\pdfoutput=1

\documentclass[11pt]{article}

\usepackage[preprint]{acl}

\usepackage{times}
\usepackage{latexsym}
\usepackage{graphicx}
\usepackage{booktabs}
\usepackage{enumitem}
\usepackage{multirow}

\usepackage{amsmath} 
\usepackage{bbm} 

\usepackage[T1]{fontenc}

\usepackage[utf8]{inputenc}

\usepackage{microtype}

\usepackage{inconsolata}
\usepackage{devanagari}

\title{Are Language Models Agnostic to Linguistically Grounded Perturbations? A Case Study of Indic Languages} 

\author{Poulami Ghosh$^{\ast}$, Raj Dabre$^{\ddagger}$, Pushpak Bhattacharyya$^{\ast}$\\ { }
$^{\ast}$IIT Bombay, India, $^{\ddagger}$NICT, Japan\\
\texttt{\{poulami, pb}\}@cse.iitb.ac.in}

\begin{document}
\maketitle
\begin{abstract}
Pre-trained language models (PLMs) are known to be susceptible to perturbations to the input text, but existing works do not explicitly focus on linguistically grounded attacks, which are subtle and more prevalent in nature. In this paper, we study whether PLMs are agnostic to linguistically grounded attacks or not. To this end, we offer the first study addressing this, investigating different Indic languages and various downstream tasks. Our findings reveal that although PLMs are susceptible to linguistic perturbations, when compared to non-linguistic attacks, PLMs exhibit a slightly lower susceptibility to linguistic attacks. This highlights that even constrained attacks are effective. Moreover, we investigate the implications of these outcomes across a range of languages, encompassing diverse language families and different scripts. 
\end{abstract}

\section{Introduction}
In this era of artificial intelligence, LLMs are everywhere. A powerful ecosystem of tools and technologies have emerged around these foundational models \cite{devlin-etal-2019-bert,liu2019roberta,radford2019language,raffel2020exploring,brown2020language}, leading to their active integration into diverse production environments and real-world applications. Originally developed and used by researchers and experts in natural language processing, language models are being used daily by people from all walks of life. Whether it's composing emails, searching the web, or interacting with virtual assistants, LLMs are now part of our daily workflow.

\begin{figure}[!t]
\centering
  \includegraphics[width=\linewidth]{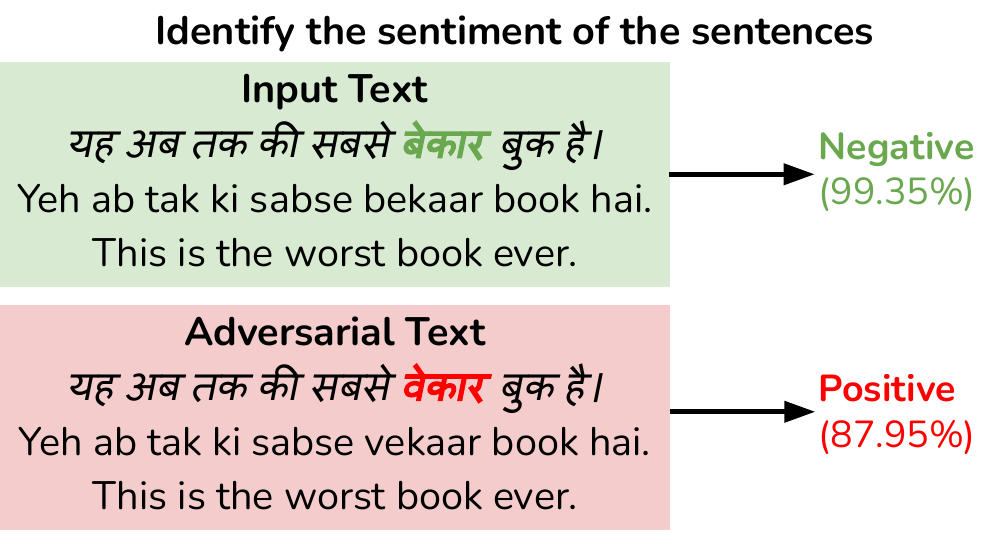}
\caption{The substitution of {\dn b}\textbf{(ba)} with orthographically similar {\dn v}\textbf{(va)} in the target word {\dn b\?kAr}\textbf{(bekaar)} causes the model to misclassify the text with high confidence. This highlights the sensitivity of language models to subtle variations in the input text, where altering a single character can lead to a significant shift in the model's output. The input text is cropped to fit within the image.}
\label{fig:overview}
\end{figure}

Despite their ground-breaking success, recent studies have found that even these powerful and advanced language models are vulnerable to adversarial attacks \cite{ribeiro2018semantically,jin2020bert,li2020bert,morris2020textattack,guo2021gradient,mehrabi2022robust,zou2023universal}. These attacks involve introducing subtle perturbations or noise into the model's input, which are undetectable to human observers, but can severely disrupt the model's performance. However, these attacks are intentionally generated in an attempt to fool the model, rather than arising organically from real-world circumstances. The artificial nature of these attacks make them less likely to be prevalent in real-world systems. As a result, they may not accurately reflect the types of challenges or threats that naturally occur in real-world settings. 

The ubiquity of language models has prompted us to investigate their robustness in settings that reflect complexities of real-world usage. We aim to evaluate how these models perform under conditions that users are likely to encounter in their daily interactions with the technology. This motivates us to design more natural adversarial attacks grounded in the principles of linguistics. Taking this into consideration, we ask a pertinent question: 
\textit{\textit{Are language models vulnerable to linguistically grounded perturbations?}}\\ 
Our contributions are:
\begin{itemize}[itemsep=1pt,topsep=2pt,parsep=0pt,partopsep=0pt]
\item We perform the first investigation on the robustness of language models to linguistically-grounded perturbations. Our findings reveal that such perturbations are effective in fooling the model.
\item We are the first to explore adversarial attacks targeting Indic language models, covering 12 Indic languages across 3 language families.
\item In our study, we develop novel adversarial attacks based on linguistic principles, with a specific emphasis on phonological and orthographic aspects. We release linguistic resources consisting of orthographically similar characters across 9 different Indic scripts (Section \ref{sec:apdx_lin_resource} in Appendix).
\end{itemize}

\section{Related work}\label{sec:related_work}
\textbf{Adversarial attacks on Text:} Adversarial attacks were first studied to test the resilience of neural networks, in the domain of computer vision \cite{kurakin2018adversarial,kurakin2016adversarial}. In recent years, research on adversarial attacks has expanded into the field of natural language processing, resulting in the introduction of many new methods for generating adversarial examples and strategies for defending against them
\cite{alzantot2018generating,li2018textbugger,gao2018black,pruthi2019combating,Jin_Jin_Zhou_Szolovits_2020}. Adversarial attacks are divided into two categories based on model access. In white-box attacks, the adversary can access the model and parameters but cannot modify it. White-box attacks are primarily gradient-based, leveraging gradient signals to effectively craft adversarial examples \cite{guo2021gradient,ebrahimi2017hotflip,wallace2019universal}. In black box attacks, the adversary can only query the target classifier without knowledge of its pre-trained weights. In our research, we carry out a black-box attack on text, where our ability to modify input samples is restricted to observing and examining the outputs of a classification model.\\
\textbf{Adversarial attacks via Token Manipulation:} The most common black-box attacks typically involve token manipulation, where a small fraction of tokens in the input text are modified to induce model failure. \citet{gao2018black,pruthi-etal-2019-combating,li2018textbugger} have investigated character-level perturbations, where a token is modified through character insertion, character deletion, neighboring character swap, or character substitution to mimic typo-based errors. Additionally, \citet{alzantot2018generating,wang2019natural,jin2020bert} have explored alternatives such as swapping a word with a similar word within a counter-fitted word embedding space.
\citet{ren2019generating} and \citet{zang2019word} use lexical knowledge bases such as WordNet and HowNet respectively, to perform  word-level substitution to generate adversarial examples. Other word-level substitution strategies include using BERT-based masked token prediction strategy to generate semantically consistent adversaries \cite{garg2020bae,li2020bert}. Prior character-level adversarial attacks studied in literature are limited in their ability to accurately reflect the real-world challenges. Moreover, while research on adversarial attacks for textual inputs has been extensively conducted for English, the exploration of adversarial attacks and robustness of large-language models for Indian languages remains largely unexplored. We draw insights from linguistics and propose linguistically grounded adversarial attacks at the character level for 12 Indic languages across 3 language families. \\ 
\textbf{Sources of Linguistic Errors:}
The recent advancements in artificial intelligence are powered by large language models (LLMs) \cite{bommasani2021opportunities}. 
LLMs \cite{devlin-etal-2019-bert,liu2019roberta,radford2019language,raffel2020exploring,brown2020language} are being increasingly integrated with various modalities beyond text, like vision and speech, enabling them to serve as interfaces bridging different modalities. Consequently, inputs to LLMs can include text extracted from images via Optical Character Recognition (OCR) and transcriptions of speech through Automatic Speech Recognition (ASR).  As LLMs become increasingly ingrained in mainstream usage, there's a growing need to assess their robustness in settings that reflect complexities of real-world usage. In OCR, the challenge lies in accurately identifying similar characters, whereas ASR struggles with distinguishing phonetically similar elements \cite{jangid2016similar,surinta2015recognition, purkaystha2017bengali,dholakia2007wavelet,choksi2012recognition,naik2017online,wakabayashi2009f}. Therefore, such phonetic and orthographic errors can easily infiltrate the system and are more likely to disrupt the model performance.  
\section{Methodology}
Based on the approaches outlined in \citet{jin2020bert,li-etal-2020-bert-attack}, we adopt the following strategy to generate adversarial text that leverages linguistic perturbations:\footnote{While our approach builds upon existing attack methodologies, implementing them for linguistically diverse languages offers insights into their effectiveness and adaptability in non-English languages. This aligns with the broader goal of extending adversarial robustness research to underexplored languages beyond English.}\\
\textbf{1. Identify Perturbation Targets:} \\ 
Under the black-box setting, only the input-output behavior of the model is accessible to the attacker. By systematically varying the inputs and observing how the outputs change, one can infer which words most affect the model’s decision. This enables the attacker to identify potential vulnerabilities without having direct access to the target model.\\ 
Let \(W = \left[ w_0, \cdots, w_{i} , \cdots, w_{n}\right]
\) be an input sentence. Our objective is to understand the importance of each word \(w_i\) in the sentence.  Specifically, we replace \(w_{i}\) with mask token \texttt{[MASK]} to nullify the influence of the word and use the difference in the prediction probability of the sentence with and without the word \(w_i\) to quantify the importance of the word. Let the modified sentence with the word \(w_{i}\) removed be $W_{\setminus w_i} = \left[ w_0, \cdots, w_{i-1}, \texttt{[MASK]}, w_{i+1}, \cdots, w_{n}\right]$. We use the notation \(p_{y}(W)\) to denote the probability assigned to the label 
\(y\) for the sentence \(W\). Let the predicted labels for the sentences \(W\) and \(W_{\setminus w_i}\) be \(y\) and \(\overline{y}\) respectively. Following \citet{jin2020bert}, we formulate the importance score \(I_{w_i}\) of a word as follows:

\begin{multline}
I_{w_i} =  
(p_{y}(W) - p_{y}(W \setminus w_i)) \\
+ \mathbbm{1}(y \neq \overline{y}) (p_{\overline{y}}(W \setminus w_i) - p_{\overline{y}}(W)), 
\end{multline}

\noindent\textbf{2. Generate Adversarial Text:} \\ 
We reorder the words in \(W\) based on decreasing importance scores. For each word, we create a candidate pool by applying various linguistic perturbations, as detailed in section \ref{sec:ling_atk}. From this pool, we choose the most appropriate candidate for substitution. Unlike \citet{jin2020bert,li-etal-2020-bert-attack}, we choose to retain stop words for two reasons: first, there are no reliable sources of stop words available for the various Indic languages included in our study. Additionally, we observe that stop words are important keywords, crucial for correct prediction. 

The candidate pool of a word \(w_i\) is generated through character-level perturbations, where a character is replaced with another that is phonetically or orthographically similar. Each candidate word only has a single character altered. From the final pool of candidates, we select the perturbed word that successfully changes the sentence's prediction upon substitution. If no such candidate is found, we select the one that results in the largest reduction in confidence for predicting label \(y\). 

\section{Linguistically-grounded Perturbations}\label{sec:ling_atk} 
Our primary objective is to design adversarial attacks by introducing perturbations grounded in linguistic principles. The motivation for this approach stems from the observation that common sources of confusion and error in NLP systems often arise from subtle linguistic variations that are often difficult to detect. This imperceptibility makes linguistic perturbations an ideal candidate for crafting adversarial attacks that can easily infiltrate and disrupt AI models. In this paper, we focus various character-level  perturbations\footnote{We used the terms perturbation and attack interchangeably throughout the paper.} which are simple yet effective at fooling even advanced AI systems. To construct adversarial attacks that exploit linguistic vulnerabilities, we develop the following perturbation strategies: 
\subsection{Phonological Perturbations}
Phonological variations can stem from various sources and can often cause ambiguity. New learners of a language usually struggle to distinguish between phonetically similar sounds that do not exist in their native language. Moreover, non-native speakers also may pronounce words differently, heavily influenced by their first language. Variations in pronunciation, speed, or stress patterns can lead to misidentification of phonemes, resulting in errors in ASR systems. \\
In our study, we focus on perturbation of the two main categories of speech sounds, vowels and consonants. For instance, short and long vowels are phonetically very similar, differing primarily in the length of the utterance. Moreover, the consonants are grouped into Vargas. Each consonant in a Varga has the same place of articulation and differs from other consonants in the Varga along a single dimension (voiced/unvoiced, aspirated/unaspirated), as depicted in Table \ref{fig:consonants_deva} (in Appendix). The sibilants are also often confused with each other due to their high phonetic resemblance. Hence, we study the following phonetic perturbations: (1) Replacing a short vowel with the corresponding long vowel and vice-versa (2) Substituting a consonant with a homorganic consonant (except the nasals) (3) Replacing a Sibilant with another Sibilant. These phonetic perturbations were designed in consultation with linguists. Based on Table \ref{fig:consonants_deva}, we curated the set of phonetically similar vowels and consonants for Devanagari script. Using the script conversion functionality present in the Indic NLP library \cite{kunchukuttan2020indicnlp}, we extended the set across other Indic scripts.

\subsection{Orthographic Perturbations}
\begin{figure}[!t]
\centering
  \includegraphics[width=\linewidth]{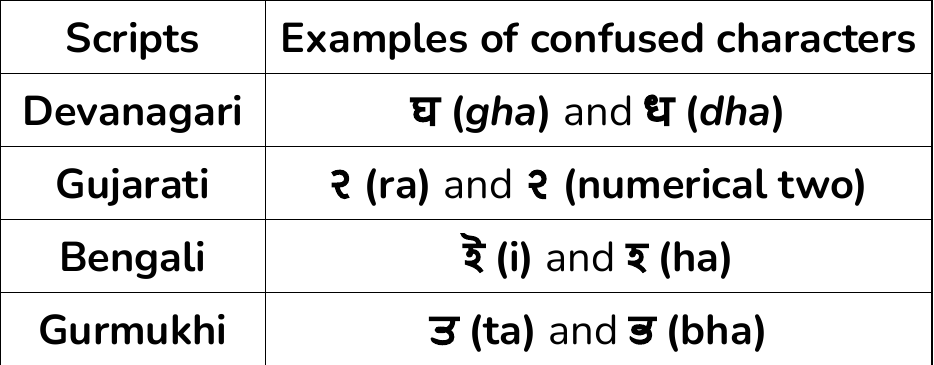}
\caption{Similar characters across different scripts}
\label{tab:sim_char}
\end{figure}

Distinguishing similar characters poses a notable challenge in handwritten and optical character recognition systems. Consequently, orthographic errors frequently originate from such systems and may cause a cascading impact on the performance of downstream tasks. It is common even for humans to get confused between similar-looking characters in a language, as evident in Figure \ref{tab:sim_char}. Moreover, new learners of a language often struggle to identify the correct order of constituents in conjunct consonants and can cause confusion. We perform the following orthographic changes in an attempt to fool the model: (1) Replace a character with a similar-looking character from the same script \footnote{In our study, both phonological and orthographic attacks involve perturbations at the character level, with some degree of overlap between them. For instance, certain short and long vowels in Indic scripts bear a high resemblance and differ by only a few strokes.} (2) Swap the constituent characters in a conjunct consonant. A collection of visually similar characters across 9 different Indic scripts was curated by an in-house expert. To ensure the objectivity and correctness of the resource, it was subsequently reviewed by multiple linguists. This resource is provided in the Appendix of the paper.

We also evaluated the robustness of Indic language models by introducing perturbations at the level of semantics. We performed experiments involving \textbf{synonym-based word substitution} as described in Section \ref{sec:synonym} in the Appendix. We utilized IndoWordNet \cite{bhattacharyya-2010-indowordnet} as a linguistic resource to obtain the synonyms of different words.

\begin{table*}[]
\centering
\resizebox{\textwidth}{!}{%
\begin{tabular}{@{}cc|cccc|cccc|cccc@{}}
\toprule
\multicolumn{2}{c|}{\multirow{2}{*}{}} &
  \multicolumn{4}{c|}{\textbf{IndicSentiment}} &
  \multicolumn{4}{c|}{\textbf{IndicParaphrase}} &
  \multicolumn{4}{c}{\textbf{IndicXNLI}} \\ \cmidrule(l){3-14} 
\multicolumn{2}{c|}{} &
  \textbf{\begin{tabular}[c]{@{}c@{}}Original\\ Accuracy\end{tabular}} &
  \textbf{\begin{tabular}[c]{@{}c@{}}After-Attack\\ Accuracy\end{tabular}} &
  \textbf{\begin{tabular}[c]{@{}c@{}}\% Perturbed\\  Words\end{tabular}} &
  \textbf{\begin{tabular}[c]{@{}c@{}}Query \\ Number\end{tabular}} &
  \textbf{\begin{tabular}[c]{@{}c@{}}Original\\ Accuracy\end{tabular}} &
  \textbf{\begin{tabular}[c]{@{}c@{}}After-Attack\\ Accuracy\end{tabular}} &
  \textbf{\begin{tabular}[c]{@{}c@{}}\% Perturbed\\  Words\end{tabular}} &
  \textbf{\begin{tabular}[c]{@{}c@{}}Query \\ Number\end{tabular}} &
  \textbf{\begin{tabular}[c]{@{}c@{}}Original\\ Accuracy\end{tabular}} &
  \textbf{\begin{tabular}[c]{@{}c@{}}After-Attack\\ Accuracy\end{tabular}} &
  \textbf{\begin{tabular}[c]{@{}c@{}}\% Perturbed\\  Words\end{tabular}} &
  \textbf{\begin{tabular}[c]{@{}c@{}}Query \\ Number\end{tabular}} \\ \midrule
\multirow{3}{*}{\textbf{IndicBERT}} &
  \textbf{Rand} &
  \multirow{3}{*}{0.937} &
  0.194 &
  16.985 &
  43.75 &
  \multirow{3}{*}{0.565} &
  0.114/0.09 &
  10.175/10.568 &
  16.438/16.853 &
  \multirow{3}{*}{0.714} &
  0.214/0.099 &
  9.056/13.511 &
  17.99/10.629 \\ \cmidrule(lr){2-2} \cmidrule(lr){4-6} \cmidrule(lr){8-10} \cmidrule(l){12-14} 
 &
  \textbf{Phono} &
   &
  0.276 &
  17.82 &
  39.187 &
   &
  0.153/0.128 &
  9.861/10.621 &
  13.889/14.539 &
   &
  0.268/0.183 &
  9.081/13.652 &
  16.027/9.543 \\ \cmidrule(lr){2-2} \cmidrule(lr){4-6} \cmidrule(lr){8-10} \cmidrule(l){12-14} 
 &
  \textbf{Ortho} &
   &
  0.462 &
  17.109 &
  28.899 &
   &
  0.198/0.168 &
  9.448/10.32 &
  11.782/12.464 &
   &
  0.32/0.236 &
  8.132/12.932 &
  13.76/8.35 \\ \midrule
\multirow{3}{*}{\textbf{MuRIL}} &
  \textbf{Rand} &
  \multirow{3}{*}{0.855} &
  0.101 &
  10.945 &
  34.861 &
  \multirow{3}{*}{0.606} &
  0.205/0.142 &
  10.406/10.253 &
  18.092/17.913 &
  \multirow{3}{*}{0.725} &
  0.188/0.093 &
  9.629/13.622 &
  19.323/10.896 \\ \cmidrule(lr){2-2} \cmidrule(lr){4-6} \cmidrule(lr){8-10} \cmidrule(l){12-14} 
 &
  \textbf{Phono} &
   &
  0.189 &
  12.295 &
  32.092 &
   &
  0.36/0.281 &
  8.263/9.432 &
  12.4/13.934 &
   &
  0.296/0.222 &
  9.876/14.081 &
  17.398/10.028 \\ \cmidrule(lr){2-2} \cmidrule(lr){4-6} \cmidrule(lr){8-10} \cmidrule(l){12-14} 
 &
  \textbf{Ortho} &
   &
  0.234 &
  12.275 &
  29.064 &
   &
  0.333/0.255 &
  8.282/9.146 &
  11.322/12.443 &
   &
  0.318/0.243 &
  8.685/13.251 &
  14.549/8.678 \\ \midrule
\multirow{3}{*}{\textbf{XLMR}} &
  \textbf{Rand} &
  \multirow{3}{*}{0.814} &
  0.176 &
  10.066 &
  31.454 &
  \multirow{3}{*}{0.569} &
  0.086/0.074 &
  9.162/8.969 &
  15.814/15.676 &
  \multirow{3}{*}{0.695} &
  0.203/0.096 &
  9.193/13.026 &
  17.885/10.356 \\ \cmidrule(lr){2-2} \cmidrule(lr){4-6} \cmidrule(lr){8-10} \cmidrule(l){12-14} 
 &
  \textbf{Phono} &
   &
  0.237 &
  10.337 &
  28.422 &
   &
  0.112/0.097 &
  9.688/9.682 &
  14.336/14.376 &
   &
  0.258/0.17 &
  9.257/13.591 &
  15.91/9.359 \\ \cmidrule(lr){2-2} \cmidrule(lr){4-6} \cmidrule(lr){8-10} \cmidrule(l){12-14} 
 &
  \textbf{Ortho} &
   &
  0.373 &
  9.531 &
  22.266 &
   &
  0.125/0.117 &
  9.562/9.428 &
  12.478/12.485 &
   &
  0.315/0.238 &
  8.295/12.402 &
  13.586/8.032 \\ \midrule
\multirow{3}{*}{\textbf{mBERT}} &
  \textbf{Rand} &
  \multirow{3}{*}{0.636} &
  0.082 &
  7.843 &
  23.431 &
  \multirow{3}{*}{0.556} &
  0.113/0.101 &
  9.275/8.582 &
  15.748/14.947 &
  \multirow{3}{*}{0.561} &
  0.128/0.068 &
  7.342/10.078 &
  14.578/8.077 \\ \cmidrule(lr){2-2} \cmidrule(lr){4-6} \cmidrule(lr){8-10} \cmidrule(l){12-14} 
 &
  \textbf{Phono} &
   &
  0.15 &
  7.737 &
  20.842 &
   &
  0.199/0.178 &
  9.25/9.05 &
  13.385/13.322 &
   &
  0.192/0.122 &
  7.494/10.335 &
  12.961/7.287 \\ \cmidrule(lr){2-2} \cmidrule(lr){4-6} \cmidrule(lr){8-10} \cmidrule(l){12-14} 
 &
  \textbf{Ortho} &
   &
  0.208 &
  7.201 &
  18.221 &
   &
  0.185/0.169 &
  8.943/8.664 &
  11.859/11.702 &
   &
  0.22/0.162 &
  6.58/9.53 &
  11.024/6.283 \\ \bottomrule
\end{tabular}%
}
\caption{The table presents the impact of different character-substitution attack strategies: random (\textbf{Rand}), phonetic (\textbf{Phono}), and orthographic (\textbf{Ortho}) on various language models, with the results averaged across different languages. Random substitution causes the largest accuracy drop compared to phonetic and orthographic attacks. For detailed results, please refer to }
\label{tab:attack_results_across_lms}
\end{table*}

\section{Experimental Setup}
To ensure the generalizability of our findings, we 
conduct an extensive investigation covering a range of languages and tasks.
\subsection{Tasks}
We selected three NLU tasks from the IndicXTREME \cite{doddapaneni2023towards} benchmark, namely \textit{IndicSentiment, IndicXParaphrase, IndicXNLI}.

\subsection{Models}
We have considered four language models for our study - IndicBERTv2 (278M) \cite{doddapaneni-etal-2023-towards}, Muril (236M) \cite{khanuja2021muril}, XLMR (270M) \cite{conneau-etal-2020-unsupervised} and mBERT (110M). For fine-tuning these models, we used the hyper-parameter settings given in \citet{doddapaneni-etal-2023-towards}. Results are reported across 3 random trials of each experiment.

\subsection{Languages and Scripts}
We choose various languages from two prominent language families in India: Indo-Aryan and Dravidian, as detailed in table \ref{tab:lang}. Within the Indo-Aryan family, we focus on languages with diverse scripts.

\subsection{Automatic Evaluation}
We impose constraints on the similarity scores to ensure semantic and syntactic consistency between the original and altered sentences: (1) Employing Language-agnostic BERT Sentence Embedding (LaBSE) \cite{feng2020language} as a sentence encoder, we encode the two sentences and utilize their cosine similarity score as an indicator of semantic similarity. (2) We use the chrF score \cite{popovic-2015-chrf} to compute the overlap between the original and adversarial text. (3) BERTScore \cite{zhang2019bertscore} is employed to capture both semantic and overlap-based similarity between the two sentences. (4) Additionally, we incorporate a phonetic similarity measure from the IndicNLP library \cite{kunchukuttan2020indicnlp}.\\
Based on experiments with various threshold values, we selected a threshold of 0.6 to balance the attack success rate with imperceptibility requirements, ensuring that adversarial attacks remain both effective and difficult to detect. This is supported by results from human evaluation, where participants consistently found that adversarial examples generated with this threshold were challenging to distinguish from the original samples. 

\section{Results}
In this section, we perform experiments to assess the robustness of LLMs from a linguistic standpoint. 
\begin{itemize}[itemsep=1pt,topsep=2pt,parsep=0pt,partopsep=0pt]
\item[Q1.] Can large language models (LLMs) handle linguistic perturbations in the input text effectively? (Section \ref{sec:lin_perturb}) 
\item[Q2.] How does the nature of the attack (linguistic or non-linguistic) impact the model's performance? ((Section \ref{sec:lin_vs_nonlin})
\item[Q3.] Are specific languages or language families more susceptible to certain attacks? (Section \ref{sec:lang_families})
\item[Q4.] How does the robustness of language models vary across different tasks? \ref{sec:lin_perturb})
\end{itemize}

\subsection{Adversarial Robustness against Linguistic Perturbations}
\subsubsection{Robustness across Language Models}\label{sec:lin_perturb}

In this section, we investigate how subjecting language models to various linguistic perturbations impacts their performance. For our study, we have focused on three fundamental tasks: \textit{sentiment analysis, paraphrasing, and natural language inference}. The results presented in Table \ref{tab:attack_results_across_lms} highlight varying trends in the robustness of language models across different tasks.

For the IndicSentiment dataset, models with higher pre-attack accuracy generally exhibit greater resilience to linguistic perturbations, with IndicBERTv2 outperforming others in both performance and robustness. In contrast, models like MURIL, XLM-R, and mBERT experience a more significant decline in performance with fewer perturbations, indicating lower resistance to such perturbations. For the IndicXNLI dataset, the impact of linguistic perturbations on different models remains relatively consistent, with MURIL and IndicBERTv2 showing comparatively higher robustness. However, the observations in the IndicXParaphrase dataset differs from the other datasets in our study. Here, MURIL demonstrates the highest overall performance and robustness against perturbations, while XLM-R struggles the most with maintaining performance when subjected to linguistic perturbations compared to the other models.

\begin{figure*}[!t]
\centering
  \includegraphics[width=\textwidth]{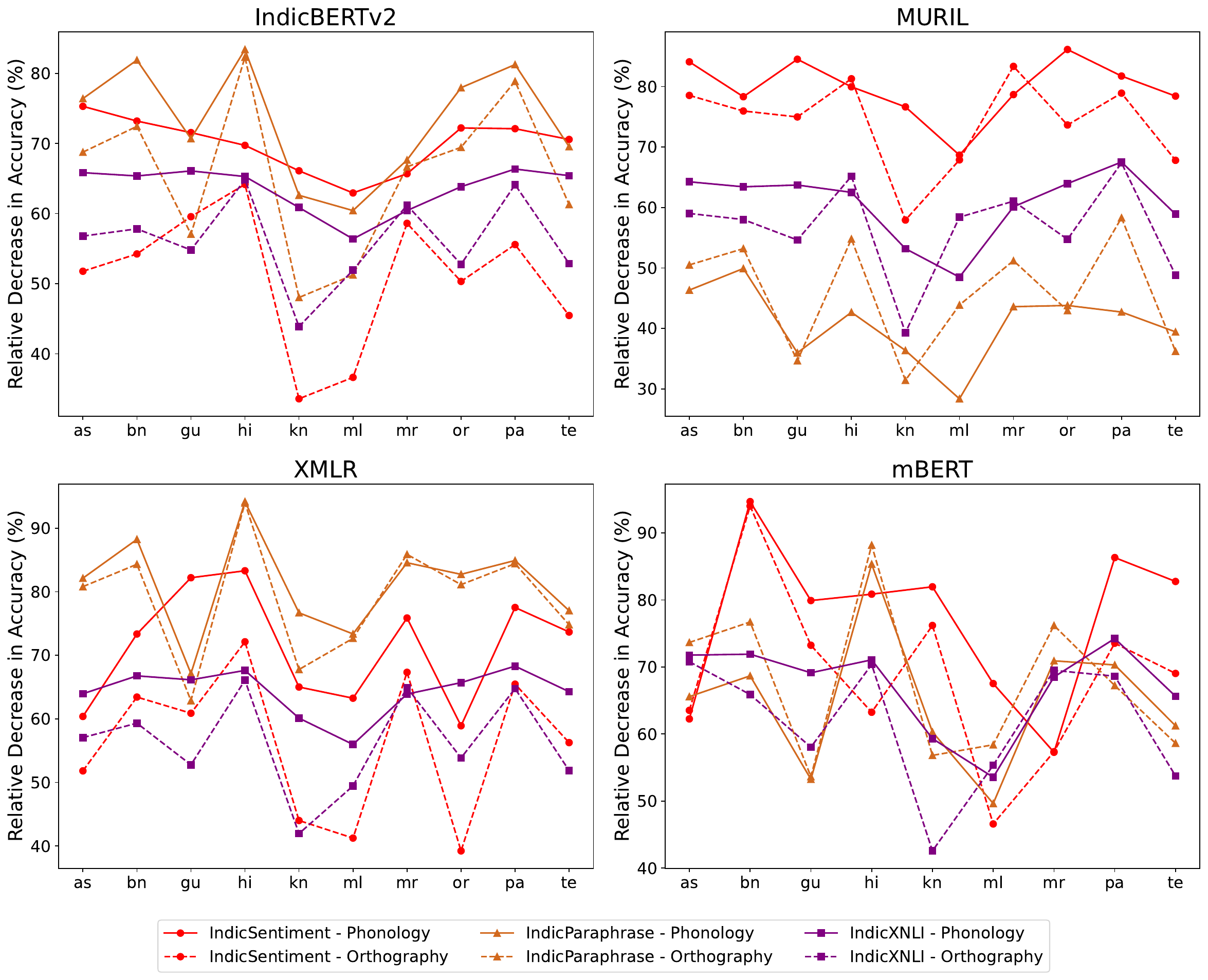}
\caption{The figure highlights the impact of linguistic perturbations across different languages. The x-axis lists the ISO codes for the different languages, as specified in Table \ref{tab:lang}.  Bodo (bd) and Tamil (ta) are not present across all tasks. Therefore, to maintain consistency across all tasks, we have present the remaining 10 languages in the plot. For IndicParaphrase and IndicXNLI, we illustrate the relative decrease in performance due to perturbations in sentence1 and premise, respectively.}
\label{fig:lang_adv_attack}
\end{figure*}

\subsubsection{Vulnerability to Linguistic Perturbations}
The results indicate that different language models have different levels of resistance to linguistic perturbations across different tasks.\\
For the IndicSentiment dataset, we observe that phonological perturbations cause more damage than orthographic perturbations. While MURIL and mBERT are similarly impacted by both types of perturbations, IndicBERTv2 and XLMR perform significantly worse when subjected to phonological perturbations and exhibit greater resilience against orthographic attacks. A similar trend is observed in the IndicXNLI dataset, where phonological perturbations more effectively disrupt model performance compared to orthographic perturbations.
The effectiveness of phonological perturbations can be attributed to the larger pool of perturbed candidates they generate, compared to orthographic perturbations. With a effectively larger search space, phonological perturbations have a greater chance of generating perturbations that can effectively mislead a model. However, in the IndicParaphrase dataset, a different pattern emerges: IndicBERTv2 and XLM-R are more sensitive to phonological perturbations, while MURIL and mBERT are more affected by orthographic perturbations.\\ 
Overall, the evaluation reveals that different models have varying strengths and weaknesses in handling linguistic perturbations. Moreover, the notable decline in performance observed across various languages models and tasks highlight the vulnerability of language models to linguistic noise introduced in their inputs.

\subsubsection{Linguistic v/s Non-Linguistic Perturbations}\label{sec:lin_vs_nonlin}
Among the various random attacks existing in literature, the phonological and orthographic attacks are most akin to random character substitution attacks. Various Indic scripts comprise of three different character types: consonant letters, independent vowels, and dependent vowel signs. Hence, for random character substitution attack, we substitute a letter with a random letter of the same type i.e. replace a vowel (both independent vowel and dependent vowel sign) with a random vowel (excluding the original vowel/vowel sign), replace a consonant with a random consonant (excluding the original consonant). Note that, in this case the original and perturbed character may or may not be linguistically related due to the random nature of the attack. However, in linguistic attacks, we enforce substitution of a character with a linguistically (phonetically/orthographically) similar character. Due to unconstrained search space, higher number of perturbed candidates are generated in case of random attacks compared to linguistic attacks, contributing to their higher attack success. Therefore, we observe that linguistic attacks successfully deceive the model but cause less harm compared to non-linguistic random attacks.
Given that IndicBERTv2 demonstrated consistent performance across different tasks and languages, significant robustness against various perturbations, and has the highest coverage of Indic languages, we selected this model for further experimentation.

\begin{figure*}[!t]
\centering
  \includegraphics[width=\textwidth]{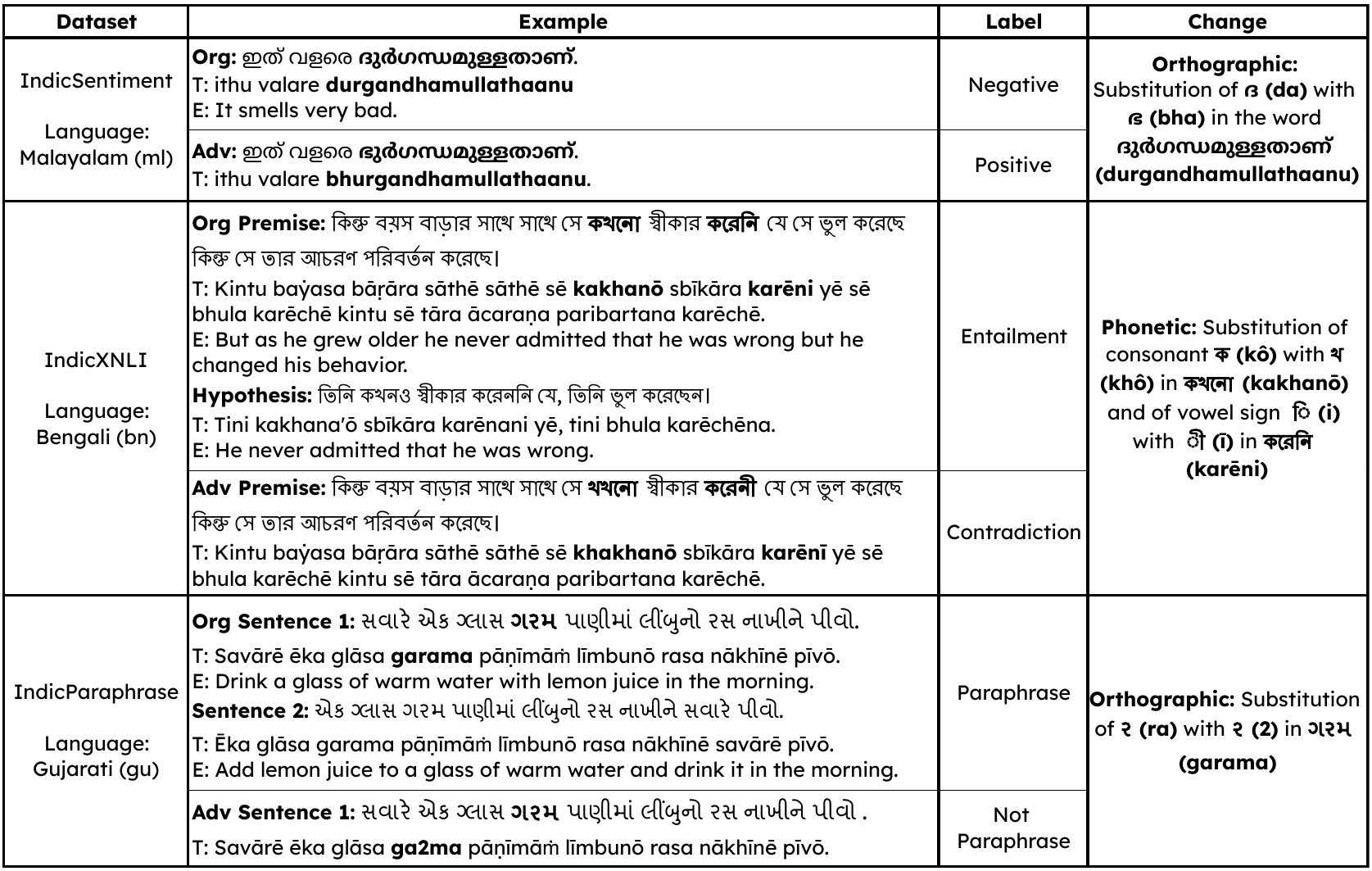}
\caption{Examples of generated adversarial text across different languages}
\label{fig:adv_attack_example}
\end{figure*}

\subsubsection{Robustness across Languages and Language Families}\label{sec:lang_families}
The impact of linguistic perturbations across different languages is presented in the figure \ref{fig:lang_adv_attack}. Detailed results for a few selected languages are presented in Table 3. In this section, we focus our discussion on IndicBERTv2.
Hindi, Marathi, and Bodo all share the same script, Devanagari. Bodo, which is only available in the IndicSentiment dataset, shows the least robustness and worst performance across all languages. Across all tasks, Hindi is more vulnerable than Marathi to both phonetic and orthographic perturbations. Assamese and Bengali share the same script, but their sensitivity to different perturbations varies by task and type of perturbation. In the IndicParaphrase dataset, Bengali is more sensitive to linguistic perturbations than Assamese. For the IndicXNLI dataset, both languages exhibit similar robustness. However, in the IndicSentiment dataset, Bengali is more robust to phonological perturbations, while Assamese shows greater resistance to orthographic perturbations.
Across different language families considered in our study, we find that languages from the Sino-Tibetan family are generally the least robust to linguistic perturbations, while those from the Dravidian family demonstrate the highest robustness. Languages belonging to the Indo-Aryan language family lie in between in the spectrum of robustness. These observations can be attributed to two probable factors. Firstly, languages belonging to the Sino-Tibetan language family are low resource and under represented in the training corpus, making them more susceptible to perturbations. Secondly, languages in the Dravidian family are linguistically rich with complex morphology. Moreover, due to their agglutinative nature, words in these languages are often longer, meaning that perturbing a single character may have a lesser impact on the overall word structure. This likely contributes to their greater robustness against linguistic perturbations.

\subsubsection{Fine-grained Analysis of Phonological Perturbations}
We conducted a fine-grained study of phonological perturbations, subdividing these attacks into three categories: i) substitution of homorganic consonants, ii) substitution of short/long vowels, and iii) substitution of sibilants. The homorganic consonant substitution attack is further divided into a) substitution of aspirated/unaspirated consonants and b) substitution of voiced/voiceless consonants. The detailed results of these fine-grained phonological attacks on the IndicSentiment dataset for IndicBERTv2 model are summarized in Table \ref{tab:fg_phono_attacks} in the Appendix.
Orthographic attacks do not easily allow for a fine-grained categorization based on linguistic principles.

For most languages, except for Tamil (ta), consonant-based perturbation causes more damage to accuracy compared to vowel-based perturbation. This could be attributed to the relatively fewer consonants in the Tamil language than others. Among the variants of consonant-based substitution, both aspirated/unaspirated and voiced/unvoiced are similarly impactful. However, substitutions involving sibilants result in the least reduction in accuracy, likely due to the limited number of sibilants present across languages.

\begin{table}[!h]
\centering
\resizebox{\columnwidth}{!}{%
\begin{tabular}{@{}cccccc@{}}
\toprule
\multicolumn{3}{c}{\textbf{Language}} & \textbf{Accuracy} & \textbf{Grammar} & \textbf{Similarity} \\ \midrule
\multirow{6}{*}{Hindi}   & \multirow{2}{*}{Phonology}   & Org & 0.94 & 4.607 & \multirow{2}{*}{0.887} \\ \cmidrule(lr){3-5}
                         &                              & Adv & 0.9  & 4.28  &                        \\ \cmidrule(l){2-6} 
                         & \multirow{2}{*}{Orthography} & Org & 0.98 & 4.627 & \multirow{2}{*}{0.843} \\ \cmidrule(lr){3-5}
                         &                              & Adv & 0.86 & 3.84  &                        \\ \cmidrule(l){2-6} 
                         & \multirow{2}{*}{Random}      & Org & 0.94 & 4.579 & \multirow{2}{*}{0.789} \\ \cmidrule(lr){3-5}
                         &                              & Adv & 0.86 & 3.407 &                        \\ \midrule
\multirow{6}{*}{Bengali} & \multirow{2}{*}{Phonology}   & Org & 0.91 & 4.16  & \multirow{2}{*}{0.86}  \\ \cmidrule(lr){3-5}
                         &                              & Adv & 0.82 & 3.896 &                        \\ \cmidrule(l){2-6} 
                         & \multirow{2}{*}{Orthography} & Org & 0.94 & 4.247 & \multirow{2}{*}{0.848} \\ \cmidrule(lr){3-5}
                         &                              & Adv & 0.8  & 3.721 &                        \\ \cmidrule(l){2-6} 
                         & \multirow{2}{*}{Random}      & Org & 0.98 & 4.28  & \multirow{2}{*}{0.797} \\ \cmidrule(lr){3-5}
                         &                              & Adv & 0.68 & 3.38  &                        \\ \bottomrule
\end{tabular}%
}
\caption{Human evaluation results}
\label{tab:human_eval}
\end{table}
\subsection{Human Evaluation}
Following the process described in the paper by \cite{jin2020bert}, we conducted a human evaluation for three types of adversarial attacks: random, phonology-based, and orthography-based character substitutions. We utilized the IndicSentiment dataset along with the IndicBERTv2 model for two languages, Hindi and Bengali. We randomly selected 50 test sentences from each language to generate adversarial examples. For label prediction, we determined the predicted label based on the majority class. The final score for semantic similarity and grammaticality were calculated by averaging the scores given by all annotators. For each language, the evaluation was performed by three
independent humans who are native speakers in the language and have university-level education background. For phonetic perturbations, we provided audio samples of the examples to the human evaluators while for orthographic perturbations, written text was provided.\\
As shown in Table \ref{tab:human_eval}, linguistic perturbations result in grammaticality and semantic similarity scores that are much closer to the original text when compared to random perturbations. Additionally, the semantic similarity and grammaticality scores for random attacks are significantly lower, highlighting that linguistic perturbations are generally more subtle and less detectable by humans.

\section{Conclusion}
In conclusion, this paper addresses the vulnerability of PLMs to adversarial attacks, focusing specifically on linguistically grounded attacks, which are subtle and prevalent in real-world settings but often overlooked in existing research. Our study investigates whether PLMs are affected by linguistically grounded attacks, marking the first comprehensive examination across various Indic languages and tasks. The findings demonstrate that PLMs are indeed vulnerable to linguistic perturbations. We further analyze the results across diverse languages, spanning different language families and scripts. Our analysis reveals that while both linguistic and non-linguistic attacks pose challenges to PLMs, the latter are more effective at deceiving the model.

Future research could investigate the effectiveness of applying the proposed methods to decoder-based language models for Indic languages in order to assess their broader applicability. Additionally, future work may explore the potential of adversarial training to mitigate the effects of linguistic perturbations.\\

\section{Limitations}
A limitation of our study is the inability to assess the effectiveness of syntactic and morphological perturbations for Indian languages due to insufficient resources. Although we designed these attacks to replace words with their inflected forms, the lack of adequate linguistic resources restricted our ability to fully evaluate their impact. Additionally, the unavailability of high-quality dependency parsers for Indic languages hindered our ability to investigate the effect of syntactic perturbations.

\section{Ethical Considerations}
Perturbations can cause models to hallucinate and potentially produce toxic and harmful content. Our research focuses on the impact of linguistic attacks to understand whether the model recognizes the difference between linguistic and non-linguistic perturbations. We do not intend it to be used for harmful purposes but for model understanding and expect that our perturbation methods will be used for identifying cases where harmful content will be produced in order to mitigate it. All our perturbation mapping datasets were manually curated by ourselves and no other humans were involved.

\appendix

\section{Languages}
Details of languages included in our study is outlined in Table \ref{tab:lang}.

\begin{table}[!h]
\centering
\resizebox{\columnwidth}{!}{%
\begin{tabular}{@{}ccc@{}}
\toprule
\textbf{Language} & \textbf{Languages Family} & \textbf{Script}  \\ \midrule
Assamese (as)     & Indo-Aryan                & Bengali–Assamese \\
Bodo (bd)         & Sino-Tibetan              & Devanagari       \\
Bengali (bn)      & Indo-Aryan                & Bengali–Assamese \\
Gujarati (gu)     & Indo-Aryan                & Gujarati         \\
Hindi (hi)        & Indo-Aryan                & Devanagari       \\
Kannada (kn)      & Dravidian                 & Kannada          \\
Malayalam (ml)    & Dravidian                 & Malayalam        \\
Marathi (mr)      & Indo-Aryan                & Devanagari       \\
Odia (or)         & Indo-Aryan                & Odia             \\
Punjabi (pa)      & Indo-Aryan                & Gurmukhi         \\
Tamil (ta)        & Dravidian                 & Tamil            \\
Telugu (te)       & Dravidian                 & Telugu           \\ \bottomrule
\end{tabular}%
}
\caption{Overview of different languages}
\label{tab:lang}
\end{table}

\section{Linguistic Resources}\label{sec:apdx_lin_resource}
Table \ref{fig:consonants_deva} displays the different consonants in the Devanagari script, along with their various features of articulation. Tables \ref{fig:Visually_similar_character_1} and \ref{fig:Visually_similar_character_2} present the list of visually similar characters across different Indic scripts.

\section{Fine-grained Analysis of Phonological Attacks}
Table \ref{tab:fg_phono_attacks} provide the detailed results on the fine-grained analysis of phonological perturbations performed on the IndicSentiment dataset with the IndicBERTv2 model. 
\begin{table*}[!h]
\centering
\resizebox{\textwidth}{!}{%
\begin{tabular}{@{}ccccccccc@{}}
\toprule
\textbf{Language} &
  \textbf{Type of Perturbation} &
  \textbf{Original Accuracy} &
  \textbf{After-Attack Accuracy} &
  \textbf{\% Perturbed Words} &
  \textbf{Semantic Similarity} &
  \textbf{Overlap Similarity} &
  \textbf{BERTScore based Similarity} &
  \textbf{Phonetic Similarity} \\ \midrule
\multirow{6}{*}{\textbf{as}} & \textbf{Consonants} & \multirow{6}{*}{0.931} & 0.371 & 17.854 & 0.93  & 0.868 & 0.959 & 0.87  \\ \cmidrule(lr){2-2} \cmidrule(l){4-9} 
                             & \textbf{Vowels}     &                        & 0.413 & 18.955 & 0.938 & 0.861 & 0.961 & 0.888 \\ \cmidrule(lr){2-2} \cmidrule(l){4-9} 
                             & \textbf{Aspirated}  &                        & 0.387 & 18.357 & 0.931 & 0.866 & 0.959 & 0.873 \\ \cmidrule(lr){2-2} \cmidrule(l){4-9} 
                             & \textbf{Voiced}     &                        & 0.341 & 17.899 & 0.932 & 0.87  & 0.961 & 0.873 \\ \cmidrule(lr){2-2} \cmidrule(l){4-9} 
                             & \textbf{Sibilants}  &                        & 0.83  & 5.454  & 0.974 & 0.953 & 0.981 & 0.877 \\ \cmidrule(lr){2-2} \cmidrule(l){4-9} 
                             & \textbf{Overall}    &                        & 0.23  & 18.11  & 0.932 & 0.867 & 0.96  & 0.882 \\ \midrule
\multirow{6}{*}{\textbf{bd}} & \textbf{Consonants} & \multirow{6}{*}{0.859} & 0.239 & 15.691 & 0.965 & 0.897 & 0.976 & -     \\ \cmidrule(lr){2-2} \cmidrule(l){4-9} 
                             & \textbf{Vowels}     &                        & 0.194 & 15.331 & 0.979 & 0.898 & 0.978 & -     \\ \cmidrule(lr){2-2} \cmidrule(l){4-9} 
                             & \textbf{Aspirated}  &                        & 0.246 & 15.761 & 0.973 & 0.896 & 0.976 & -     \\ \cmidrule(lr){2-2} \cmidrule(l){4-9} 
                             & \textbf{Voiced}     &                        & 0.243 & 15.624 & 0.962 & 0.896 & 0.976 & -     \\ \cmidrule(lr){2-2} \cmidrule(l){4-9} 
                             & \textbf{Sibilants}  &                        & 0.784 & 3.225  & 0.993 & 0.979 & 0.992 & -     \\ \cmidrule(lr){2-2} \cmidrule(l){4-9} 
                             & \textbf{Overall}    &                        & 0.138 & 14.866 & 0.972 & 0.902 & 0.977 & -     \\ \midrule
\multirow{6}{*}{\textbf{bn}} & \textbf{Consonants} & \multirow{6}{*}{0.955} & 0.368 & 17.799 & 0.912 & 0.866 & 0.95  & 0.971 \\ \cmidrule(lr){2-2} \cmidrule(l){4-9} 
                             & \textbf{Vowels}     &                        & 0.456 & 18.525 & 0.927 & 0.864 & 0.954 & 0.927 \\ \cmidrule(lr){2-2} \cmidrule(l){4-9} 
                             & \textbf{Aspirated}  &                        & 0.37  & 18.067 & 0.914 & 0.866 & 0.95  & 0.973 \\ \cmidrule(lr){2-2} \cmidrule(l){4-9} 
                             & \textbf{Voiced}     &                        & 0.387 & 17.143 & 0.915 & 0.872 & 0.954 & 0.974 \\ \cmidrule(lr){2-2} \cmidrule(l){4-9} 
                             & \textbf{Sibilants}  &                        & 0.85  & 6.402  & 0.97  & 0.947 & 0.975 & 0.979 \\ \cmidrule(lr){2-2} \cmidrule(l){4-9} 
                             & \textbf{Overall}    &                        & 0.256 & 17.759 & 0.919 & 0.867 & 0.953 & 0.947 \\ \midrule
\multirow{6}{*}{\textbf{gu}} & \textbf{Consonants} & \multirow{6}{*}{0.943} & 0.431 & 17.9   & 0.907 & 0.858 & 0.953 & 0.983 \\ \cmidrule(lr){2-2} \cmidrule(l){4-9} 
                             & \textbf{Vowels}     &                        & 0.521 & 18.984 & 0.922 & 0.85  & 0.957 & 0.93  \\ \cmidrule(lr){2-2} \cmidrule(l){4-9} 
                             & \textbf{Aspirated}  &                        & 0.447 & 18.159 & 0.913 & 0.855 & 0.953 & 0.985 \\ \cmidrule(lr){2-2} \cmidrule(l){4-9} 
                             & \textbf{Voiced}     &                        & 0.471 & 17.105 & 0.912 & 0.862 & 0.956 & 0.985 \\ \cmidrule(lr){2-2} \cmidrule(l){4-9} 
                             & \textbf{Sibilants}  &                        & 0.875 & 5.923  & 0.968 & 0.946 & 0.979 & 0.99  \\ \cmidrule(lr){2-2} \cmidrule(l){4-9} 
                             & \textbf{Overall}    &                        & 0.268 & 18.951 & 0.907 & 0.851 & 0.953 & 0.951 \\ \midrule
\multirow{6}{*}{\textbf{hi}} & \textbf{Consonants} & \multirow{6}{*}{0.958} & 0.376 & 15.203 & 0.913 & 0.859 & 0.952 & 0.983 \\ \cmidrule(lr){2-2} \cmidrule(l){4-9} 
                             & \textbf{Vowels}     &                        & 0.486 & 16.179 & 0.93  & 0.852 & 0.96  & 0.941 \\ \cmidrule(lr){2-2} \cmidrule(l){4-9} 
                             & \textbf{Aspirated}  &                        & 0.369 & 14.827 & 0.918 & 0.862 & 0.952 & 0.985 \\ \cmidrule(lr){2-2} \cmidrule(l){4-9} 
                             & \textbf{Voiced}     &                        & 0.38  & 14.886 & 0.916 & 0.862 & 0.954 & 0.986 \\ \cmidrule(lr){2-2} \cmidrule(l){4-9} 
                             & \textbf{Sibilants}  &                        & 0.865 & 5.426  & 0.969 & 0.942 & 0.977 & 0.99  \\ \cmidrule(lr){2-2} \cmidrule(l){4-9} 
                             & \textbf{Overall}    &                        & 0.29  & 16.426 & 0.921 & 0.852 & 0.954 & 0.959 \\ \midrule
\multirow{6}{*}{\textbf{kn}} & \textbf{Consonants} & \multirow{6}{*}{0.938} & 0.462 & 18.519 & 0.931 & 0.899 & 0.955 & 0.985 \\ \cmidrule(lr){2-2} \cmidrule(l){4-9} 
                             & \textbf{Vowels}     &                        & 0.504 & 18.353 & 0.94  & 0.898 & 0.962 & 0.951 \\ \cmidrule(lr){2-2} \cmidrule(l){4-9} 
                             & \textbf{Aspirated}  &                        & 0.473 & 18.525 & 0.929 & 0.898 & 0.954 & 0.987 \\ \cmidrule(lr){2-2} \cmidrule(l){4-9} 
                             & \textbf{Voiced}     &                        & 0.448 & 18.306 & 0.931 & 0.9   & 0.957 & 0.987 \\ \cmidrule(lr){2-2} \cmidrule(l){4-9} 
                             & \textbf{Sibilants}  &                        & 0.833 & 7.191  & 0.973 & 0.959 & 0.979 & 0.99  \\ \cmidrule(lr){2-2} \cmidrule(l){4-9} 
                             & \textbf{Overall}    &                        & 0.318 & 18.682 & 0.931 & 0.898 & 0.957 & 0.969 \\ \midrule
\multirow{6}{*}{\textbf{ml}} & \textbf{Consonants} & \multirow{6}{*}{0.939} & 0.558 & 17.273 & 0.953 & 0.915 & 0.957 & 0.951 \\ \cmidrule(lr){2-2} \cmidrule(l){4-9} 
                             & \textbf{Vowels}     &                        & 0.517 & 19.872 & 0.944 & 0.902 & 0.961 & 0.935 \\ \cmidrule(lr){2-2} \cmidrule(l){4-9} 
                             & \textbf{Aspirated}  &                        & 0.554 & 16.946 & 0.953 & 0.916 & 0.956 & 0.953 \\ \cmidrule(lr){2-2} \cmidrule(l){4-9} 
                             & \textbf{Voiced}     &                        & 0.563 & 16.847 & 0.953 & 0.917 & 0.959 & 0.952 \\ \cmidrule(lr){2-2} \cmidrule(l){4-9} 
                             & \textbf{Sibilants}  &                        & 0.864 & 5.41   & 0.979 & 0.966 & 0.981 & 0.957 \\ \cmidrule(lr){2-2} \cmidrule(l){4-9} 
                             & \textbf{Overall}    &                        & 0.348 & 18.991 & 0.945 & 0.905 & 0.957 & 0.943 \\ \midrule
\multirow{6}{*}{\textbf{mr}} & \textbf{Consonants} & \multirow{6}{*}{0.947} & 0.401 & 17.663 & 0.922 & 0.879 & 0.954 & 0.983 \\ \cmidrule(lr){2-2} \cmidrule(l){4-9} 
                             & \textbf{Vowels}     &                        & 0.496 & 18.745 & 0.937 & 0.874 & 0.961 & 0.928 \\ \cmidrule(lr){2-2} \cmidrule(l){4-9} 
                             & \textbf{Aspirated}  &                        & 0.399 & 16.647 & 0.929 & 0.886 & 0.956 & 0.985 \\ \cmidrule(lr){2-2} \cmidrule(l){4-9} 
                             & \textbf{Voiced}     &                        & 0.408 & 17.841 & 0.919 & 0.877 & 0.955 & 0.985 \\ \cmidrule(lr){2-2} \cmidrule(l){4-9} 
                             & \textbf{Sibilants}  &                        & 0.847 & 7.569  & 0.969 & 0.944 & 0.975 & 0.988 \\ \cmidrule(lr){2-2} \cmidrule(l){4-9} 
                             & \textbf{Overall}    &                        & 0.325 & 17.914 & 0.93  & 0.878 & 0.957 & 0.954 \\ \midrule
\multirow{6}{*}{\textbf{or}} & \textbf{Consonants} & \multirow{6}{*}{0.933} & 0.371 & 17.524 & 0.916 & 0.876 & 0.959 & 0.96  \\ \cmidrule(lr){2-2} \cmidrule(l){4-9} 
                             & \textbf{Vowels}     &                        & 0.461 & 17.837 & 0.926 & 0.874 & 0.962 & 0.932 \\ \cmidrule(lr){2-2} \cmidrule(l){4-9} 
                             & \textbf{Aspirated}  &                        & 0.373 & 17.797 & 0.916 & 0.875 & 0.958 & 0.962 \\ \cmidrule(lr){2-2} \cmidrule(l){4-9} 
                             & \textbf{Voiced}     &                        & 0.38  & 17.296 & 0.915 & 0.877 & 0.961 & 0.962 \\ \cmidrule(lr){2-2} \cmidrule(l){4-9} 
                             & \textbf{Sibilants}  &                        & 0.807 & 7.021  & 0.966 & 0.948 & 0.981 & 0.965 \\ \cmidrule(lr){2-2} \cmidrule(l){4-9} 
                             & \textbf{Overall}    &                        & 0.259 & 17.214 & 0.92  & 0.877 & 0.961 & 0.945 \\ \midrule
\multirow{6}{*}{\textbf{pa}} & \textbf{Consonants} & \multirow{6}{*}{0.951} & 0.411 & 14.996 & 0.912 & 0.859 & 0.954 & 0.934 \\ \cmidrule(lr){2-2} \cmidrule(l){4-9} 
                             & \textbf{Vowels}     &                        & 0.469 & 17.443 & 0.917 & 0.846 & 0.955 & 0.918 \\ \cmidrule(lr){2-2} \cmidrule(l){4-9} 
                             & \textbf{Aspirated}  &                        & 0.437 & 14.987 & 0.912 & 0.857 & 0.953 & 0.935 \\ \cmidrule(lr){2-2} \cmidrule(l){4-9} 
                             & \textbf{Voiced}     &                        & 0.418 & 14.604 & 0.915 & 0.864 & 0.956 & 0.936 \\ \cmidrule(lr){2-2} \cmidrule(l){4-9} 
                             & \textbf{Sibilants}  &                        & 0.857 & 4.471  & 0.966 & 0.951 & 0.978 & 0.937 \\ \cmidrule(lr){2-2} \cmidrule(l){4-9} 
                             & \textbf{Overall}    &                        & 0.265 & 16.789 & 0.911 & 0.846 & 0.952 & 0.923 \\ \midrule
\multirow{6}{*}{\textbf{ta}} & \textbf{Consonants} & \multirow{6}{*}{0.948} & 0.922 & 2.407  & 0.989 & 0.984 & 0.992 & 0.994 \\ \cmidrule(lr){2-2} \cmidrule(l){4-9} 
                             & \textbf{Vowels}     &                        & 0.346 & 19.741 & 0.932 & 0.9   & 0.958 & 0.955 \\ \cmidrule(lr){2-2} \cmidrule(l){4-9} 
                             & \textbf{Aspirated}  &                        & 0.946 & 0.211  & 0.999 & 0.998 & 0.999 & 0.997 \\ \cmidrule(lr){2-2} \cmidrule(l){4-9} 
                             & \textbf{Voiced}     &                        & 0.859 & 5.834  & 0.974 & 0.967 & 0.984 & 0.992 \\ \cmidrule(lr){2-2} \cmidrule(l){4-9} 
                             & \textbf{Sibilants}  &                        & 0.945 & 0.3    & 0.999 & 0.998 & 0.999 & 0.996 \\ \cmidrule(lr){2-2} \cmidrule(l){4-9} 
                             & \textbf{Overall}    &                        & 0.334 & 19.847 & 0.932 & 0.899 & 0.957 & 0.956 \\ \midrule
\multirow{6}{*}{\textbf{te}} & \textbf{Consonants} & \multirow{6}{*}{0.948} & 0.419 & 18.056 & 0.927 & 0.889 & 0.955 & 0.978 \\ \cmidrule(lr){2-2} \cmidrule(l){4-9} 
                             & \textbf{Vowels}     &                        & 0.452 & 19.467 & 0.938 & 0.887 & 0.961 & 0.936 \\ \cmidrule(lr){2-2} \cmidrule(l){4-9} 
                             & \textbf{Aspirated}  &                        & 0.419 & 17.705 & 0.929 & 0.891 & 0.955 & 0.981 \\ \cmidrule(lr){2-2} \cmidrule(l){4-9} 
                             & \textbf{Voiced}     &                        & 0.433 & 17.529 & 0.928 & 0.896 & 0.96  & 0.98  \\ \cmidrule(lr){2-2} \cmidrule(l){4-9} 
                             & \textbf{Sibilants}  &                        & 0.872 & 5.498  & 0.982 & 0.962 & 0.982 & 0.982 \\ \cmidrule(lr){2-2} \cmidrule(l){4-9} 
                             & \textbf{Overall}    &                        & 0.279 & 18.293 & 0.932 & 0.89  & 0.958 & 0.956 \\ \bottomrule
\end{tabular}%
}
\caption{Fine-grained Analysis of Phonological Attacks}
\label{tab:fg_phono_attacks}
\end{table*}

\begin{table*}[!h]
\centering
  \includegraphics[width=\linewidth]{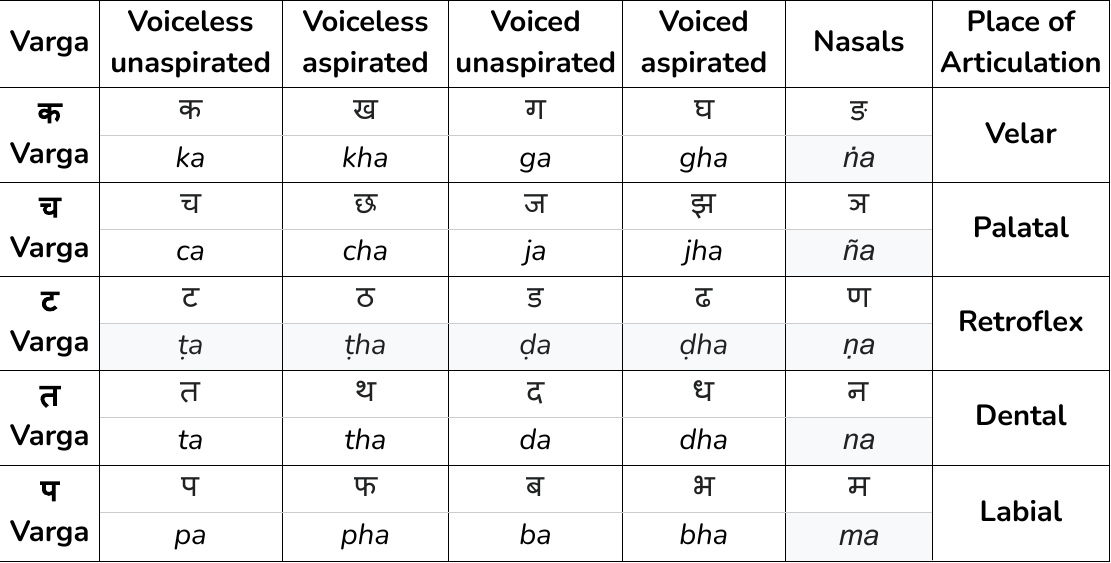}
\caption{Consonants in Devanagari script}
\label{fig:consonants_deva}
\end{table*}

\begin{table*}[!h]
\centering
  \includegraphics[width=0.5\linewidth]{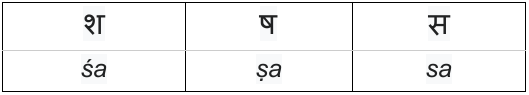}
\caption{Sibilants in Devanagari script}
\label{fig:sibilants_deva}
\end{table*}

\begin{table*}[!h]
\centering
  \includegraphics[width=\linewidth]{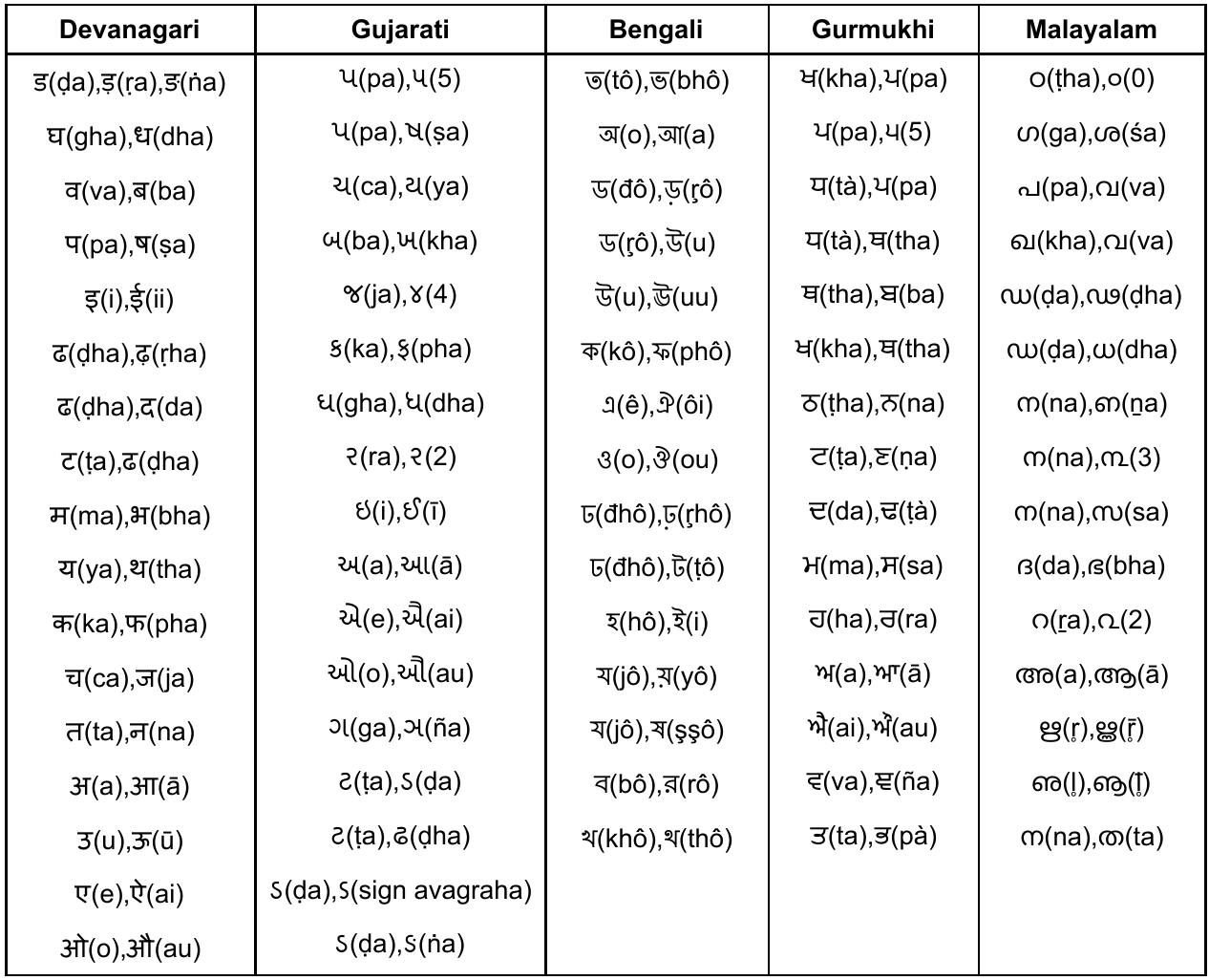}
\caption{Visually similar characters across different scripts}
\label{fig:Visually_similar_character_1}
\end{table*}

\begin{table*}[!h]
\centering
  \includegraphics[width=0.8\linewidth]{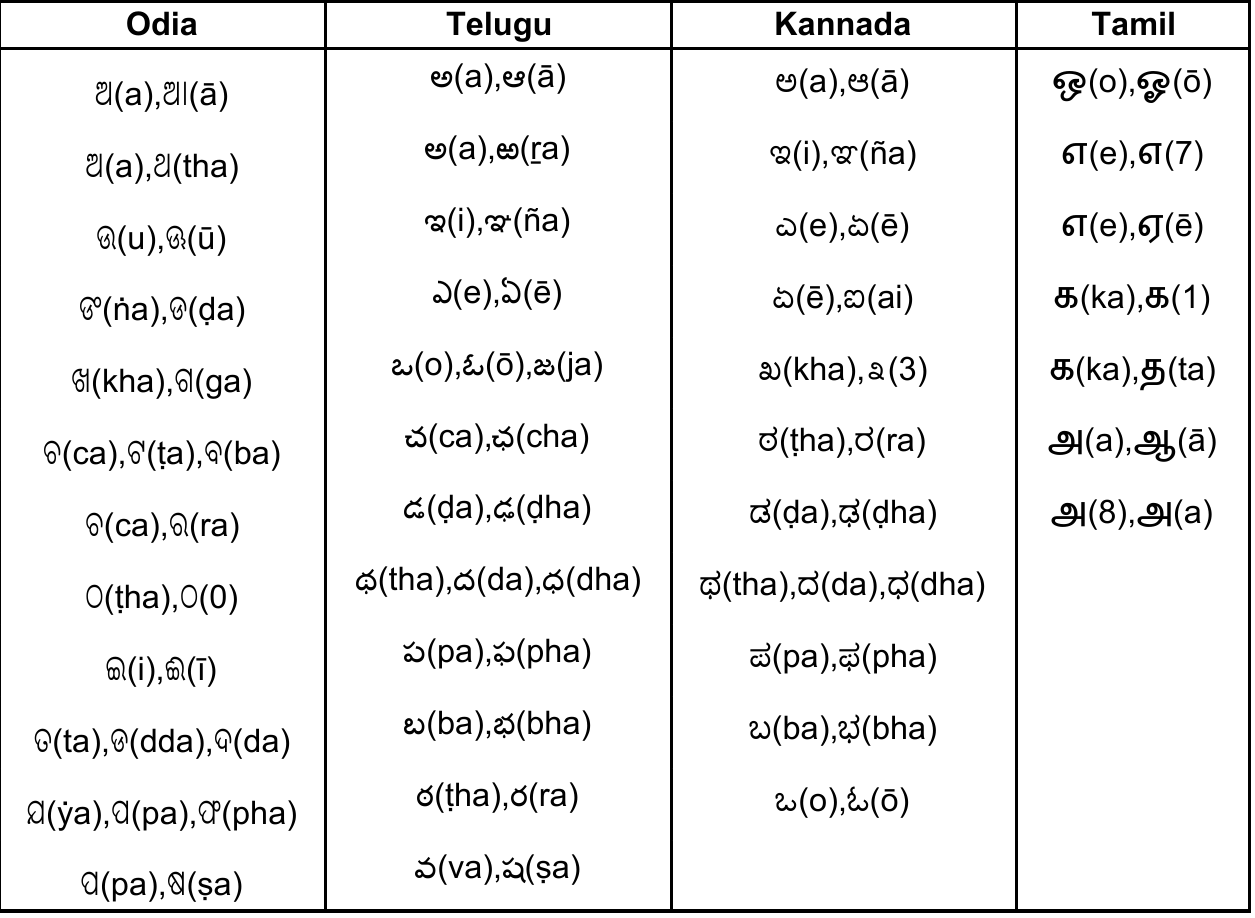}
\caption{Visually similar characters across different scripts}
\label{fig:Visually_similar_character_2}
\end{table*}

\section{Detailed Results}
We evaluate several widely used Indic language models on various NLP tasks, including sentiment analysis, paraphrasing, and natural language inference, to assess their robustness against linguistically grounded perturbations. The detailed results are presented as follows: Tables \ref{tab:senti_indicbert}-\ref{tab:xnli_indicbert} for IndicBERTv2, Tables \ref{tab:senti_muril}-\ref{tab:xnli_muril} for MuRIL, Tables \ref{tab:senti_xlmr}-\ref{tab:xnli_xlmr} for XLM-R, and Tables \ref{tab:senti_mbert}-\ref{tab:xnli_mbert} for mBERT. In each table, the \textit{Lang} column lists the ISO codes for the different languages, as specified in Table \ref{tab:lang}.

Moreover, Odia script is not handled by mBERT tokenizer, resulting in <UNK> tokens upon tokenization. Hence results for Odia are not present for mBERT.

\begin{table*}[]
\centering
\resizebox{\textwidth}{!}{%
\begin{tabular}{@{}ccccccccccccc@{}}
\toprule
\textbf{Lang} &
  \textbf{\begin{tabular}[c]{@{}c@{}}Type\\ of\\ Perturbation\end{tabular}} &
  \textbf{\begin{tabular}[c]{@{}c@{}}Original\\ Accuracy\end{tabular}} &
  \textbf{\begin{tabular}[c]{@{}c@{}}After\\ Attack\\ Accuracy\end{tabular}} &
  \textbf{\begin{tabular}[c]{@{}c@{}}\% \\ Perturbed\\ Words\end{tabular}} &
  \textbf{\begin{tabular}[c]{@{}c@{}}Semantic\\ Similarity\end{tabular}} &
  \textbf{\begin{tabular}[c]{@{}c@{}}Overlap\\ Similarity\end{tabular}} &
  \textbf{\begin{tabular}[c]{@{}c@{}}BERTScore\\ based\\ Similarity\end{tabular}} &
  \textbf{\begin{tabular}[c]{@{}c@{}}Phonetic\\ Similarity\end{tabular}} &
  \textbf{\begin{tabular}[c]{@{}c@{}}Avg. No. of \\ Candidates \\ per word\end{tabular}} &
  \textbf{\begin{tabular}[c]{@{}c@{}}Avg. \\ Word \\ Length\end{tabular}} &
  \textbf{\begin{tabular}[c]{@{}c@{}}Query \\ Number\end{tabular}} &
  \textbf{\begin{tabular}[c]{@{}c@{}}Avg. \\ Sentence \\ Length\end{tabular}} \\ \midrule
\multirow{3}{*}{\textbf{as}} & \textbf{Rand}  & \multirow{3}{*}{0.931} & 0.171 & 16.824 & 0.929 & 0.876 & 0.96  & -     & 4.273 & 5.34  & 41.705 & \multirow{3}{*}{24.003} \\ \cmidrule(lr){2-2} \cmidrule(lr){4-12}
                             & \textbf{Phono} &                        & 0.23  & 18.11  & 0.932 & 0.867 & 0.96  & 0.882 & 3.365 & 5.331 & 40.287 &                         \\ \cmidrule(lr){2-2} \cmidrule(lr){4-12}
                             & \textbf{Ortho} &                        & 0.449 & 17.632 & 0.927 & 0.861 & 0.958 & -     & 1.198 & 5.288 & 29.244 &                         \\ \midrule
\multirow{3}{*}{\textbf{bd}} & \textbf{Rand}  & \multirow{3}{*}{0.859} & 0.109 & 14.507 & 0.968 & 0.904 & 0.975 & -     & 5.423 & 5.841 & 35.68  & \multirow{3}{*}{22.205} \\ \cmidrule(lr){2-2} \cmidrule(lr){4-12}
                             & \textbf{Phono} &                        & 0.138 & 14.866 & 0.972 & 0.902 & 0.977 & -     & 3.865 & 5.84  & 31.643 &                         \\ \cmidrule(lr){2-2} \cmidrule(lr){4-12}
                             & \textbf{Ortho} &                        & 0.301 & 15.371 & 0.97  & 0.898 & 0.975 & -     & 1.05  & 5.838 & 23.671 &                         \\ \midrule
\multirow{3}{*}{\textbf{bn}} & \textbf{Rand}  & \multirow{3}{*}{0.955} & 0.185 & 16.665 & 0.915 & 0.874 & 0.952 & -     & 4.764 & 5.255 & 43.601 & \multirow{3}{*}{23.721} \\ \cmidrule(lr){2-2} \cmidrule(lr){4-12}
                             & \textbf{Phono} &                        & 0.256 & 17.759 & 0.919 & 0.867 & 0.953 & 0.947 & 3.445 & 5.229 & 40.995 &                         \\ \cmidrule(lr){2-2} \cmidrule(lr){4-12}
                             & \textbf{Ortho} &                        & 0.437 & 18.834 & 0.912 & 0.852 & 0.949 & -     & 1.222 & 5.198 & 30.491 &                         \\ \midrule
\multirow{3}{*}{\textbf{gu}} & \textbf{Rand}  & \multirow{3}{*}{0.943} & 0.184 & 17.943 & 0.91  & 0.858 & 0.951 & -     & 4.287 & 4.798 & 49.145 & \multirow{3}{*}{26.129} \\ \cmidrule(lr){2-2} \cmidrule(lr){4-12}
                             & \textbf{Phono} &                        & 0.268 & 18.951 & 0.907 & 0.851 & 0.953 & 0.951 & 3.059 & 4.763 & 44.077 &                         \\ \cmidrule(lr){2-2} \cmidrule(lr){4-12}
                             & \textbf{Ortho} &                        & 0.381 & 17.893 & 0.905 & 0.849 & 0.949 & -     & 1.685 & 4.754 & 37.195 &                         \\ \midrule
\multirow{3}{*}{\textbf{hi}} & \textbf{Rand}  & \multirow{3}{*}{0.957} & 0.198 & 15.477 & 0.918 & 0.86  & 0.951 & -     & 3.75  & 4.2   & 50.787 & \multirow{3}{*}{30.107} \\ \cmidrule(lr){2-2} \cmidrule(lr){4-12}
                             & \textbf{Phono} &                        & 0.29  & 16.426 & 0.921 & 0.852 & 0.954 & 0.959 & 2.725 & 4.166 & 47.856 &                         \\ \cmidrule(lr){2-2} \cmidrule(lr){4-12}
                             & \textbf{Ortho} &                        & 0.343 & 15.186 & 0.918 & 0.854 & 0.952 & -     & 1.075 & 4.153 & 36.864 &                         \\ \midrule
\multirow{3}{*}{\textbf{kn}} & \textbf{Rand}  & \multirow{3}{*}{0.938} & 0.23  & 18.461 & 0.928 & 0.9   & 0.955 & -     & 6.131 & 7.718 & 42.375 & \multirow{3}{*}{20.034} \\ \cmidrule(lr){2-2} \cmidrule(lr){4-12}
                             & \textbf{Phono} &                        & 0.318 & 18.682 & 0.931 & 0.898 & 0.957 & 0.969 & 4.204 & 7.666 & 35.36  &                         \\ \cmidrule(lr){2-2} \cmidrule(lr){4-12}
                             & \textbf{Ortho} &                        & 0.623 & 16.104 & 0.941 & 0.905 & 0.959 & -     & 1.284 & 7.569 & 22.125 &                         \\ \midrule
\multirow{3}{*}{\textbf{ml}} & \textbf{Rand}  & \multirow{3}{*}{0.939} & 0.262 & 18.551 & 0.937 & 0.907 & 0.955 & -     & 6.347 & 8.836 & 39.805 & \multirow{3}{*}{19.254} \\ \cmidrule(lr){2-2} \cmidrule(lr){4-12}
                             & \textbf{Phono} &                        & 0.348 & 18.991 & 0.945 & 0.905 & 0.957 & 0.943 & 4.094 & 8.719 & 34.639 &                         \\ \cmidrule(lr){2-2} \cmidrule(lr){4-12}
                             & \textbf{Ortho} &                        & 0.595 & 17.606 & 0.942 & 0.907 & 0.957 & -     & 1.501 & 8.633 & 21.704 &                         \\ \midrule
\multirow{3}{*}{\textbf{mr}} & \textbf{Rand}  & \multirow{3}{*}{0.947} & 0.217 & 17.597 & 0.921 & 0.879 & 0.954 & -     & 4.952 & 5.591 & 43.821 & \multirow{3}{*}{23.172} \\ \cmidrule(lr){2-2} \cmidrule(lr){4-12}
                             & \textbf{Phono} &                        & 0.325 & 17.914 & 0.93  & 0.878 & 0.957 & 0.954 & 3.528 & 5.599 & 38.665 &                         \\ \cmidrule(lr){2-2} \cmidrule(lr){4-12}
                             & \textbf{Ortho} &                        & 0.392 & 18.737 & 0.919 & 0.865 & 0.953 & -     & 1.291 & 5.604 & 30.082 &                         \\ \midrule
\multirow{3}{*}{\textbf{or}} & \textbf{Rand}  & \multirow{3}{*}{0.932} & 0.2   & 16.876 & 0.918 & 0.88  & 0.959 & -     & 4.719 & 5.437 & 43.382 & \multirow{3}{*}{23.491} \\ \cmidrule(lr){2-2} \cmidrule(lr){4-12}
                             & \textbf{Phono} &                        & 0.259 & 17.214 & 0.92  & 0.877 & 0.961 & 0.945 & 3.447 & 5.449 & 38.83  &                         \\ \cmidrule(lr){2-2} \cmidrule(lr){4-12}
                             & \textbf{Ortho} &                        & 0.463 & 17.001 & 0.921 & 0.869 & 0.958 & -     & 1.174 & 5.435 & 29.598 &                         \\ \midrule
\multirow{3}{*}{\textbf{pa}} & \textbf{Rand}  & \multirow{3}{*}{0.95}  & 0.176 & 15.409 & 0.913 & 0.857 & 0.952 & -     & 3.714 & 4.171 & 49.891 & \multirow{3}{*}{30.062} \\ \cmidrule(lr){2-2} \cmidrule(lr){4-12}
                             & \textbf{Phono} &                        & 0.265 & 16.789 & 0.911 & 0.846 & 0.952 & 0.923 & 2.714 & 4.15  & 47.255 &                         \\ \cmidrule(lr){2-2} \cmidrule(lr){4-12}
                             & \textbf{Ortho} &                        & 0.422 & 15.509 & 0.914 & 0.855 & 0.951 & -     & 0.931 & 4.12  & 35.503 &                         \\ \midrule
\multirow{3}{*}{\textbf{ta}} & \textbf{Rand}  & \multirow{3}{*}{0.948} & 0.161 & 17.618 & 0.929 & 0.908 & 0.957 & -     & 5.882 & 8.049 & 41.332 & \multirow{3}{*}{20.874} \\ \cmidrule(lr){2-2} \cmidrule(lr){4-12}
                             & \textbf{Phono} &                        & 0.334 & 19.847 & 0.932 & 0.899 & 0.957 & 0.956 & 4.164 & 7.927 & 31.829 &                         \\ \cmidrule(lr){2-2} \cmidrule(lr){4-12}
                             & \textbf{Ortho} &                        & 0.622 & 17.019 & 0.944 & 0.906 & 0.958 & -     & 1.331 & 7.838 & 24.026 &                         \\ \midrule
\multirow{3}{*}{\textbf{te}} & \textbf{Rand}  & \multirow{3}{*}{0.948} & 0.234 & 17.895 & 0.928 & 0.892 & 0.955 & -     & 5.677 & 6.897 & 43.477 & \multirow{3}{*}{21.76}  \\ \cmidrule(lr){2-2} \cmidrule(lr){4-12}
                             & \textbf{Phono} &                        & 0.279 & 18.293 & 0.932 & 0.89  & 0.958 & 0.956 & 4.069 & 6.878 & 38.809 &                         \\ \cmidrule(lr){2-2} \cmidrule(lr){4-12}
                             & \textbf{Ortho} &                        & 0.517 & 18.418 & 0.92  & 0.878 & 0.952 & -     & 1.269 & 6.795 & 26.286 &                         \\ \bottomrule
\end{tabular}%
}
\caption{The table presents the impact of different character-substitution attack strategies: random (\textbf{Rand}), phonetic (\textbf{Phono}), and orthographic (\textbf{Ortho}) on the \textit{IndicBERTv2} language model for the \textit{IndicSentiment} dataset.} 
\label{tab:senti_indicbert}
\end{table*}

\begin{table*}[]
\centering
\resizebox{\textwidth}{!}{%
\begin{tabular}{@{}ccccccccccccc@{}}
\toprule
\textbf{Lang} &
  \textbf{\begin{tabular}[c]{@{}c@{}}Type\\ of\\ Perturbation\end{tabular}} &
  \textbf{\begin{tabular}[c]{@{}c@{}}Original\\ Accuracy\end{tabular}} &
  \textbf{\begin{tabular}[c]{@{}c@{}}After\\ Attack\\ Accuracy\end{tabular}} &
  \textbf{\begin{tabular}[c]{@{}c@{}}\% \\ Perturbed\\ Words\end{tabular}} &
  \textbf{\begin{tabular}[c]{@{}c@{}}Semantic\\ Similarity\end{tabular}} &
  \textbf{\begin{tabular}[c]{@{}c@{}}Overlap\\ Similarity\end{tabular}} &
  \textbf{\begin{tabular}[c]{@{}c@{}}BERTScore\\ based\\ Similarity\end{tabular}} &
  \textbf{\begin{tabular}[c]{@{}c@{}}Phonetic\\ Similarity\end{tabular}} &
  \textbf{\begin{tabular}[c]{@{}c@{}}Avg. No. of \\ Candidates \\ per word\end{tabular}} &
  \textbf{\begin{tabular}[c]{@{}c@{}}Avg. \\ Word \\ Length\end{tabular}} &
  \textbf{\begin{tabular}[c]{@{}c@{}}Query \\ Number\end{tabular}} &
  \textbf{\begin{tabular}[c]{@{}c@{}}Avg. \\ Sentence \\ Length\end{tabular}} \\ \midrule
\multirow{3}{*}{\textbf{as}} &
  \textbf{Rand} &
  \multirow{3}{*}{0.572} &
  0.103/0.094 &
  10.654/11.05 &
  0.955/0.953 &
  0.928/0.925 &
  0.976/0.975 &
  - &
  4.958/5.068 &
  6.212/6.348 &
  18.59/18.758 &
  \multirow{3}{*}{17.814/17.628} \\ \cmidrule(lr){2-2} \cmidrule(lr){4-12}
 &
  \textbf{Phono} &
   &
  0.135/0.122 &
  10.498/11.222 &
  0.958/0.956 &
  0.93/0.924 &
  0.977/0.976 &
  0.896/0.896 &
  3.836/3.936 &
  6.129/6.298 &
  16.462/17.024 &
   \\ \cmidrule(lr){2-2} \cmidrule(lr){4-12}
 &
  \textbf{Ortho} &
   &
  0.179/0.164 &
  10.18/11.022 &
  0.956/0.952 &
  0.927/0.92 &
  0.976/0.975 &
  - &
  2.614/2.637 &
  6.05/6.177 &
  14.177/14.504 &
   \\ \midrule
\multirow{3}{*}{\textbf{bn}} &
  \textbf{Rand} &
  \multirow{3}{*}{0.498} &
  0.069/0.064 &
  8.645/8.947 &
  0.96/0.957 &
  0.944/0.939 &
  0.976/0.975 &
  - &
  5.478/5.505 &
  6.2/6.224 &
  12.9/13.3 &
  \multirow{3}{*}{15.031/15.299} \\ \cmidrule(lr){2-2} \cmidrule(lr){4-12}
 &
  \textbf{Phono} &
   &
  0.09/0.084 &
  8.61/8.986 &
  0.963/0.96 &
  0.944/0.939 &
  0.978/0.977 &
  0.979/0.976 &
  3.9/3.918 &
  6.148/6.218 &
  11.209/11.629 &
   \\ \cmidrule(lr){2-2} \cmidrule(lr){4-12}
 &
  \textbf{Ortho} &
   &
  0.137/0.123 &
  8.592/9.104 &
  0.962/0.958 &
  0.942/0.935 &
  0.977/0.975 &
  - &
  2.448/2.471 &
  6.059/6.108 &
  9.622/10.153 &
   \\ \midrule
\multirow{3}{*}{\textbf{gu}} &
  \textbf{Rand} &
  \multirow{3}{*}{0.723} &
  0.168/0.095 &
  12.115/12.553 &
  0.947/0.94 &
  0.915/0.912 &
  0.97/0.968 &
  - &
  5.033/5.132 &
  5.656/5.768 &
  21.138/21.365 &
  \multirow{3}{*}{16.53/16.594} \\ \cmidrule(lr){2-2} \cmidrule(lr){4-12}
 &
  \textbf{Phono} &
   &
  0.212/0.142 &
  12.349/13.032 &
  0.95/0.946 &
  0.914/0.909 &
  0.972/0.97 &
  0.972/0.97 &
  3.422/3.489 &
  5.589/5.698 &
  18.28/18.843 &
   \\ \cmidrule(lr){2-2} \cmidrule(lr){4-12}
 &
  \textbf{Ortho} &
   &
  0.31/0.232 &
  11.706/12.081 &
  0.947/0.942 &
  0.914/0.91 &
  0.969/0.967 &
  - &
  1.851/1.881 &
  5.422/5.521 &
  15.07/15.546 &
   \\ \midrule
\multirow{3}{*}{\textbf{hi}} &
  \textbf{Rand} &
  \multirow{3}{*}{0.498} &
  0.05/0.04 &
  8.262/7.897 &
  0.956/0.954 &
  0.933/0.934 &
  0.974/0.973 &
  - &
  4.361/4.5 &
  4.926/5.091 &
  16.06/16.187 &
  \multirow{3}{*}{19.749/20.201} \\ \cmidrule(lr){2-2} \cmidrule(lr){4-12}
 &
  \textbf{Phono} &
   &
  0.083/0.067 &
  8.151/8.464 &
  0.96/0.955 &
  0.934/0.929 &
  0.976/0.974 &
  0.979/0.977 &
  3.007/3.056 &
  4.844/4.961 &
  14.11/14.553 &
   \\ \cmidrule(lr){2-2} \cmidrule(lr){4-12}
 &
  \textbf{Ortho} &
   &
  0.088/0.064 &
  8.29/8.623 &
  0.955/0.949 &
  0.929/0.924 &
  0.974/0.972 &
  - &
  2.179/2.257 &
  4.771/4.915 &
  13.058/13.55 &
   \\ \midrule
\multirow{3}{*}{\textbf{kn}} &
  \textbf{Rand} &
  \multirow{3}{*}{0.583} &
  0.156/0.145 &
  11.117/11.417 &
  0.959/0.958 &
  0.94/0.939 &
  0.974/0.974 &
  - &
  5.839/5.89 &
  7.331/7.367 &
  16.063/15.952 &
  \multirow{3}{*}{14.402/14.236} \\ \cmidrule(lr){2-2} \cmidrule(lr){4-12}
 &
  \textbf{Phono} &
   &
  0.218/0.199 &
  10.332/11.067 &
  0.968/0.967 &
  0.944/0.941 &
  0.978/0.977 &
  0.978/0.978 &
  3.995/4.018 &
  7.311/7.372 &
  12.818/13.086 &
   \\ \cmidrule(lr){2-2} \cmidrule(lr){4-12}
 &
  \textbf{Ortho} &
   &
  0.303/0.274 &
  9.045/10.381 &
  0.971/0.965 &
  0.947/0.941 &
  0.978/0.976 &
  - &
  2.157/2.176 &
  7.129/7.246 &
  9.208/10.045 &
   \\ \midrule
\multirow{3}{*}{\textbf{ml}} &
  \textbf{Rand} &
  \multirow{3}{*}{0.566} &
  0.178/0.144 &
  10.898/12.133 &
  0.964/0.962 &
  0.947/0.943 &
  0.976/0.974 &
  - &
  6.173/6.25 &
  8.522/8.597 &
  14.597/15.905 &
  \multirow{3}{*}{13.238/13.265} \\ \cmidrule(lr){2-2} \cmidrule(lr){4-12}
 &
  \textbf{Phono} &
   &
  0.224/0.21 &
  10.108/11.152 &
  0.973/0.969 &
  0.951/0.947 &
  0.979/0.977 &
  0.95/0.95 &
  4.204/4.25 &
  8.501/8.575 &
  11.502/12.217 &
   \\ \cmidrule(lr){2-2} \cmidrule(lr){4-12}
 &
  \textbf{Ortho} &
   &
  0.276/0.255 &
  9.021/10.446 &
  0.971/0.965 &
  0.954/0.948 &
  0.979/0.977 &
  - &
  3.146/3.17 &
  8.359/8.485 &
  9.102/9.974 &
   \\ \midrule
\multirow{3}{*}{\textbf{mr}} &
  \textbf{Rand} &
  \multirow{3}{*}{0.544} &
  0.11/0.086 &
  10.729/10.96 &
  0.954/0.953 &
  0.934/0.933 &
  0.973/0.973 &
  - &
  5.609/5.678 &
  6.366/6.452 &
  17.001/17.155 &
  \multirow{3}{*}{15.938/16.051} \\ \cmidrule(lr){2-2} \cmidrule(lr){4-12}
 &
  \textbf{Phono} &
   &
  0.176/0.136 &
  9.815/10.879 &
  0.963/0.958 &
  0.94/0.934 &
  0.977/0.975 &
  0.978/0.975 &
  3.872/3.881 &
  6.297/6.358 &
  13.305/14.318 &
   \\ \cmidrule(lr){2-2} \cmidrule(lr){4-12}
 &
  \textbf{Ortho} &
   &
  0.181/0.138 &
  9.789/11.136 &
  0.96/0.953 &
  0.938/0.929 &
  0.976/0.973 &
  - &
  2.945/2.981 &
  6.259/6.318 &
  12.1/13.3 &
   \\ \midrule
\multirow{3}{*}{\textbf{or}} &
  \textbf{Rand} &
  \multirow{3}{*}{0.576} &
  0.11/0.082 &
  10.309/10.719 &
  0.958/0.955 &
  0.936/0.932 &
  0.975/0.974 &
  - &
  5.369/5.388 &
  6.158/6.188 &
  16.542/17.029 &
  \multirow{3}{*}{15.884/16.035} \\ \cmidrule(lr){2-2} \cmidrule(lr){4-12}
 &
  \textbf{Phono} &
   &
  0.127/0.106 &
  10.379/11.099 &
  0.96/0.957 &
  0.935/0.931 &
  0.976/0.975 &
  0.962/0.961 &
  3.805/3.824 &
  6.141/6.168 &
  14.302/15.047 &
   \\ \cmidrule(lr){2-2} \cmidrule(lr){4-12}
 &
  \textbf{Ortho} &
   &
  0.176/0.145 &
  9.548/10.766 &
  0.958/0.952 &
  0.936/0.927 &
  0.976/0.973 &
  - &
  2.479/2.463 &
  6.041/6.046 &
  11.995/13.025 &
   \\ \midrule
\multirow{3}{*}{\textbf{pa}} &
  \textbf{Rand} &
  \multirow{3}{*}{0.543} &
  0.067/0.047 &
  9.158/9.112 &
  0.945/0.941 &
  0.923/0.922 &
  0.973/0.973 &
  - &
  4.227/4.295 &
  4.738/4.808 &
  17.961/18.416 &
  \multirow{3}{*}{20.124/20.372} \\ \cmidrule(lr){2-2} \cmidrule(lr){4-12}
 &
  \textbf{Phono} &
   &
  0.102/0.079 &
  9.046/9.452 &
  0.947/0.945 &
  0.924/0.92 &
  0.975/0.974 &
  0.938/0.935 &
  3.023/3.058 &
  4.677/4.719 &
  15.874/16.629 &
   \\ \cmidrule(lr){2-2} \cmidrule(lr){4-12}
 &
  \textbf{Ortho} &
   &
  0.115/0.081 &
  9.161/9.528 &
  0.943/0.941 &
  0.923/0.918 &
  0.973/0.972 &
  - &
  2.076/2.087 &
  4.625/4.67 &
  14.479/15.139 &
   \\ \midrule
\multirow{3}{*}{\textbf{te}} &
  \textbf{Rand} &
  \multirow{3}{*}{0.545} &
  0.127/0.098 &
  9.86/10.887 &
  0.963/0.957 &
  0.946/0.939 &
  0.977/0.975 &
  - &
  5.837/5.825 &
  7.178/7.149 &
  13.529/14.458 &
  \multirow{3}{*}{13.726/13.731} \\ \cmidrule(lr){2-2} \cmidrule(lr){4-12}
 &
  \textbf{Phono} &
   &
  0.166/0.139 &
  9.324/10.861 &
  0.969/0.963 &
  0.949/0.94 &
  0.98/0.978 &
  0.98/0.978 &
  4.007/4.006 &
  7.128/7.109 &
  11.026/12.046 &
   \\ \cmidrule(lr){2-2} \cmidrule(lr){4-12}
 &
  \textbf{Ortho} &
   &
  0.211/0.199 &
  9.151/10.116 &
  0.968/0.964 &
  0.946/0.94 &
  0.978/0.977 &
  - &
  2.294/2.274 &
  7.101/7.052 &
  9.012/9.407 &
   \\ \bottomrule
\end{tabular}%
}
\caption{The table presents the impact of different character-substitution attack strategies: random (\textbf{Rand}), phonetic (\textbf{Phono}), and orthographic (\textbf{Ortho}) on the \textit{IndicBERTv2} language model for the \textit{IndicParaphrase} dataset. Results are shown for perturbing both \textit{sentence1} and \textit{sentence2} of the IndicParaphrase dataset, separated by a forward slash (/).}
\label{tab:para_indicbert}
\end{table*}

\begin{table*}[]
\centering
\resizebox{\textwidth}{!}{%
\begin{tabular}{@{}ccccccccccccc@{}}
\toprule
\textbf{Lang} &
  \textbf{\begin{tabular}[c]{@{}c@{}}Type\\ of\\ Perturbation\end{tabular}} &
  \textbf{\begin{tabular}[c]{@{}c@{}}Original\\ Accuracy\end{tabular}} &
  \textbf{\begin{tabular}[c]{@{}c@{}}After\\ Attack\\ Accuracy\end{tabular}} &
  \textbf{\begin{tabular}[c]{@{}c@{}}\%\\ Perturbed\\ Words\end{tabular}} &
  \textbf{\begin{tabular}[c]{@{}c@{}}Semantic\\ Similarity\end{tabular}} &
  \textbf{\begin{tabular}[c]{@{}c@{}}Overlap\\ Similarity\end{tabular}} &
  \textbf{\begin{tabular}[c]{@{}c@{}}BERTScore\\ based\\ Similarity\end{tabular}} &
  \textbf{\begin{tabular}[c]{@{}c@{}}Phonetic\\ Similarity\end{tabular}} &
  \textbf{\begin{tabular}[c]{@{}c@{}}Avg. No. of \\ Candidates \\ per word\end{tabular}} &
  \textbf{\begin{tabular}[c]{@{}c@{}}Avg. \\ Word \\ Length\end{tabular}} &
  \textbf{\begin{tabular}[c]{@{}c@{}}Query \\ Number\end{tabular}} &
  \textbf{\begin{tabular}[c]{@{}c@{}}Avg. \\ Sentence \\ Length\end{tabular}} \\ \midrule
\multirow{3}{*}{\textbf{as}} &
  \textbf{Rand} &
  \multirow{3}{*}{0.676} &
  0.206/0.077 &
  8.405/11.86 &
  0.952/0.926 &
  0.926/0.901 &
  0.975/0.97 &
  - &
  4.105/4.348 &
  5.156/5.348 &
  17.001/9.922 &
  \multirow{3}{*}{17.091/9.459} \\ \cmidrule(lr){2-2} \cmidrule(lr){4-12}
 &
  \textbf{Phono} &
   &
  0.232/0.126 &
  8.76/12.403 &
  0.954/0.933 &
  0.924/0.899 &
  0.976/0.972 &
  0.89/0.892 &
  3.267/3.372 &
  5.135/5.321 &
  15.911/9.49 &
   \\ \cmidrule(lr){2-2} \cmidrule(lr){4-12}
 &
  \textbf{Ortho} &
   &
  0.293/0.209 &
  7.833/12.17 &
  0.955/0.934 &
  0.929/0.899 &
  0.976/0.97 &
  - &
  2.177/2.133 &
  5.111/5.222 &
  13.542/8.402 &
   \\ \midrule
\multirow{3}{*}{\textbf{bn}} &
  \textbf{Rand} &
  \multirow{3}{*}{0.711} &
  0.206/0.085 &
  9.256/13.099 &
  0.941/0.912 &
  0.917/0.888 &
  0.967/0.961 &
  - &
  4.487/4.507 &
  4.962/4.959 &
  18.448/10.476 &
  \multirow{3}{*}{16.971/9.221} \\ \cmidrule(lr){2-2} \cmidrule(lr){4-12}
 &
  \textbf{Phono} &
   &
  0.248/0.142 &
  9.063/13.611 &
  0.946/0.92 &
  0.918/0.886 &
  0.97/0.963 &
  0.975/0.978 &
  3.245/3.266 &
  4.927/4.964 &
  16.177/9.781 &
   \\ \cmidrule(lr){2-2} \cmidrule(lr){4-12}
 &
  \textbf{Ortho} &
   &
  0.302/0.215 &
  8.984/13.149 &
  0.944/0.919 &
  0.918/0.888 &
  0.969/0.964 &
  - &
  2.061/1.973 &
  4.888/4.894 &
  14.464/8.686 &
   \\ \midrule
\multirow{3}{*}{\textbf{gu}} &
  \textbf{Rand} &
  \multirow{3}{*}{0.735} &
  0.197/0.087 &
  9.304/12.735 &
  0.938/0.913 &
  0.91/0.881 &
  0.967/0.963 &
  - &
  4.037/4.063 &
  4.503/4.56 &
  20.014/11.184 &
  \multirow{3}{*}{18.389/9.956} \\ \cmidrule(lr){2-2} \cmidrule(lr){4-12}
 &
  \textbf{Phono} &
   &
  0.25/0.15 &
  9.305/12.931 &
  0.942/0.917 &
  0.912/0.881 &
  0.97/0.966 &
  0.973/0.979 &
  2.841/2.849 &
  4.483/4.529 &
  18.242/10.419 &
   \\ \cmidrule(lr){2-2} \cmidrule(lr){4-12}
 &
  \textbf{Ortho} &
   &
  0.334/0.243 &
  7.66/11.917 &
  0.947/0.92 &
  0.921/0.885 &
  0.97/0.964 &
  - &
  1.423/1.428 &
  4.419/4.472 &
  15.209/9.07 &
   \\ \midrule
\multirow{3}{*}{\textbf{hi}} &
  \textbf{Rand} &
  \multirow{3}{*}{0.718} &
  0.185/0.067 &
  7.94/10.86 &
  0.94/0.913 &
  0.91/0.884 &
  0.967/0.962 &
  - &
  3.457/3.577 &
  3.85/4.041 &
  21.273/11.616 &
  \multirow{3}{*}{21.825/11.492} \\ \cmidrule(lr){2-2} \cmidrule(lr){4-12}
 &
  \textbf{Phono} &
   &
  0.251/0.161 &
  7.94/11.27 &
  0.946/0.921 &
  0.913/0.884 &
  0.97/0.966 &
  0.979/0.981 &
  2.448/2.538 &
  3.799/3.961 &
  19.727/11.154 &
   \\ \cmidrule(lr){2-2} \cmidrule(lr){4-12}
 &
  \textbf{Ortho} &
   &
  0.254/0.131 &
  7.222/10.673 &
  0.943/0.916 &
  0.916/0.884 &
  0.97/0.963 &
  - &
  1.667/1.716 &
  3.793/3.985 &
  17.706/10.253 &
   \\ \midrule
\multirow{3}{*}{\textbf{kn}} &
  \textbf{Rand} &
  \multirow{3}{*}{0.717} &
  0.211/0.121 &
  10.374/15.709 &
  0.946/0.924 &
  0.932/0.908 &
  0.971/0.965 &
  - &
  5.915/6.541 &
  7.423/8.195 &
  17.721/10.649 &
  \multirow{3}{*}{14.072/7.533} \\ \cmidrule(lr){2-2} \cmidrule(lr){4-12}
 &
  \textbf{Phono} &
   &
  0.281/0.219 &
  10.138/15.101 &
  0.956/0.938 &
  0.933/0.911 &
  0.973/0.969 &
  0.983/0.985 &
  3.975/4.312 &
  7.347/8.011 &
  15.059/8.997 &
   \\ \cmidrule(lr){2-2} \cmidrule(lr){4-12}
 &
  \textbf{Ortho} &
   &
  0.403/0.362 &
  8.054/13.431 &
  0.966/0.943 &
  0.945/0.917 &
  0.976/0.97 &
  - &
  2.018/2.131 &
  7.204/7.708 &
  10.709/6.576 &
   \\ \midrule
\multirow{3}{*}{\textbf{ml}} &
  \textbf{Rand} &
  \multirow{3}{*}{0.727} &
  0.257/0.163 &
  10.054/16.153 &
  0.949/0.931 &
  0.935/0.916 &
  0.968/0.964 &
  - &
  6.119/6.923 &
  8.61/9.555 &
  15.657/9.979 &
  \multirow{3}{*}{12.98/6.837} \\ \cmidrule(lr){2-2} \cmidrule(lr){4-12}
 &
  \textbf{Phono} &
   &
  0.318/0.259 &
  9.683/16.019 &
  0.96/0.939 &
  0.939/0.917 &
  0.972/0.967 &
  0.953/0.951 &
  4.035/4.319 &
  8.569/9.319 &
  13.276/8.111 &
   \\ \cmidrule(lr){2-2} \cmidrule(lr){4-12}
 &
  \textbf{Ortho} &
   &
  0.35/0.298 &
  9.248/15.143 &
  0.954/0.934 &
  0.94/0.919 &
  0.971/0.966 &
  - &
  3.117/3.29 &
  8.54/9.205 &
  11.47/6.985 &
   \\ \midrule
\multirow{3}{*}{\textbf{mr}} &
  \textbf{Rand} &
  \multirow{3}{*}{0.696} &
  0.222/0.091 &
  8.799/13.349 &
  0.942/0.918 &
  0.923/0.894 &
  0.97/0.964 &
  - &
  4.745/4.822 &
  5.349/5.381 &
  16.502/9.981 &
  \multirow{3}{*}{16.154/8.773} \\ \cmidrule(lr){2-2} \cmidrule(lr){4-12}
 &
  \textbf{Phono} &
   &
  0.277/0.204 &
  8.956/13.707 &
  0.949/0.927 &
  0.924/0.896 &
  0.972/0.968 &
  0.978/0.979 &
  3.252/3.301 &
  5.295/5.343 &
  14.771/9.19 &
   \\ \cmidrule(lr){2-2} \cmidrule(lr){4-12}
 &
  \textbf{Ortho} &
   &
  0.272/0.136 &
  8.362/13.669 &
  0.946/0.916 &
  0.925/0.893 &
  0.972/0.965 &
  - &
  2.501/2.465 &
  5.309/5.343 &
  13.638/8.572 &
   \\ \midrule
\multirow{3}{*}{\textbf{or}} &
  \textbf{Rand} &
  \multirow{3}{*}{0.703} &
  0.22/0.107 &
  8.577/13.536 &
  0.945/0.922 &
  0.922/0.896 &
  0.973/0.966 &
  - &
  4.8/4.917 &
  5.409/5.54 &
  16.451/10.388 &
  \multirow{3}{*}{16.173/8.847} \\ \cmidrule(lr){2-2} \cmidrule(lr){4-12}
 &
  \textbf{Phono} &
   &
  0.255/0.157 &
  8.684/13.831 &
  0.948/0.926 &
  0.922/0.894 &
  0.974/0.967 &
  0.968/0.968 &
  3.527/3.575 &
  5.395/5.511 &
  15.066/9.571 &
   \\ \cmidrule(lr){2-2} \cmidrule(lr){4-12}
 &
  \textbf{Ortho} &
   &
  0.333/0.269 &
  7.779/12.801 &
  0.951/0.928 &
  0.928/0.897 &
  0.975/0.969 &
  - &
  1.979/1.943 &
  5.323/5.447 &
  12.673/8.083 &
   \\ \midrule
\multirow{3}{*}{\textbf{pa}} &
  \textbf{Rand} &
  \multirow{3}{*}{0.738} &
  0.203/0.071 &
  8.064/11.236 &
  0.94/0.914 &
  0.91/0.88 &
  0.968/0.964 &
  - &
  3.454/3.486 &
  3.896/3.968 &
  22.214/12.334 &
  \multirow{3}{*}{21.909/11.725} \\ \cmidrule(lr){2-2} \cmidrule(lr){4-12}
 &
  \textbf{Phono} &
   &
  0.25/0.168 &
  8.48/11.781 &
  0.941/0.915 &
  0.909/0.877 &
  0.97/0.966 &
  0.942/0.939 &
  2.549/2.515 &
  3.872/3.913 &
  21.027/11.779 &
   \\ \cmidrule(lr){2-2} \cmidrule(lr){4-12}
 &
  \textbf{Ortho} &
   &
  0.267/0.131 &
  7.926/11.599 &
  0.941/0.919 &
  0.911/0.878 &
  0.968/0.963 &
  - &
  1.649/1.662 &
  3.871/3.918 &
  18.867/11.001 &
   \\ \midrule
\multirow{3}{*}{\textbf{ta}} &
  \textbf{Rand} &
  \multirow{3}{*}{0.717} &
  0.223/0.109 &
  9.394/15.238 &
  0.944/0.92 &
  0.939/0.914 &
  0.971/0.965 &
  - &
  5.412/6.039 &
  7.785/8.377 &
  16.282/10.456 &
  \multirow{3}{*}{14.721/7.826} \\ \cmidrule(lr){2-2} \cmidrule(lr){4-12}
 &
  \textbf{Phono} &
   &
  0.335/0.269 &
  9.196/14.51 &
  0.953/0.933 &
  0.941/0.919 &
  0.974/0.97 &
  0.974/0.977 &
  3.764/4.063 &
  7.66/8.132 &
  12.464/7.424 &
   \\ \cmidrule(lr){2-2} \cmidrule(lr){4-12}
 &
  \textbf{Ortho} &
   &
  0.372/0.34 &
  8.23/13.674 &
  0.958/0.94 &
  0.943/0.921 &
  0.973/0.969 &
  - &
  2.105/2.245 &
  7.591/8.016 &
  11.523/7.03 &
   \\ \midrule
\multirow{3}{*}{\textbf{te}} &
  \textbf{Rand} &
  \multirow{3}{*}{0.713} &
  0.223/0.115 &
  9.445/14.841 &
  0.949/0.927 &
  0.929/0.902 &
  0.971/0.965 &
  - &
  5.253/5.818 &
  6.398/7.004 &
  16.326/9.937 &
  \multirow{3}{*}{14.655/7.706} \\ \cmidrule(lr){2-2} \cmidrule(lr){4-12}
 &
  \textbf{Phono} &
   &
  0.247/0.159 &
  9.681/15.007 &
  0.953/0.931 &
  0.928/0.902 &
  0.973/0.968 &
  0.974/0.98 &
  3.675/4.098 &
  6.378/6.91 &
  14.582/9.056 &
   \\ \cmidrule(lr){2-2} \cmidrule(lr){4-12}
 &
  \textbf{Ortho} &
   &
  0.337/0.264 &
  8.149/14.028 &
  0.956/0.93 &
  0.936/0.905 &
  0.974/0.968 &
  - &
  1.94/2.112 &
  6.312/6.772 &
  11.557/7.187 &
   \\ \bottomrule
\end{tabular}%
}
\caption{The table presents the impact of different character-substitution attack strategies: random (\textbf{Rand}), phonetic (\textbf{Phono}), and orthographic (\textbf{Ortho}) on the \textit{IndicBERTv2} language model for the \textit{IndicXNLI} dataset. Results are shown for perturbing both \textit{premise} and \textit{hypothesis} of the IndicXNLI dataset, separated by a forward slash (/).}
\label{tab:xnli_indicbert}
\end{table*}

\begin{table*}[]
\centering
\resizebox{\textwidth}{!}{%
\begin{tabular}{@{}ccccccccccccc@{}}
\toprule
\textbf{Lang} &
  \textbf{\begin{tabular}[c]{@{}c@{}}Type\\ of\\ Perturbation\end{tabular}} &
  \textbf{\begin{tabular}[c]{@{}c@{}}Original\\ Accuracy\end{tabular}} &
  \textbf{\begin{tabular}[c]{@{}c@{}}After\\ Attack\\ Accuracy\end{tabular}} &
  \textbf{\begin{tabular}[c]{@{}c@{}}\%\\ Perturbed\\ Words\end{tabular}} &
  \textbf{\begin{tabular}[c]{@{}c@{}}Semantic\\ Similarity\end{tabular}} &
  \textbf{\begin{tabular}[c]{@{}c@{}}Overlap\\ Similarity\end{tabular}} &
  \textbf{\begin{tabular}[c]{@{}c@{}}BERTScore\\ based\\ Similarity\end{tabular}} &
  \textbf{\begin{tabular}[c]{@{}c@{}}Phonetic\\ Similarity\end{tabular}} &
  \textbf{\begin{tabular}[c]{@{}c@{}}Avg. No. of\\ Candidates\\ per word\end{tabular}} &
  \textbf{\begin{tabular}[c]{@{}c@{}}Avg. \\ Word\\ Length\end{tabular}} &
  \textbf{\begin{tabular}[c]{@{}c@{}}Query\\ Number\end{tabular}} &
  \textbf{\begin{tabular}[c]{@{}c@{}}Avg.\\ Sentence\\ Length\end{tabular}} \\ \midrule
\multirow{3}{*}{\textbf{as}} &
  \textbf{Random} &
  \multirow{3}{*}{0.885} &
  0.086 &
  11.665 &
  0.948 &
  0.908 &
  0.968 &
  - &
  4.348 &
  5.409 &
  34.635 &
  \multirow{3}{*}{24.003} \\ \cmidrule(lr){2-2} \cmidrule(lr){4-12}
 &
  \textbf{Phonology} &
   &
  0.141 &
  12.201 &
  0.953 &
  0.904 &
  0.968 &
  0.88 &
  3.349 &
  5.343 &
  33.275 &
   \\ \cmidrule(lr){2-2} \cmidrule(lr){4-12}
 &
  \textbf{Orthography} &
   &
  0.19 &
  13.535 &
  0.944 &
  0.892 &
  0.965 &
  - &
  2.438 &
  5.286 &
  32.117 &
   \\ \midrule
\multirow{3}{*}{\textbf{bd}} &
  \textbf{Random} &
  \multirow{3}{*}{0.487} &
  0.056 &
  7.899 &
  0.978 &
  0.942 &
  0.985 &
  - &
  5.282 &
  5.648 &
  18.058 &
  \multirow{3}{*}{22.205} \\ \cmidrule(lr){2-2} \cmidrule(lr){4-12}
 &
  \textbf{Phonology} &
   &
  0.103 &
  7.656 &
  0.981 &
  0.945 &
  0.987 &
  - &
  3.628 &
  5.618 &
  15.188 &
   \\ \cmidrule(lr){2-2} \cmidrule(lr){4-12}
 &
  \textbf{Orthography} &
   &
  0.172 &
  6.73 &
  0.98 &
  0.949 &
  0.987 &
  - &
  2.202 &
  5.681 &
  12.081 &
   \\ \midrule
\multirow{3}{*}{\textbf{bn}} &
  \textbf{Random} &
  \multirow{3}{*}{0.891} &
  0.125 &
  11.208 &
  0.944 &
  0.91 &
  0.962 &
  - &
  4.715 &
  5.205 &
  35.019 &
  \multirow{3}{*}{23.721} \\ \cmidrule(lr){2-2} \cmidrule(lr){4-12}
 &
  \textbf{Phonology} &
   &
  0.193 &
  11.674 &
  0.949 &
  0.908 &
  0.964 &
  0.956 &
  3.386 &
  5.168 &
  32.864 &
   \\ \cmidrule(lr){2-2} \cmidrule(lr){4-12}
 &
  \textbf{Orthography} &
   &
  0.214 &
  13.101 &
  0.939 &
  0.896 &
  0.96 &
  - &
  2.349 &
  5.133 &
  31.508 &
   \\ \midrule
\multirow{3}{*}{\textbf{gu}} &
  \textbf{Random} &
  \multirow{3}{*}{0.89} &
  0.087 &
  11.52 &
  0.94 &
  0.902 &
  0.962 &
  - &
  4.259 &
  4.758 &
  37.929 &
  \multirow{3}{*}{26.129} \\ \cmidrule(lr){2-2} \cmidrule(lr){4-12}
 &
  \textbf{Phonology} &
   &
  0.138 &
  12.433 &
  0.94 &
  0.896 &
  0.963 &
  0.966 &
  3.067 &
  4.722 &
  35.817 &
   \\ \cmidrule(lr){2-2} \cmidrule(lr){4-12}
 &
  \textbf{Orthography} &
   &
  0.223 &
  12.301 &
  0.935 &
  0.893 &
  0.959 &
  - &
  1.646 &
  4.684 &
  31.953 &
   \\ \midrule
\multirow{3}{*}{\textbf{hi}} &
  \textbf{Random} &
  \multirow{3}{*}{0.902} &
  0.101 &
  9.641 &
  0.944 &
  0.904 &
  0.963 &
  - &
  3.781 &
  4.215 &
  40.589 &
  \multirow{3}{*}{30.107} \\ \cmidrule(lr){2-2} \cmidrule(lr){4-12}
 &
  \textbf{Phonology} &
   &
  0.181 &
  10.692 &
  0.946 &
  0.896 &
  0.964 &
  0.971 &
  2.748 &
  4.157 &
  39.562 &
   \\ \cmidrule(lr){2-2} \cmidrule(lr){4-12}
 &
  \textbf{Orthography} &
   &
  0.169 &
  9.785 &
  0.943 &
  0.899 &
  0.963 &
  - &
  1.883 &
  4.138 &
  36.064 &
   \\ \midrule
\multirow{3}{*}{\textbf{kn}} &
  \textbf{Random} &
  \multirow{3}{*}{0.903} &
  0.102 &
  9.593 &
  0.945 &
  0.905 &
  0.963 &
  - &
  3.785 &
  4.219 &
  40.565 &
  \multirow{3}{*}{20.034} \\ \cmidrule(lr){2-2} \cmidrule(lr){4-12}
 &
  \textbf{Phonology} &
   &
  0.211 &
  14.77 &
  0.95 &
  0.917 &
  0.962 &
  0.972 &
  4.388 &
  7.912 &
  32.446 &
   \\ \cmidrule(lr){2-2} \cmidrule(lr){4-12}
 &
  \textbf{Orthography} &
   &
  0.38 &
  14.729 &
  0.948 &
  0.913 &
  0.961 &
  - &
  2.101 &
  7.723 &
  27.1 &
   \\ \midrule
\multirow{3}{*}{\textbf{ml}} &
  \textbf{Random} &
  \multirow{3}{*}{0.887} &
  0.134 &
  13.224 &
  0.955 &
  0.93 &
  0.963 &
  - &
  6.695 &
  9.141 &
  34.133 &
  \multirow{3}{*}{19.254} \\ \cmidrule(lr){2-2} \cmidrule(lr){4-12}
 &
  \textbf{Phonology} &
   &
  0.278 &
  15.01 &
  0.959 &
  0.925 &
  0.963 &
  0.944 &
  4.165 &
  8.865 &
  32.098 &
   \\ \cmidrule(lr){2-2} \cmidrule(lr){4-12}
 &
  \textbf{Orthography} &
   &
  0.285 &
  14.817 &
  0.953 &
  0.923 &
  0.96 &
  - &
  3.232 &
  8.874 &
  26.926 &
   \\ \midrule
\multirow{3}{*}{\textbf{mr}} &
  \textbf{Random} &
  \multirow{3}{*}{0.9} &
  0.103 &
  11.694 &
  0.946 &
  0.911 &
  0.963 &
  - &
  4.918 &
  5.529 &
  35.263 &
  \multirow{3}{*}{23.172} \\ \cmidrule(lr){2-2} \cmidrule(lr){4-12}
 &
  \textbf{Phonology} &
   &
  0.192 &
  12.74 &
  0.951 &
  0.908 &
  0.965 &
  0.96 &
  3.514 &
  5.539 &
  33.117 &
   \\ \cmidrule(lr){2-2} \cmidrule(lr){4-12}
 &
  \textbf{Orthography} &
   &
  0.15 &
  13.063 &
  0.942 &
  0.901 &
  0.962 &
  - &
  2.626 &
  5.526 &
  31.208 &
   \\ \midrule
\multirow{3}{*}{\textbf{or}} &
  \textbf{Random} &
  \multirow{3}{*}{0.865} &
  0.076 &
  10.646 &
  0.95 &
  0.917 &
  0.97 &
  - &
  4.72 &
  5.424 &
  32.816 &
  \multirow{3}{*}{23.491} \\ \cmidrule(lr){2-2} \cmidrule(lr){4-12}
 &
  \textbf{Phonology} &
   &
  0.12 &
  11.808 &
  0.948 &
  0.91 &
  0.969 &
  0.952 &
  3.414 &
  5.403 &
  31.641 &
   \\ \cmidrule(lr){2-2} \cmidrule(lr){4-12}
 &
  \textbf{Orthography} &
   &
  0.228 &
  11.968 &
  0.95 &
  0.907 &
  0.968 &
  - &
  2.138 &
  5.377 &
  30.164 &
   \\ \midrule
\multirow{3}{*}{\textbf{pa}} &
  \textbf{Random} &
  \multirow{3}{*}{0.876} &
  0.1 &
  8.914 &
  0.948 &
  0.908 &
  0.965 &
  - &
  3.787 &
  4.264 &
  38.7 &
  \multirow{3}{*}{30.062} \\ \cmidrule(lr){2-2} \cmidrule(lr){4-12}
 &
  \textbf{Phonology} &
   &
  0.16 &
  10.06 &
  0.947 &
  0.901 &
  0.965 &
  0.93 &
  2.744 &
  4.199 &
  38.168 &
   \\ \cmidrule(lr){2-2} \cmidrule(lr){4-12}
 &
  \textbf{Orthography} &
   &
  0.185 &
  8.993 &
  0.951 &
  0.908 &
  0.964 &
  - &
  1.817 &
  4.167 &
  34.765 &
   \\ \midrule
\multirow{3}{*}{\textbf{ta}} &
  \textbf{Random} &
  \multirow{3}{*}{0.892} &
  0.125 &
  12.612 &
  0.948 &
  0.931 &
  0.965 &
  - &
  5.997 &
  8.148 &
  34.436 &
  \multirow{3}{*}{20.874} \\ \cmidrule(lr){2-2} \cmidrule(lr){4-12}
 &
  \textbf{Phonology} &
   &
  0.358 &
  14.95 &
  0.95 &
  0.923 &
  0.965 &
  0.963 &
  4.161 &
  7.916 &
  27.714 &
   \\ \cmidrule(lr){2-2} \cmidrule(lr){4-12}
 &
  \textbf{Orthography} &
   &
  0.331 &
  13.743 &
  0.959 &
  0.924 &
  0.962 &
  - &
  2.224 &
  7.919 &
  26.771 &
   \\ \midrule
\multirow{3}{*}{\textbf{te}} &
  \textbf{Random} &
  \multirow{3}{*}{0.885} &
  0.116 &
  12.729 &
  0.949 &
  0.919 &
  0.963 &
  - &
  5.875 &
  7.062 &
  36.191 &
  \multirow{3}{*}{21.76} \\ \cmidrule(lr){2-2} \cmidrule(lr){4-12}
 &
  \textbf{Phonology} &
   &
  0.191 &
  13.548 &
  0.951 &
  0.917 &
  0.965 &
  0.964 &
  4.19 &
  6.977 &
  33.21 &
   \\ \cmidrule(lr){2-2} \cmidrule(lr){4-12}
 &
  \textbf{Orthography} &
   &
  0.285 &
  14.535 &
  0.943 &
  0.908 &
  0.96 &
  - &
  2.087 &
  6.847 &
  28.113 &
   \\ \bottomrule
\end{tabular}%
}
\caption{The table presents the impact of different character-substitution attack strategies: random (\textbf{Rand}), phonetic (\textbf{Phono}), and orthographic (\textbf{Ortho}) on the \textit{MuRIL} language model for the \textit{IndicSentiment} dataset.}
\label{tab:senti_muril}
\end{table*}

\begin{table*}[]
\centering
\resizebox{\textwidth}{!}{%
\begin{tabular}{@{}ccccccccccccc@{}}
\toprule
\textbf{Lang} &
  \textbf{\begin{tabular}[c]{@{}c@{}}Type\\ of\\ Perturbation\end{tabular}} &
  \textbf{\begin{tabular}[c]{@{}c@{}}Original\\ Accuracy\end{tabular}} &
  \textbf{\begin{tabular}[c]{@{}c@{}}After\\ Attack\\ Accuracy\end{tabular}} &
  \textbf{\begin{tabular}[c]{@{}c@{}}\%\\ Perturbed\\ Words\end{tabular}} &
  \textbf{\begin{tabular}[c]{@{}c@{}}Semantic\\ Similarity\end{tabular}} &
  \textbf{\begin{tabular}[c]{@{}c@{}}Overlap\\ Similarity\end{tabular}} &
  \textbf{\begin{tabular}[c]{@{}c@{}}BERTScore\\ based\\ Similarity\end{tabular}} &
  \textbf{\begin{tabular}[c]{@{}c@{}}Phonetic\\ Similarity\end{tabular}} &
  \textbf{\begin{tabular}[c]{@{}c@{}}Avg. No. of\\ Candidates\\ per word\end{tabular}} &
  \textbf{\begin{tabular}[c]{@{}c@{}}Avg.\\ Word\\ Length\end{tabular}} &
  \textbf{\begin{tabular}[c]{@{}c@{}}Query\\ Number\end{tabular}} &
  \textbf{\begin{tabular}[c]{@{}c@{}}Avg.\\ Sentence\\ Length\end{tabular}} \\ \midrule
\multirow{3}{*}{\textbf{as}} &
  \textbf{Rand} &
  \multirow{3}{*}{0.601} &
  0.222/0.16 &
  10.052/9.917 &
  0.961/0.961 &
  0.931/0.933 &
  0.976/0.976 &
  - &
  4.593/4.55 &
  5.739/5.689 &
  18.702/18.568 &
  \multirow{3}{*}{17.814/17.628} \\ \cmidrule(lr){2-2} \cmidrule(lr){4-12}
 &
  \textbf{Phono} &
   &
  0.322/0.272 &
  8.953/9.329 &
  0.971/0.967 &
  0.939/0.937 &
  0.98/0.98 &
  0.895/0.896 &
  3.606/3.569 &
  5.691/5.62 &
  14.663/15.258 &
   \\ \cmidrule(lr){2-2} \cmidrule(lr){4-12}
 &
  \textbf{Ortho} &
   &
  0.297/0.241 &
  8.909/9.214 &
  0.965/0.963 &
  0.936/0.934 &
  0.979/0.978 &
  - &
  2.47/2.428 &
  5.683/5.608 &
  13.277/13.79 &
   \\ \midrule
\multirow{3}{*}{\textbf{bn}} &
  \textbf{Rand} &
  \multirow{3}{*}{0.517} &
  0.129/0.063 &
  9.336/8.745 &
  0.959/0.961 &
  0.938/0.942 &
  0.975/0.975 &
  - &
  5.24/5.231 &
  5.908/5.897 &
  14.987/14.49 &
  \multirow{3}{*}{15.031/15.299} \\ \cmidrule(lr){2-2} \cmidrule(lr){4-12}
 &
  \textbf{Phono} &
   &
  0.259/0.161 &
  7.557/8.545 &
  0.969/0.962 &
  0.949/0.943 &
  0.981/0.977 &
  0.984/0.979 &
  3.691/3.675 &
  5.862/5.849 &
  10.314/11.779 &
   \\ \cmidrule(lr){2-2} \cmidrule(lr){4-12}
 &
  \textbf{Ortho} &
   &
  0.242/0.131 &
  7.426/8.459 &
  0.966/0.962 &
  0.949/0.942 &
  0.98/0.977 &
  - &
  2.388/2.377 &
  5.859/5.842 &
  9.195/10.688 &
   \\ \midrule
\multirow{3}{*}{\textbf{gu}} &
  \textbf{Rand} &
  \multirow{3}{*}{0.837} &
  0.32/0.202 &
  14.997/14.926 &
  0.934/0.934 &
  0.892/0.894 &
  0.962/0.961 &
  - &
  4.772/4.75 &
  5.368/5.331 &
  27.097/26.391 &
  \multirow{3}{*}{16.53/16.594} \\ \cmidrule(lr){2-2} \cmidrule(lr){4-12}
 &
  \textbf{Phono} &
   &
  0.536/0.408 &
  13.74/15.085 &
  0.947/0.941 &
  0.903/0.892 &
  0.968/0.964 &
  0.972/0.968 &
  3.242/3.244 &
  5.306/5.295 &
  21.155/21.958 &
   \\ \cmidrule(lr){2-2} \cmidrule(lr){4-12}
 &
  \textbf{Ortho} &
   &
  0.547/0.403 &
  12.029/13.807 &
  0.945/0.935 &
  0.907/0.896 &
  0.967/0.961 &
  - &
  1.812/1.815 &
  5.279/5.283 &
  17.204/18.787 &
   \\ \midrule
\multirow{3}{*}{\textbf{hi}} &
  \textbf{Rand} &
  \multirow{3}{*}{0.532} &
  0.126/0.064 &
  8.795/7.979 &
  0.957/0.956 &
  0.928/0.932 &
  0.973/0.972 &
  - &
  4.023/4.016 &
  4.517/4.531 &
  18.416/18.032 &
  \multirow{3}{*}{19.749/20.201} \\ \cmidrule(lr){2-2} \cmidrule(lr){4-12}
 &
  \textbf{Phono} &
   &
  0.305/0.207 &
  6.743/7.604 &
  0.97/0.961 &
  0.945/0.935 &
  0.981/0.976 &
  0.985/0.982 &
  2.784/2.751 &
  4.435/4.424 &
  11.995/14.325 &
   \\ \cmidrule(lr){2-2} \cmidrule(lr){4-12}
 &
  \textbf{Ortho} &
   &
  0.241/0.135 &
  7.35/7.465 &
  0.963/0.957 &
  0.939/0.934 &
  0.978/0.975 &
  - &
  2.009/2.01 &
  4.429/4.461 &
  12.403/13.733 &
   \\ \midrule
\multirow{3}{*}{\textbf{kn}} &
  \textbf{Rand} &
  \multirow{3}{*}{0.608} &
  0.225/0.184 &
  10.66/10.383 &
  0.963/0.964 &
  0.941/0.943 &
  0.975/0.975 &
  - &
  5.766/5.754 &
  7.252/7.241 &
  16.581/16.359 &
  \multirow{3}{*}{14.402/14.236} \\ \cmidrule(lr){2-2} \cmidrule(lr){4-12}
 &
  \textbf{Phono} &
   &
  0.387/0.331 &
  8.206/8.843 &
  0.976/0.972 &
  0.955/0.953 &
  0.982/0.981 &
  0.984/0.984 &
  3.886/3.882 &
  7.136/7.188 &
  10.706/11.359 &
   \\ \cmidrule(lr){2-2} \cmidrule(lr){4-12}
 &
  \textbf{Ortho} &
   &
  0.417/0.368 &
  7.011/7.581 &
  0.979/0.975 &
  0.959/0.956 &
  0.983/0.981 &
  - &
  2.177/2.183 &
  7.09/7.146 &
  7.813/8.389 &
   \\ \midrule
\multirow{3}{*}{\textbf{ml}} &
  \textbf{Rand} &
  \multirow{3}{*}{0.596} &
  0.269/0.219 &
  9.803/9.989 &
  0.968/0.967 &
  0.951/0.952 &
  0.977/0.976 &
  - &
  6.075/6.127 &
  8.439/8.496 &
  14.382/14.43 &
  \multirow{3}{*}{13.238/13.265} \\ \cmidrule(lr){2-2} \cmidrule(lr){4-12}
 &
  \textbf{Phono} &
   &
  0.427/0.377 &
  6.389/7.513 &
  0.983/0.979 &
  0.969/0.964 &
  0.986/0.983 &
  0.954/0.954 &
  4.079/4.129 &
  8.276/8.407 &
  7.895/9.047 &
   \\ \cmidrule(lr){2-2} \cmidrule(lr){4-12}
 &
  \textbf{Ortho} &
   &
  0.334/0.3 &
  8.517/9.119 &
  0.973/0.969 &
  0.957/0.955 &
  0.979/0.977 &
  - &
  3.142/3.155 &
  8.31/8.388 &
  8.992/9.341 &
   \\ \midrule
\multirow{3}{*}{\textbf{mr}} &
  \textbf{Rand} &
  \multirow{3}{*}{0.625} &
  0.183/0.118 &
  11.179/11.002 &
  0.96/0.96 &
  0.93/0.931 &
  0.971/0.971 &
  - &
  5.338/5.247 &
  6.025/5.943 &
  18.915/19.039 &
  \multirow{3}{*}{15.938/16.051} \\ \cmidrule(lr){2-2} \cmidrule(lr){4-12}
 &
  \textbf{Phono} &
   &
  0.352/0.272 &
  8.736/10.156 &
  0.972/0.968 &
  0.945/0.937 &
  0.979/0.976 &
  0.98/0.978 &
  3.673/3.63 &
  6/5.938 &
  12.801/14.666 &
   \\ \cmidrule(lr){2-2} \cmidrule(lr){4-12}
 &
  \textbf{Ortho} &
   &
  0.305/0.239 &
  9.669/10.122 &
  0.965/0.963 &
  0.938/0.935 &
  0.976/0.974 &
  - &
  2.771/2.731 &
  5.989/5.925 &
  12.79/13.772 &
   \\ \midrule
\multirow{3}{*}{\textbf{or}} &
  \textbf{Rand} &
  \multirow{3}{*}{0.596} &
  0.213/0.143 &
  10.14/10.743 &
  0.962/0.959 &
  0.936/0.932 &
  0.975/0.974 &
  - &
  5.234/5.122 &
  5.988/5.865 &
  17.524/18.157 &
  \multirow{3}{*}{15.884/16.035} \\ \cmidrule(lr){2-2} \cmidrule(lr){4-12}
 &
  \textbf{Phono} &
   &
  0.335/0.24 &
  8.236/10.443 &
  0.97/0.963 &
  0.948/0.935 &
  0.981/0.976 &
  0.966/0.963 &
  3.67/3.64 &
  5.901/5.886 &
  12.076/14.596 &
   \\ \cmidrule(lr){2-2} \cmidrule(lr){4-12}
 &
  \textbf{Ortho} &
   &
  0.34/0.253 &
  7.854/9.331 &
  0.968/0.963 &
  0.948/0.938 &
  0.98/0.976 &
  - &
  2.401/2.395 &
  5.88/5.854 &
  10.423/12.249 &
   \\ \midrule
\multirow{3}{*}{\textbf{pa}} &
  \textbf{Rand} &
  \multirow{3}{*}{0.561} &
  0.15/0.112 &
  9.051/8.45 &
  0.954/0.957 &
  0.922/0.927 &
  0.973/0.974 &
  - &
  3.879/3.789 &
  4.325/4.218 &
  19.472/18.528 &
  \multirow{3}{*}{20.124/20.372} \\ \cmidrule(lr){2-2} \cmidrule(lr){4-12}
 &
  \textbf{Phono} &
   &
  0.322/0.243 &
  6.482/7.936 &
  0.969/0.963 &
  0.945/0.933 &
  0.981/0.978 &
  0.941/0.938 &
  2.784/2.737 &
  4.244/4.186 &
  12.647/15.219 &
   \\ \cmidrule(lr){2-2} \cmidrule(lr){4-12}
 &
  \textbf{Ortho} &
   &
  0.234/0.163 &
  7.664/8.23 &
  0.963/0.959 &
  0.934/0.929 &
  0.977/0.974 &
  - &
  1.883/1.866 &
  4.263/4.199 &
  13.676/14.854 &
   \\ \midrule
\multirow{3}{*}{\textbf{te}} &
  \textbf{Rand} &
  \multirow{3}{*}{0.591} &
  0.208/0.156 &
  10.044/10.392 &
  0.966/0.963 &
  0.943/0.94 &
  0.976/0.975 &
  - &
  5.724/5.659 &
  7.018/6.943 &
  14.843/15.132 &
  \multirow{3}{*}{13.726/13.731} \\ \cmidrule(lr){2-2} \cmidrule(lr){4-12}
 &
  \textbf{Phono} &
   &
  0.358/0.301 &
  7.587/8.863 &
  0.979/0.974 &
  0.958/0.95 &
  0.984/0.981 &
  0.986/0.983 &
  3.954/3.909 &
  6.98/6.911 &
  9.749/11.136 &
   \\ \cmidrule(lr){2-2} \cmidrule(lr){4-12}
 &
  \textbf{Ortho} &
   &
  0.377/0.315 &
  6.395/8.133 &
  0.981/0.972 &
  0.961/0.95 &
  0.984/0.98 &
  - &
  2.258/2.245 &
  6.96/6.89 &
  7.443/8.831 &
   \\ \bottomrule
\end{tabular}%
}
\caption{The table presents the impact of different character-substitution attack strategies: random (\textbf{Rand}), phonetic (\textbf{Phono}), and orthographic (\textbf{Ortho}) on the \textit{MuRIL} language model for the \textit{IndicParaphrase} dataset. Results are shown for perturbing both \textit{sentence1} and \textit{sentence2} of the IndicParaphrase dataset, separated by a forward slash (/).}
\label{tab:para_muril}
\end{table*}

\begin{table*}[]
\centering
\resizebox{\textwidth}{!}{%
\begin{tabular}{@{}ccccccccccccc@{}}
\toprule
\textbf{Lang} &
  \textbf{\begin{tabular}[c]{@{}c@{}}Type\\ of\\ Perturbation\end{tabular}} &
  \textbf{\begin{tabular}[c]{@{}c@{}}Original\\ Accuracy\end{tabular}} &
  \textbf{\begin{tabular}[c]{@{}c@{}}After\\ Attack\\ Accuracy\end{tabular}} &
  \textbf{\begin{tabular}[c]{@{}c@{}}\%\\ Perturbed\\ Words\end{tabular}} &
  \textbf{\begin{tabular}[c]{@{}c@{}}Semantic\\ Similarity\end{tabular}} &
  \textbf{\begin{tabular}[c]{@{}c@{}}Overlap\\ Similarity\end{tabular}} &
  \textbf{\begin{tabular}[c]{@{}c@{}}BERTScore\\ based\\ Similarity\end{tabular}} &
  \textbf{\begin{tabular}[c]{@{}c@{}}Phonetic\\ Similarity\end{tabular}} &
  \textbf{\begin{tabular}[c]{@{}c@{}}Avg. No. of\\ Candidates\\ per word\end{tabular}} &
  \textbf{\begin{tabular}[c]{@{}c@{}}Avg.\\ Word\\ Length\end{tabular}} &
  \textbf{\begin{tabular}[c]{@{}c@{}}Query\\ Number\end{tabular}} &
  \textbf{\begin{tabular}[c]{@{}c@{}}Avg.\\ Sentence\\ Length\end{tabular}} \\ \midrule
\multirow{3}{*}{\textbf{as}} &
  \textbf{Rand} &
  \multirow{3}{*}{0.688} &
  0.175/0.075 &
  8.88/12.17 &
  0.949/0.925 &
  0.922/0.897 &
  0.973/0.97 &
  - &
  4.136/4.307 &
  5.178/5.308 &
  17.864/10.214 &
  \multirow{3}{*}{17.091/9.459} \\ \cmidrule(lr){2-2} \cmidrule(lr){4-12}
 &
  \textbf{Phono} &
   &
  0.246/0.153 &
  9.091/12.406 &
  0.954/0.935 &
  0.922/0.899 &
  0.975/0.971 &
  0.89/0.891 &
  3.291/3.342 &
  5.161/5.253 &
  17.047/9.932 &
   \\ \cmidrule(lr){2-2} \cmidrule(lr){4-12}
 &
  \textbf{Ortho} &
   &
  0.282/0.191 &
  8.322/12.529 &
  0.954/0.935 &
  0.926/0.898 &
  0.975/0.969 &
  - &
  2.188/2.14 &
  5.12/5.188 &
  14.452/8.882 &
   \\ \midrule
\multirow{3}{*}{\textbf{bn}} &
  \textbf{Rand} &
  \multirow{3}{*}{0.749} &
  0.177/0.084 &
  10.146/13.473 &
  0.932/0.905 &
  0.909/0.885 &
  0.964/0.959 &
  - &
  4.503/4.527 &
  4.988/4.996 &
  20.477/10.986 &
  \multirow{3}{*}{16.971/9.221} \\ \cmidrule(lr){2-2} \cmidrule(lr){4-12}
 &
  \textbf{Phono} &
   &
  0.274/0.185 &
  10.336/14.281 &
  0.939/0.912 &
  0.908/0.881 &
  0.967/0.961 &
  0.973/0.978 &
  3.254/3.281 &
  4.954/4.984 &
  18.682/10.669 &
   \\ \cmidrule(lr){2-2} \cmidrule(lr){4-12}
 &
  \textbf{Ortho} &
   &
  0.315/0.242 &
  9.211/13.646 &
  0.941/0.915 &
  0.915/0.883 &
  0.967/0.961 &
  - &
  2.063/2.01 &
  4.906/4.936 &
  15.837/9.308 &
   \\ \midrule
\multirow{3}{*}{\textbf{gu}} &
  \textbf{Rand} &
  \multirow{3}{*}{0.744} &
  0.173/0.071 &
  9.671/12.869 &
  0.934/0.907 &
  0.904/0.881 &
  0.966/0.962 &
  - &
  4.055/4.05 &
  4.526/4.538 &
  21.037/11.449 &
  \multirow{3}{*}{18.389/9.956} \\ \cmidrule(lr){2-2} \cmidrule(lr){4-12}
 &
  \textbf{Phono} &
   &
  0.27/0.178 &
  9.794/13.588 &
  0.942/0.916 &
  0.908/0.878 &
  0.969/0.965 &
  0.973/0.979 &
  2.85/2.828 &
  4.495/4.491 &
  19.198/10.957 &
   \\ \cmidrule(lr){2-2} \cmidrule(lr){4-12}
 &
  \textbf{Ortho} &
   &
  0.339/0.268 &
  8.292/12.172 &
  0.942/0.918 &
  0.914/0.886 &
  0.968/0.964 &
  - &
  1.432/1.434 &
  4.441/4.445 &
  15.766/9.228 &
   \\ \midrule
\multirow{3}{*}{\textbf{hi}} &
  \textbf{Rand} &
  \multirow{3}{*}{0.752} &
  0.178/0.069 &
  8.086/11.217 &
  0.939/0.91 &
  0.908/0.879 &
  0.965/0.96 &
  - &
  3.477/3.631 &
  3.87/4.079 &
  22.863/12.476 &
  \multirow{3}{*}{21.825/11.492} \\ \cmidrule(lr){2-2} \cmidrule(lr){4-12}
 &
  \textbf{Phono} &
   &
  0.282/0.211 &
  9.377/12.511 &
  0.941/0.913 &
  0.899/0.873 &
  0.967/0.963 &
  0.976/0.98 &
  2.459/2.558 &
  3.819/3.977 &
  22.484/12.197 &
   \\ \cmidrule(lr){2-2} \cmidrule(lr){4-12}
 &
  \textbf{Ortho} &
   &
  0.262/0.141 &
  8.193/11.943 &
  0.938/0.908 &
  0.907/0.876 &
  0.967/0.961 &
  - &
  1.699/1.74 &
  3.835/3.974 &
  19.655/11.344 &
   \\ \midrule
\multirow{3}{*}{\textbf{kn}} &
  \textbf{Rand} &
  \multirow{3}{*}{0.737} &
  0.203/0.118 &
  11.048/15.749 &
  0.943/0.922 &
  0.927/0.908 &
  0.968/0.965 &
  - &
  5.874/6.394 &
  7.391/8.024 &
  18.663/10.809 &
  \multirow{3}{*}{14.072/7.533} \\ \cmidrule(lr){2-2} \cmidrule(lr){4-12}
 &
  \textbf{Phono} &
   &
  0.345/0.266 &
  10.355/16.166 &
  0.956/0.932 &
  0.933/0.905 &
  0.973/0.968 &
  0.984/0.985 &
  3.932/4.166 &
  7.278/7.744 &
  15.588/9.673 &
   \\ \cmidrule(lr){2-2} \cmidrule(lr){4-12}
 &
  \textbf{Ortho} &
   &
  0.448/0.408 &
  8.328/13.624 &
  0.963/0.942 &
  0.942/0.917 &
  0.975/0.97 &
  - &
  2.013/2.093 &
  7.214/7.585 &
  10.99/6.709 &
   \\ \midrule
\multirow{3}{*}{\textbf{ml}} &
  \textbf{Rand} &
  \multirow{3}{*}{0.721} &
  0.229/0.133 &
  9.996/16.219 &
  0.946/0.921 &
  0.936/0.917 &
  0.967/0.963 &
  - &
  6.192/6.838 &
  8.705/9.464 &
  16.139/10.16 &
  \multirow{3}{*}{12.98/6.837} \\ \cmidrule(lr){2-2} \cmidrule(lr){4-12}
 &
  \textbf{Phono} &
   &
  0.372/0.306 &
  10.022/16.124 &
  0.96/0.939 &
  0.94/0.918 &
  0.972/0.968 &
  0.954/0.952 &
  3.999/4.24 &
  8.521/9.111 &
  13.99/8.632 &
   \\ \cmidrule(lr){2-2} \cmidrule(lr){4-12}
 &
  \textbf{Ortho} &
   &
  0.3/0.251 &
  9.464/15.102 &
  0.951/0.93 &
  0.938/0.92 &
  0.968/0.963 &
  - &
  3.109/3.288 &
  8.543/9.176 &
  11.866/7.054 &
   \\ \midrule
\multirow{3}{*}{\textbf{mr}} &
  \textbf{Rand} &
  \multirow{3}{*}{0.711} &
  0.189/0.082 &
  9.797/13.337 &
  0.937/0.913 &
  0.913/0.892 &
  0.966/0.963 &
  - &
  4.759/4.872 &
  5.376/5.434 &
  18.264/10.069 &
  \multirow{3}{*}{16.154/8.773} \\ \cmidrule(lr){2-2} \cmidrule(lr){4-12}
 &
  \textbf{Phono} &
   &
  0.284/0.216 &
  10.73/13.988 &
  0.942/0.922 &
  0.909/0.892 &
  0.969/0.966 &
  0.973/0.977 &
  3.255/3.291 &
  5.305/5.346 &
  16.821/9.395 &
   \\ \cmidrule(lr){2-2} \cmidrule(lr){4-12}
 &
  \textbf{Ortho} &
   &
  0.277/0.15 &
  9.723/13.726 &
  0.938/0.912 &
  0.913/0.89 &
  0.968/0.964 &
  - &
  2.488/2.477 &
  5.306/5.379 &
  14.614/8.81 &
   \\ \midrule
\multirow{3}{*}{\textbf{or}} &
  \textbf{Rand} &
  \multirow{3}{*}{0.709} &
  0.181/0.08 &
  9.452/13.537 &
  0.944/0.921 &
  0.917/0.895 &
  0.97/0.965 &
  - &
  4.839/4.846 &
  5.454/5.475 &
  18.008/10.561 &
  \multirow{3}{*}{16.173/8.847} \\ \cmidrule(lr){2-2} \cmidrule(lr){4-12}
 &
  \textbf{Phono} &
   &
  0.256/0.179 &
  10.198/14.244 &
  0.945/0.921 &
  0.914/0.893 &
  0.972/0.967 &
  0.965/0.967 &
  3.529/3.517 &
  5.389/5.427 &
  17.026/9.946 &
   \\ \cmidrule(lr){2-2} \cmidrule(lr){4-12}
 &
  \textbf{Ortho} &
   &
  0.321/0.265 &
  8.512/13.123 &
  0.953/0.93 &
  0.925/0.899 &
  0.974/0.968 &
  - &
  1.969/1.889 &
  5.317/5.394 &
  13.582/8.358 &
   \\ \midrule
\multirow{3}{*}{\textbf{pa}} &
  \textbf{Rand} &
  \multirow{3}{*}{0.747} &
  0.159/0.056 &
  8.667/11.343 &
  0.936/0.909 &
  0.903/0.879 &
  0.966/0.963 &
  - &
  3.496/3.567 &
  3.949/4.056 &
  23.732/12.642 &
  \multirow{3}{*}{21.909/11.725} \\ \cmidrule(lr){2-2} \cmidrule(lr){4-12}
 &
  \textbf{Phono} &
   &
  0.243/0.182 &
  9.246/11.701 &
  0.936/0.913 &
  0.899/0.878 &
  0.967/0.965 &
  0.94/0.939 &
  2.57/2.547 &
  3.912/3.954 &
  22.522/12.101 &
   \\ \cmidrule(lr){2-2} \cmidrule(lr){4-12}
 &
  \textbf{Ortho} &
   &
  0.244/0.13 &
  8.114/11.71 &
  0.943/0.916 &
  0.909/0.877 &
  0.967/0.962 &
  - &
  1.642/1.673 &
  3.884/3.954 &
  19.627/11.354 &
   \\ \midrule
\multirow{3}{*}{\textbf{ta}} &
  \textbf{Rand} &
  \multirow{3}{*}{0.725} &
  0.192/0.122 &
  10.086/14.891 &
  0.938/0.91 &
  0.936/0.917 &
  0.969/0.965 &
  - &
  5.421/5.892 &
  7.778/8.202 &
  17.819/10.298 &
  \multirow{3}{*}{14.721/7.826} \\ \cmidrule(lr){2-2} \cmidrule(lr){4-12}
 &
  \textbf{Phono} &
   &
  0.385/0.335 &
  9.396/14.625 &
  0.953/0.93 &
  0.942/0.92 &
  0.975/0.97 &
  0.974/0.978 &
  3.748/3.949 &
  7.628/7.962 &
  12.525/7.537 &
   \\ \cmidrule(lr){2-2} \cmidrule(lr){4-12}
 &
  \textbf{Ortho} &
   &
  0.341/0.318 &
  8.636/13.826 &
  0.96/0.943 &
  0.943/0.921 &
  0.972/0.966 &
  - &
  2.085/2.207 &
  7.583/7.968 &
  11.936/7.107 &
   \\ \midrule
\multirow{3}{*}{\textbf{te}} &
  \textbf{Rand} &
  \multirow{3}{*}{0.72} &
  0.209/0.131 &
  10.089/15.039 &
  0.942/0.92 &
  0.924/0.901 &
  0.969/0.965 &
  - &
  5.272/5.63 &
  6.411/6.798 &
  17.684/10.19 &
  \multirow{3}{*}{14.655/7.706} \\ \cmidrule(lr){2-2} \cmidrule(lr){4-12}
 &
  \textbf{Phono} &
   &
  0.296/0.227 &
  10.095/15.255 &
  0.952/0.927 &
  0.926/0.901 &
  0.973/0.968 &
  0.971/0.98 &
  3.671/3.98 &
  6.362/6.732 &
  15.49/9.268 &
   \\ \cmidrule(lr){2-2} \cmidrule(lr){4-12}
 &
  \textbf{Ortho} &
   &
  0.369/0.313 &
  8.743/14.355 &
  0.952/0.925 &
  0.933/0.905 &
  0.973/0.968 &
  - &
  1.927/2.097 &
  6.299/6.652 &
  11.711/7.302 &
   \\ \bottomrule
\end{tabular}%
}
\caption{The table presents the impact of different character-substitution attack strategies: random (\textbf{Rand}), phonetic (\textbf{Phono}), and orthographic (\textbf{Ortho}) on the \textit{MuRIL} language model for the \textit{IndicXNLI} dataset. Results are shown for perturbing both \textit{premise} and \textit{hypothesis} of the IndicXNLI dataset, separated by a forward slash (/).}
\label{tab:xnli_muril}
\end{table*}

\begin{table*}[]
\centering
\resizebox{\textwidth}{!}{%
\begin{tabular}{@{}ccccccccccccc@{}}
\toprule
\textbf{Language} &
  \textbf{\begin{tabular}[c]{@{}c@{}}Type\\ of\\ Perturbation\end{tabular}} &
  \textbf{\begin{tabular}[c]{@{}c@{}}Original\\ Accuracy\end{tabular}} &
  \textbf{\begin{tabular}[c]{@{}c@{}}After\\ Attack\\ Accuracy\end{tabular}} &
  \textbf{\begin{tabular}[c]{@{}c@{}}\%\\ Perturbed\\ Words\end{tabular}} &
  \textbf{\begin{tabular}[c]{@{}c@{}}Semantic\\ Similarity\end{tabular}} &
  \textbf{\begin{tabular}[c]{@{}c@{}}Overlap\\ Similarity\end{tabular}} &
  \textbf{\begin{tabular}[c]{@{}c@{}}BERTScore\\ based\\ Similarity\end{tabular}} &
  \textbf{\begin{tabular}[c]{@{}c@{}}Phonetic\\ Similarity\end{tabular}} &
  \textbf{\begin{tabular}[c]{@{}c@{}}Avg. No. of\\ Candidates\\ per word\end{tabular}} &
  \textbf{\begin{tabular}[c]{@{}c@{}}Avg. \\ Word\\ Length\end{tabular}} &
  \textbf{\begin{tabular}[c]{@{}c@{}}Query\\ Number\end{tabular}} &
  \textbf{\begin{tabular}[c]{@{}c@{}}Avg.\\ Sentence\\ Length\end{tabular}} \\ \midrule
\multirow{3}{*}{\textbf{as}} & \textbf{Rand}  & \multirow{3}{*}{0.742} & 0.225 & 8.541  & 0.958 & 0.929 & 0.975 & -     & 4.154 & 5.066 & 25.656 & \multirow{3}{*}{24.003} \\ \cmidrule(lr){2-2} \cmidrule(lr){4-12}
                             & \textbf{Phono} &                        & 0.294 & 8.016  & 0.965 & 0.933 & 0.977 & 0.878 & 3.226 & 5.086 & 23.326 &                         \\ \cmidrule(lr){2-2} \cmidrule(lr){4-12}
                             & \textbf{Ortho} &                        & 0.358 & 7.161  & 0.968 & 0.938 & 0.978 & -     & 1.164 & 5.106 & 18.31  &                         \\ \midrule
\multirow{3}{*}{\textbf{bd}} & \textbf{Rand}  & \multirow{3}{*}{0.481} & 0.157 & 7.412  & 0.978 & 0.945 & 0.985 & -     & 5.616 & 6.009 & 16.224 & \multirow{3}{*}{22.205} \\ \cmidrule(lr){2-2} \cmidrule(lr){4-12}
                             & \textbf{Phono} &                        & 0.199 & 6.678  & 0.983 & 0.947 & 0.988 & -     & 3.98  & 5.998 & 12.55  &                         \\ \cmidrule(lr){2-2} \cmidrule(lr){4-12}
                             & \textbf{Ortho} &                        & 0.34  & 3.579  & 0.988 & 0.969 & 0.992 & 0     & 1.073 & 5.924 & 5.481  &                         \\ \midrule
\multirow{3}{*}{\textbf{bn}} & \textbf{Rand}  & \multirow{3}{*}{0.863} & 0.158 & 10.648 & 0.944 & 0.915 & 0.964 & -     & 4.477 & 4.899 & 33.325 & \multirow{3}{*}{23.721} \\ \cmidrule(lr){2-2} \cmidrule(lr){4-12}
                             & \textbf{Phono} &                        & 0.23  & 11.438 & 0.944 & 0.91  & 0.964 & 0.955 & 3.245 & 4.936 & 31.408 &                         \\ \cmidrule(lr){2-2} \cmidrule(lr){4-12}
                             & \textbf{Ortho} &                        & 0.316 & 11.129 & 0.946 & 0.911 & 0.965 & -     & 1.19  & 4.982 & 25.362 &                         \\ \midrule
\multirow{3}{*}{\textbf{gu}} & \textbf{Rand}  & \multirow{3}{*}{0.882} & 0.108 & 10.828 & 0.942 & 0.906 & 0.963 & -     & 4.099 & 4.58  & 36.327 & \multirow{3}{*}{26.129} \\ \cmidrule(lr){2-2} \cmidrule(lr){4-12}
                             & \textbf{Phono} &                        & 0.157 & 11.889 & 0.939 & 0.901 & 0.964 & 0.966 & 2.972 & 4.589 & 34.81  &                         \\ \cmidrule(lr){2-2} \cmidrule(lr){4-12}
                             & \textbf{Ortho} &                        & 0.345 & 11.344 & 0.937 & 0.9   & 0.961 & -     & 0.987 & 4.644 & 28.433 &                         \\ \midrule
\multirow{3}{*}{\textbf{hi}} & \textbf{Rand}  & \multirow{3}{*}{0.886} & 0.113 & 9.478  & 0.944 & 0.906 & 0.963 & -     & 3.698 & 4.13  & 39.355 & \multirow{3}{*}{30.107} \\ \cmidrule(lr){2-2} \cmidrule(lr){4-12}
                             & \textbf{Phono} &                        & 0.148 & 9.944  & 0.951 & 0.901 & 0.966 & 0.975 & 2.684 & 4.077 & 37.241 &                         \\ \cmidrule(lr){2-2} \cmidrule(lr){4-12}
                             & \textbf{Ortho} &                        & 0.247 & 10.171 & 0.946 & 0.9   & 0.963 & -     & 1.057 & 4.052 & 32.866 &                         \\ \midrule
\multirow{3}{*}{\textbf{kn}} & \textbf{Rand}  & \multirow{3}{*}{0.843} & 0.229 & 9.951  & 0.96  & 0.942 & 0.969 & -     & 6.196 & 7.824 & 29.52  & \multirow{3}{*}{20.034} \\ \cmidrule(lr){2-2} \cmidrule(lr){4-12}
                             & \textbf{Phono} &                        & 0.295 & 9.715  & 0.963 & 0.943 & 0.972 & 0.98  & 4.28  & 7.75  & 26.087 &                         \\ \cmidrule(lr){2-2} \cmidrule(lr){4-12}
                             & \textbf{Ortho} &                        & 0.472 & 9.521  & 0.97  & 0.945 & 0.972 & -     & 1.3   & 7.661 & 19.831 &                         \\ \midrule
\multirow{3}{*}{\textbf{ml}} & \textbf{Rand}  & \multirow{3}{*}{0.868} & 0.199 & 11.799 & 0.959 & 0.938 & 0.966 & -     & 6.37  & 8.754 & 31.368 & \multirow{3}{*}{19.254} \\ \cmidrule(lr){2-2} \cmidrule(lr){4-12}
                             & \textbf{Phono} &                        & 0.319 & 11.735 & 0.966 & 0.939 & 0.968 & 0.946 & 4.031 & 8.649 & 26.866 &                         \\ \cmidrule(lr){2-2} \cmidrule(lr){4-12}
                             & \textbf{Ortho} &                        & 0.51  & 11.283 & 0.967 & 0.941 & 0.969 & -     & 1.496 & 8.571 & 19.288 &                         \\ \midrule
\multirow{3}{*}{\textbf{mr}} & \textbf{Rand}  & \multirow{3}{*}{0.866} & 0.14  & 12.045 & 0.944 & 0.91  & 0.963 & -     & 4.849 & 5.432 & 35.102 & \multirow{3}{*}{23.172} \\ \cmidrule(lr){2-2} \cmidrule(lr){4-12}
                             & \textbf{Phono} &                        & 0.209 & 12.153 & 0.953 & 0.91  & 0.966 & 0.965 & 3.456 & 5.456 & 31.128 &                         \\ \cmidrule(lr){2-2} \cmidrule(lr){4-12}
                             & \textbf{Ortho} &                        & 0.283 & 11.559 & 0.948 & 0.911 & 0.965 & -     & 1.274 & 5.504 & 24.98  &                         \\ \midrule
\multirow{3}{*}{\textbf{or}} & \textbf{Rand}  & \multirow{3}{*}{0.752} & 0.297 & 7.312  & 0.962 & 0.942 & 0.978 & -     & 4.555 & 5.209 & 25.642 & \multirow{3}{*}{23.491} \\ \cmidrule(lr){2-2} \cmidrule(lr){4-12}
                             & \textbf{Phono} &                        & 0.309 & 7.478  & 0.965 & 0.943 & 0.979 & 0.959 & 3.331 & 5.23  & 23.52  &                         \\ \cmidrule(lr){2-2} \cmidrule(lr){4-12}
                             & \textbf{Ortho} &                        & 0.457 & 6.242  & 0.975 & 0.951 & 0.982 & -     & 1.124 & 5.298 & 17.825 &                         \\ \midrule
\multirow{3}{*}{\textbf{pa}} & \textbf{Rand}  & \multirow{3}{*}{0.854} & 0.148 & 9.079  & 0.947 & 0.91  & 0.965 & -     & 3.593 & 4.048 & 38.12  & \multirow{3}{*}{30.062} \\ \cmidrule(lr){2-2} \cmidrule(lr){4-12}
                             & \textbf{Phono} &                        & 0.192 & 9.872  & 0.948 & 0.906 & 0.965 & 0.931 & 2.637 & 4.024 & 36.872 &                         \\ \cmidrule(lr){2-2} \cmidrule(lr){4-12}
                             & \textbf{Ortho} &                        & 0.295 & 8.983  & 0.952 & 0.912 & 0.965 & -     & 0.922 & 4.042 & 30.303 &                         \\ \midrule
\multirow{3}{*}{\textbf{ta}} & \textbf{Rand}  & \multirow{3}{*}{0.866} & 0.138 & 12.43  & 0.95  & 0.933 & 0.966 & -     & 5.877 & 8.017 & 33.05  & \multirow{3}{*}{20.874} \\ \cmidrule(lr){2-2} \cmidrule(lr){4-12}
                             & \textbf{Phono} &                        & 0.264 & 13.679 & 0.954 & 0.929 & 0.967 & 0.965 & 4.141 & 7.879 & 26.695 &                         \\ \cmidrule(lr){2-2} \cmidrule(lr){4-12}
                             & \textbf{Ortho} &                        & 0.48  & 12.323 & 0.965 & 0.935 & 0.968 & -     & 1.332 & 7.815 & 21.765 &                         \\ \midrule
\multirow{3}{*}{\textbf{te}} & \textbf{Rand}  & \multirow{3}{*}{0.862} & 0.198 & 11.263 & 0.956 & 0.928 & 0.967 & -     & 5.727 & 6.941 & 33.754 & \multirow{3}{*}{21.76}  \\ \cmidrule(lr){2-2} \cmidrule(lr){4-12}
                             & \textbf{Phono} &                        & 0.227 & 11.448 & 0.958 & 0.929 & 0.969 & 0.967 & 4.143 & 6.848 & 30.561 &                         \\ \cmidrule(lr){2-2} \cmidrule(lr){4-12}
                             & \textbf{Ortho} &                        & 0.377 & 11.075 & 0.959 & 0.928 & 0.967 & -     & 1.266 & 6.829 & 22.748 &                         \\ \bottomrule
\end{tabular}%
}
\caption{The table presents the impact of different character-substitution attack strategies: random (\textbf{Rand}), phonetic (\textbf{Phono}), and orthographic (\textbf{Ortho}) on the \textit{XLMR} language model for the \textit{IndicSentiment} dataset.}
\label{tab:senti_xlmr}
\end{table*}

\begin{table*}[]
\centering
\resizebox{\textwidth}{!}{%
\begin{tabular}{@{}ccccccccccccc@{}}
\toprule
\textbf{Lang} &
  \textbf{\begin{tabular}[c]{@{}c@{}}Type\\ of\\ Perturbation\end{tabular}} &
  \textbf{\begin{tabular}[c]{@{}c@{}}Original\\ Accuracy\end{tabular}} &
  \textbf{\begin{tabular}[c]{@{}c@{}}After\\ Attack\\ Accuracy\end{tabular}} &
  \textbf{\begin{tabular}[c]{@{}c@{}}\%\\ Perturbed\\ Words\end{tabular}} &
  \textbf{\begin{tabular}[c]{@{}c@{}}Semantic\\ Similarity\end{tabular}} &
  \textbf{\begin{tabular}[c]{@{}c@{}}Overlap\\ Similarity\end{tabular}} &
  \textbf{\begin{tabular}[c]{@{}c@{}}BERTScore\\ based\\ Similarity\end{tabular}} &
  \textbf{\begin{tabular}[c]{@{}c@{}}Phonetic\\ Similarity\end{tabular}} &
  \textbf{\begin{tabular}[c]{@{}c@{}}Avg. No. of\\ Candidates\\ per word\end{tabular}} &
  \textbf{\begin{tabular}[c]{@{}c@{}}Avg. \\ Word \\ Length\end{tabular}} &
  \textbf{\begin{tabular}[c]{@{}c@{}}Query\\ Number\end{tabular}} &
  \textbf{\begin{tabular}[c]{@{}c@{}}Avg.\\ Sentence\\ Length\end{tabular}} \\ \midrule
\multirow{3}{*}{\textbf{as}} &
  \textbf{Rand} &
  \multirow{3}{*}{0.526} &
  0.075/0.074 &
  8.07/8.589 &
  0.972/0.97 &
  0.944/0.94 &
  0.979/0.978 &
  - &
  4.821/4.827 &
  6.028/6.045 &
  15.405/15.748 &
  \multirow{3}{*}{17.814/17.628} \\ \cmidrule(lr){2-2} \cmidrule(lr){4-12}
 &
  \textbf{Phono} &
   &
  0.094/0.094 &
  8.807/8.891 &
  0.972/0.971 &
  0.94/0.938 &
  0.979/0.979 &
  0.895/0.896 &
  3.729/3.765 &
  5.967/6.013 &
  14.666/14.65 &
   \\ \cmidrule(lr){2-2} \cmidrule(lr){4-12}
 &
  \textbf{Ortho} &
   &
  0.101/0.108 &
  8.728/8.71 &
  0.97/0.969 &
  0.936/0.936 &
  0.978/0.978 &
  - &
  2.518/2.53 &
  5.898/5.972 &
  12.899/12.733 &
   \\ \midrule
\multirow{3}{*}{\textbf{bn}} &
  \textbf{Rand} &
  \multirow{3}{*}{0.51} &
  0.037/0.03 &
  9.034/8.757 &
  0.963/0.965 &
  0.939/0.94 &
  0.975/0.975 &
  - &
  5.564/5.55 &
  6.28/6.24 &
  13.942/13.805 &
  \multirow{3}{*}{15.031/15.299} \\ \cmidrule(lr){2-2} \cmidrule(lr){4-12}
 &
  \textbf{Phono} &
   &
  0.06/0.053 &
  9.661/9.369 &
  0.965/0.964 &
  0.935/0.936 &
  0.975/0.975 &
  0.979/0.976 &
  3.905/3.897 &
  6.205/6.213 &
  12.585/12.51 &
   \\ \cmidrule(lr){2-2} \cmidrule(lr){4-12}
 &
  \textbf{Ortho} &
   &
  0.08/0.075 &
  9.427/9.286 &
  0.963/0.963 &
  0.932/0.932 &
  0.974/0.974 &
  - &
  2.462/2.459 &
  6.13/6.111 &
  10.627/10.742 &
   \\ \midrule
\multirow{3}{*}{\textbf{gu}} &
  \textbf{Rand} &
  \multirow{3}{*}{0.806} &
  0.239/0.181 &
  12.348/12.61 &
  0.953/0.95 &
  0.909/0.908 &
  0.967/0.966 &
  - &
  4.97/5.001 &
  5.576/5.632 &
  22.709/23.217 &
  \multirow{3}{*}{16.53/16.594} \\ \cmidrule(lr){2-2} \cmidrule(lr){4-12}
 &
  \textbf{Phono} &
   &
  0.265/0.227 &
  13.267/13.53 &
  0.956/0.953 &
  0.904/0.902 &
  0.968/0.967 &
  0.97/0.969 &
  3.396/3.399 &
  5.572/5.591 &
  20.721/20.901 &
   \\ \cmidrule(lr){2-2} \cmidrule(lr){4-12}
 &
  \textbf{Ortho} &
   &
  0.299/0.268 &
  13.291/13.429 &
  0.945/0.943 &
  0.898/0.898 &
  0.962/0.962 &
  - &
  1.872/1.883 &
  5.481/5.475 &
  17.798/17.932 &
   \\ \midrule
\multirow{3}{*}{\textbf{hi}} &
  \textbf{Rand} &
  \multirow{3}{*}{0.498} &
  0.015/0.012 &
  6.711/6.227 &
  0.967/0.968 &
  0.942/0.946 &
  0.976/0.977 &
  - &
  4.659/4.646 &
  5.283/5.271 &
  15.15/14.851 &
  \multirow{3}{*}{19.749/20.201} \\ \cmidrule(lr){2-2} \cmidrule(lr){4-12}
 &
  \textbf{Phono} &
   &
  0.029/0.023 &
  7.556/7.238 &
  0.967/0.967 &
  0.935/0.937 &
  0.976/0.976 &
  0.98/0.981 &
  3.188/3.209 &
  5.205/5.217 &
  14.313/14.308 &
   \\ \cmidrule(lr){2-2} \cmidrule(lr){4-12}
 &
  \textbf{Ortho} &
   &
  0.03/0.03 &
  7.364/6.826 &
  0.962/0.963 &
  0.934/0.939 &
  0.975/0.976 &
  - &
  2.352/2.327 &
  5.144/5.125 &
  13.201/12.914 &
   \\ \midrule
\multirow{3}{*}{\textbf{kn}} &
  \textbf{Rand} &
  \multirow{3}{*}{0.549} &
  0.1/0.094 &
  9.535/9.231 &
  0.972/0.972 &
  0.947/0.95 &
  0.977/0.978 &
  - &
  5.664/5.701 &
  7.133/7.176 &
  14.341/13.628 &
  \multirow{3}{*}{14.402/14.236} \\ \cmidrule(lr){2-2} \cmidrule(lr){4-12}
 &
  \textbf{Phono} &
   &
  0.128/0.113 &
  9.725/10.033 &
  0.974/0.975 &
  0.946/0.946 &
  0.978/0.978 &
  0.98/0.979 &
  3.851/3.835 &
  7.12/7.137 &
  12.373/12.4 &
   \\ \cmidrule(lr){2-2} \cmidrule(lr){4-12}
 &
  \textbf{Ortho} &
   &
  0.177/0.164 &
  9.156/8.981 &
  0.973/0.974 &
  0.946/0.947 &
  0.977/0.978 &
  - &
  2.132/2.133 &
  7.052/7.073 &
  9.599/9.443 &
   \\ \midrule
\multirow{3}{*}{\textbf{ml}} &
  \textbf{Rand} &
  \multirow{3}{*}{0.578} &
  0.101/0.097 &
  10.95/10.519 &
  0.968/0.97 &
  0.946/0.949 &
  0.974/0.975 &
  - &
  6.024/5.996 &
  8.317/8.295 &
  14.907/14.45 &
  \multirow{3}{*}{13.238/13.265} \\ \cmidrule(lr){2-2} \cmidrule(lr){4-12}
 &
  \textbf{Phono} &
   &
  0.154/0.127 &
  10.906/11.288 &
  0.973/0.972 &
  0.946/0.945 &
  0.976/0.975 &
  0.95/0.952 &
  4.115/4.094 &
  8.253/8.241 &
  12.578/12.735 &
   \\ \cmidrule(lr){2-2} \cmidrule(lr){4-12}
 &
  \textbf{Ortho} &
   &
  0.158/0.148 &
  10.72/11.078 &
  0.967/0.966 &
  0.944/0.944 &
  0.974/0.974 &
  - &
  3.065/3.045 &
  8.203/8.17 &
  10.92/11.079 &
   \\ \midrule
\multirow{3}{*}{\textbf{mr}} &
  \textbf{Rand} &
  \multirow{3}{*}{0.531} &
  0.049/0.046 &
  8.035/7.509 &
  0.971/0.973 &
  0.948/0.952 &
  0.977/0.978 &
  - &
  5.614/5.684 &
  6.365/6.434 &
  14.284/13.813 &
  \multirow{3}{*}{15.938/16.051} \\ \cmidrule(lr){2-2} \cmidrule(lr){4-12}
 &
  \textbf{Phono} &
   &
  0.082/0.065 &
  8.551/8.449 &
  0.974/0.974 &
  0.945/0.947 &
  0.978/0.978 &
  0.979/0.98 &
  3.887/3.875 &
  6.335/6.349 &
  12.934/13.126 &
   \\ \cmidrule(lr){2-2} \cmidrule(lr){4-12}
 &
  \textbf{Ortho} &
   &
  0.075/0.065 &
  8.441/8.157 &
  0.97/0.971 &
  0.944/0.946 &
  0.976/0.977 &
  - &
  2.915/2.915 &
  6.314/6.314 &
  11.746/11.876 &
   \\ \midrule
\multirow{3}{*}{\textbf{or}} &
  \textbf{Rand} &
  \multirow{3}{*}{0.556} &
  0.073/0.065 &
  9.841/9.395 &
  0.966/0.967 &
  0.936/0.941 &
  0.975/0.976 &
  - &
  5.53/5.461 &
  6.329/6.249 &
  16.363/15.896 &
  \multirow{3}{*}{15.884/16.035} \\ \cmidrule(lr){2-2} \cmidrule(lr){4-12}
 &
  \textbf{Phono} &
   &
  0.096/0.084 &
  10.02/9.81 &
  0.969/0.969 &
  0.935/0.938 &
  0.976/0.976 &
  0.962/0.963 &
  3.865/3.813 &
  6.267/6.205 &
  14.393/14.103 &
   \\ \cmidrule(lr){2-2} \cmidrule(lr){4-12}
 &
  \textbf{Ortho} &
   &
  0.105/0.1 &
  10.449/9.826 &
  0.963/0.963 &
  0.928/0.934 &
  0.974/0.975 &
  - &
  2.573/2.534 &
  6.222/6.152 &
  12.917/12.766 &
   \\ \midrule
\multirow{3}{*}{\textbf{pa}} &
  \textbf{Rand} &
  \multirow{3}{*}{0.591} &
  0.075/0.063 &
  7.894/7.866 &
  0.957/0.958 &
  0.928/0.929 &
  0.974/0.974 &
  - &
  4.249/4.213 &
  4.748/4.714 &
  17.844/18.461 &
  \multirow{3}{*}{20.124/20.372} \\ \cmidrule(lr){2-2} \cmidrule(lr){4-12}
 &
  \textbf{Phono} &
   &
  0.089/0.072 &
  8.914/8.555 &
  0.956/0.957 &
  0.919/0.923 &
  0.973/0.974 &
  0.938/0.936 &
  3.043/3.034 &
  4.701/4.706 &
  17.217/17.455 &
   \\ \cmidrule(lr){2-2} \cmidrule(lr){4-12}
 &
  \textbf{Ortho} &
   &
  0.092/0.081 &
  8.499/8.332 &
  0.953/0.953 &
  0.923/0.925 &
  0.973/0.973 &
  - &
  2.087/2.064 &
  4.668/4.633 &
  15.329/15.502 &
   \\ \midrule
\multirow{3}{*}{\textbf{te}} &
  \textbf{Rand} &
  \multirow{3}{*}{0.544} &
  0.092/0.082 &
  9.199/8.984 &
  0.971/0.971 &
  0.947/0.948 &
  0.978/0.979 &
  - &
  5.761/5.709 &
  7.101/7.042 &
  13.199/12.894 &
  \multirow{3}{*}{13.726/13.731} \\ \cmidrule(lr){2-2} \cmidrule(lr){4-12}
 &
  \textbf{Phono} &
   &
  0.125/0.114 &
  9.474/9.654 &
  0.975/0.974 &
  0.946/0.945 &
  0.979/0.979 &
  0.982/0.981 &
  3.957/3.933 &
  7.054/7.024 &
  11.582/11.574 &
   \\ \cmidrule(lr){2-2} \cmidrule(lr){4-12}
 &
  \textbf{Ortho} &
   &
  0.137/0.128 &
  9.54/9.653 &
  0.97/0.97 &
  0.941/0.94 &
  0.976/0.977 &
  - &
  2.277/2.279 &
  7.04/6.999 &
  9.744/9.866 &
   \\ \bottomrule
\end{tabular}%
}
\caption{The table presents the impact of different character-substitution attack strategies: random (\textbf{Rand}), phonetic (\textbf{Phono}), and orthographic (\textbf{Ortho}) on the \textit{XLMR} language model for the \textit{IndicParaphrase} dataset. Results are shown for perturbing both \textit{sentence1} and \textit{sentence2} of the IndicParaphrase dataset, separated by a forward slash (/).}
\label{tab:para_xlmr}
\end{table*}

\begin{table*}[]
\centering
\resizebox{\textwidth}{!}{%
\begin{tabular}{@{}ccccccccccccc@{}}
\toprule
\textbf{Lang} &
  \textbf{\begin{tabular}[c]{@{}c@{}}Type\\ of\\ Perturbation\end{tabular}} &
  \textbf{\begin{tabular}[c]{@{}c@{}}Original\\ Accuracy\end{tabular}} &
  \textbf{\begin{tabular}[c]{@{}c@{}}After\\ Attack\\ Accuracy\end{tabular}} &
  \textbf{\begin{tabular}[c]{@{}c@{}}\%\\ Perturbed\\ Words\end{tabular}} &
  \textbf{\begin{tabular}[c]{@{}c@{}}Semantic\\ Similarity\end{tabular}} &
  \textbf{\begin{tabular}[c]{@{}c@{}}Overlap\\ Similarity\end{tabular}} &
  \textbf{\begin{tabular}[c]{@{}c@{}}BERTScore\\ based\\ Similarity\end{tabular}} &
  \textbf{\begin{tabular}[c]{@{}c@{}}Phonetic\\ Similarity\end{tabular}} &
  \textbf{\begin{tabular}[c]{@{}c@{}}Avg. No. of\\ Candidates\\ per word\end{tabular}} &
  \textbf{\begin{tabular}[c]{@{}c@{}}Avg.\\ Word\\ Length\end{tabular}} &
  \textbf{\begin{tabular}[c]{@{}c@{}}Query\\ Number\end{tabular}} &
  \textbf{\begin{tabular}[c]{@{}c@{}}Avg.\\ Sentence\\ Length\end{tabular}} \\ \midrule
\multirow{3}{*}{\textbf{as}} &
  \textbf{Rand} &
  \multirow{3}{*}{0.638} &
  0.193/0.084 &
  8.182/11.156 &
  0.955/0.93 &
  0.929/0.908 &
  0.975/0.971 &
  - &
  4.114/4.306 &
  5.146/5.246 &
  15.576/9.282 &
  \multirow{3}{*}{17.091/9.459} \\ \cmidrule(lr){2-2} \cmidrule(lr){4-12}
 &
  \textbf{Phono} &
   &
  0.23/0.135 &
  8.24/11.405 &
  0.961/0.942 &
  0.929/0.908 &
  0.977/0.973 &
  0.889/0.893 &
  3.255/3.331 &
  5.102/5.212 &
  14.451/8.688 &
   \\ \cmidrule(lr){2-2} \cmidrule(lr){4-12}
 &
  \textbf{Ortho} &
   &
  0.274/0.201 &
  7.927/11.016 &
  0.96/0.945 &
  0.931/0.91 &
  0.977/0.972 &
  - &
  2.173/2.097 &
  5.084/5.117 &
  12.843/7.798 &
   \\ \midrule
\multirow{3}{*}{\textbf{bn}} &
  \textbf{Rand} &
  \multirow{3}{*}{0.706} &
  0.195/0.068 &
  9.087/12.369 &
  0.939/0.918 &
  0.917/0.894 &
  0.967/0.962 &
  - &
  4.464/4.466 &
  4.935/4.911 &
  17.852/10.022 &
  \multirow{3}{*}{16.971/9.221} \\ \cmidrule(lr){2-2} \cmidrule(lr){4-12}
 &
  \textbf{Phono} &
   &
  0.235/0.125 &
  9.414/13.606 &
  0.944/0.922 &
  0.915/0.888 &
  0.968/0.962 &
  0.973/0.977 &
  3.237/3.221 &
  4.904/4.911 &
  16.58/9.662 &
   \\ \cmidrule(lr){2-2} \cmidrule(lr){4-12}
 &
  \textbf{Ortho} &
   &
  0.288/0.204 &
  9.001/13.116 &
  0.942/0.921 &
  0.916/0.89 &
  0.968/0.963 &
  - &
  2.045/1.938 &
  4.881/4.861 &
  14.449/8.72 &
   \\ \midrule
\multirow{3}{*}{\textbf{gu}} &
  \textbf{Rand} &
  \multirow{3}{*}{0.702} &
  0.191/0.098 &
  9.241/12.251 &
  0.937/0.917 &
  0.909/0.889 &
  0.968/0.964 &
  - &
  4.022/4.036 &
  4.496/4.54 &
  19.293/10.884 &
  \multirow{3}{*}{18.389/9.956} \\ \cmidrule(lr){2-2} \cmidrule(lr){4-12}
 &
  \textbf{Phono} &
   &
  0.237/0.144 &
  8.845/12.838 &
  0.944/0.924 &
  0.913/0.887 &
  0.971/0.966 &
  0.974/0.979 &
  2.833/2.836 &
  4.464/4.524 &
  16.951/10.083 &
   \\ \cmidrule(lr){2-2} \cmidrule(lr){4-12}
 &
  \textbf{Ortho} &
   &
  0.332/0.279 &
  7.662/11.221 &
  0.948/0.929 &
  0.923/0.896 &
  0.971/0.967 &
  - &
  1.407/1.419 &
  4.39/4.437 &
  14.396/8.362 &
   \\ \midrule
\multirow{3}{*}{\textbf{hi}} &
  \textbf{Rand} &
  \multirow{3}{*}{0.733} &
  0.179/0.053 &
  8.096/10.888 &
  0.939/0.913 &
  0.909/0.883 &
  0.966/0.961 &
  - &
  3.5/3.635 &
  3.898/4.113 &
  22.085/11.968 &
  \multirow{3}{*}{21.825/11.492} \\ \cmidrule(lr){2-2} \cmidrule(lr){4-12}
 &
  \textbf{Phono} &
   &
  0.238/0.139 &
  8.756/12.049 &
  0.944/0.919 &
  0.906/0.876 &
  0.968/0.963 &
  0.975/0.979 &
  2.466/2.573 &
  3.834/4.035 &
  21.264/11.497 &
   \\ \cmidrule(lr){2-2} \cmidrule(lr){4-12}
 &
  \textbf{Ortho} &
   &
  0.249/0.109 &
  7.935/10.944 &
  0.939/0.916 &
  0.911/0.882 &
  0.968/0.961 &
  - &
  1.689/1.741 &
  3.829/4.023 &
  18.87/10.53 &
   \\ \midrule
\multirow{3}{*}{\textbf{kn}} &
  \textbf{Rand} &
  \multirow{3}{*}{0.707} &
  0.221/0.122 &
  10.238/14.904 &
  0.948/0.931 &
  0.932/0.912 &
  0.97/0.966 &
  - &
  5.953/6.552 &
  7.479/8.196 &
  17.399/10.415 &
  \multirow{3}{*}{14.072/7.533} \\ \cmidrule(lr){2-2} \cmidrule(lr){4-12}
 &
  \textbf{Phono} &
   &
  0.282/0.207 &
  10.069/15.281 &
  0.958/0.937 &
  0.934/0.911 &
  0.973/0.968 &
  0.984/0.984 &
  3.989/4.29 &
  7.376/7.945 &
  14.814/9.005 &
   \\ \cmidrule(lr){2-2} \cmidrule(lr){4-12}
 &
  \textbf{Ortho} &
   &
  0.411/0.369 &
  8.141/12.721 &
  0.965/0.948 &
  0.942/0.923 &
  0.975/0.972 &
  - &
  2.035/2.117 &
  7.246/7.681 &
  10.279/6.27 &
   \\ \midrule
\multirow{3}{*}{\textbf{ml}} &
  \textbf{Rand} &
  \multirow{3}{*}{0.719} &
  0.255/0.127 &
  10.246/16.184 &
  0.951/0.932 &
  0.936/0.916 &
  0.968/0.963 &
  - &
  6.077/7.041 &
  8.561/9.655 &
  15.83/10.237 &
  \multirow{3}{*}{12.98/6.837} \\ \cmidrule(lr){2-2} \cmidrule(lr){4-12}
 &
  \textbf{Phono} &
   &
  0.317/0.252 &
  10.519/16.006 &
  0.958/0.943 &
  0.936/0.918 &
  0.971/0.967 &
  0.953/0.952 &
  4.029/4.347 &
  8.548/9.322 &
  13.868/8.293 &
   \\ \cmidrule(lr){2-2} \cmidrule(lr){4-12}
 &
  \textbf{Ortho} &
   &
  0.364/0.308 &
  9.242/14.869 &
  0.953/0.934 &
  0.941/0.921 &
  0.971/0.966 &
  - &
  3.094/3.259 &
  8.482/9.129 &
  11.12/6.781 &
   \\ \midrule
\multirow{3}{*}{\textbf{mr}} &
  \textbf{Rand} &
  \multirow{3}{*}{0.687} &
  0.19/0.084 &
  9.256/12.783 &
  0.939/0.92 &
  0.918/0.897 &
  0.968/0.964 &
  - &
  4.738/4.888 &
  5.35/5.453 &
  16.723/9.637 &
  \multirow{3}{*}{16.154/8.773} \\ \cmidrule(lr){2-2} \cmidrule(lr){4-12}
 &
  \textbf{Phono} &
   &
  0.248/0.175 &
  9.592/13.816 &
  0.948/0.93 &
  0.919/0.894 &
  0.971/0.967 &
  0.976/0.977 &
  3.222/3.317 &
  5.303/5.393 &
  15.022/9.135 &
   \\ \cmidrule(lr){2-2} \cmidrule(lr){4-12}
 &
  \textbf{Ortho} &
   &
  0.241/0.131 &
  8.988/13 &
  0.944/0.92 &
  0.92/0.896 &
  0.969/0.965 &
  - &
  2.492/2.508 &
  5.305/5.411 &
  13.83/8.328 &
   \\ \midrule
\multirow{3}{*}{\textbf{or}} &
  \textbf{Rand} &
  \multirow{3}{*}{0.687} &
  0.195/0.101 &
  8.969/12.591 &
  0.946/0.928 &
  0.921/0.901 &
  0.971/0.967 &
  - &
  4.831/4.918 &
  5.456/5.54 &
  16.794/9.84 &
  \multirow{3}{*}{16.173/8.847} \\ \cmidrule(lr){2-2} \cmidrule(lr){4-12}
 &
  \textbf{Phono} &
   &
  0.236/0.154 &
  9.158/13.213 &
  0.948/0.928 &
  0.921/0.899 &
  0.972/0.967 &
  0.966/0.969 &
  3.508/3.573 &
  5.382/5.483 &
  15.434/9.245 &
   \\ \cmidrule(lr){2-2} \cmidrule(lr){4-12}
 &
  \textbf{Ortho} &
   &
  0.317/0.279 &
  8.383/12.084 &
  0.951/0.931 &
  0.927/0.905 &
  0.974/0.969 &
  - &
  1.969/1.9 &
  5.332/5.415 &
  13.279/7.738 &
   \\ \midrule
\multirow{3}{*}{\textbf{pa}} &
  \textbf{Rand} &
  \multirow{3}{*}{0.697} &
  0.182/0.07 &
  8.001/10.623 &
  0.943/0.916 &
  0.914/0.885 &
  0.969/0.964 &
  - &
  3.491/3.523 &
  3.936/4.021 &
  21.694/11.593 &
  \multirow{3}{*}{21.909/11.725} \\ \cmidrule(lr){2-2} \cmidrule(lr){4-12}
 &
  \textbf{Phono} &
   &
  0.221/0.121 &
  8.33/11.715 &
  0.947/0.918 &
  0.911/0.878 &
  0.97/0.965 &
  0.941/0.94 &
  2.576/2.556 &
  3.905/4.01 &
  20.073/11.115 &
   \\ \cmidrule(lr){2-2} \cmidrule(lr){4-12}
 &
  \textbf{Ortho} &
   &
  0.246/0.124 &
  7.537/10.634 &
  0.948/0.922 &
  0.918/0.886 &
  0.97/0.964 &
  - &
  1.656/1.677 &
  3.887/3.927 &
  18.128/10.181 &
   \\ \midrule
\multirow{3}{*}{\textbf{ta}} &
  \textbf{Rand} &
  \multirow{3}{*}{0.701} &
  0.235/0.122 &
  9.799/14.758 &
  0.944/0.926 &
  0.937/0.918 &
  0.97/0.966 &
  - &
  5.42/5.926 &
  7.78/8.271 &
  16.792/10.255 &
  \multirow{3}{*}{14.721/7.826} \\ \cmidrule(lr){2-2} \cmidrule(lr){4-12}
 &
  \textbf{Phono} &
   &
  0.344/0.264 &
  9.221/14.761 &
  0.953/0.932 &
  0.941/0.92 &
  0.974/0.969 &
  0.974/0.979 &
  3.796/3.977 &
  7.698/8.038 &
  12.116/7.317 &
   \\ \cmidrule(lr){2-2} \cmidrule(lr){4-12}
 &
  \textbf{Ortho} &
   &
  0.409/0.356 &
  7.861/13.033 &
  0.963/0.945 &
  0.947/0.925 &
  0.975/0.97 &
  - &
  2.097/2.207 &
  7.619/7.96 &
  10.769/6.718 &
   \\ \midrule
\multirow{3}{*}{\textbf{te}} &
  \textbf{Rand} &
  \multirow{3}{*}{0.695} &
  0.201/0.124 &
  10.009/14.78 &
  0.947/0.927 &
  0.925/0.902 &
  0.97/0.966 &
  - &
  5.269/5.696 &
  6.396/6.858 &
  16.693/9.778 &
  \multirow{3}{*}{14.655/7.706} \\ \cmidrule(lr){2-2} \cmidrule(lr){4-12}
 &
  \textbf{Phono} &
   &
  0.249/0.155 &
  9.681/14.813 &
  0.954/0.934 &
  0.929/0.903 &
  0.973/0.969 &
  0.974/0.98 &
  3.68/4.09 &
  6.365/6.859 &
  14.441/8.91 &
   \\ \cmidrule(lr){2-2} \cmidrule(lr){4-12}
 &
  \textbf{Ortho} &
   &
  0.335/0.263 &
  8.573/13.787 &
  0.957/0.934 &
  0.935/0.908 &
  0.973/0.968 &
  - &
  1.944/2.101 &
  6.331/6.732 &
  11.48/6.925 &
   \\ \bottomrule
\end{tabular}%
}
\caption{The table presents the impact of different character-substitution attack strategies: random (\textbf{Rand}), phonetic (\textbf{Phono}), and orthographic (\textbf{Ortho}) on the \textit{XLMR} language model for the \textit{IndicXNLI} dataset. Results are shown for perturbing both \textit{premise} and \textit{hypothesis} of the IndicXNLI dataset, separated by a forward slash (/)}
\label{tab:xnli_xlmr}
\end{table*}

\begin{table*}[]
\centering
\resizebox{\textwidth}{!}{%
\begin{tabular}{@{}ccccccccccccc@{}}
\toprule
\textbf{Lang} &
  \textbf{\begin{tabular}[c]{@{}c@{}}Type\\ of\\ Perturbation\end{tabular}} &
  \textbf{\begin{tabular}[c]{@{}c@{}}Original\\ Accuracy\end{tabular}} &
  \textbf{\begin{tabular}[c]{@{}c@{}}After\\ Attack\\ Accuracy\end{tabular}} &
  \textbf{\begin{tabular}[c]{@{}c@{}}\%\\ Perturbed\\ Words\end{tabular}} &
  \textbf{\begin{tabular}[c]{@{}c@{}}Semantic\\ Similarity\end{tabular}} &
  \textbf{\begin{tabular}[c]{@{}c@{}}Overlap\\ Similarity\end{tabular}} &
  \textbf{\begin{tabular}[c]{@{}c@{}}BERTScore\\ based\\ Similarity\end{tabular}} &
  \textbf{\begin{tabular}[c]{@{}c@{}}Phonetic\\ Similarity\end{tabular}} &
  \textbf{\begin{tabular}[c]{@{}c@{}}Avg. No. of \\ Candidates\\ per word\end{tabular}} &
  \textbf{\begin{tabular}[c]{@{}c@{}}Avg.\\ Word\\ Length\end{tabular}} &
  \textbf{\begin{tabular}[c]{@{}c@{}}Query\\ Number\end{tabular}} &
  \textbf{\begin{tabular}[c]{@{}c@{}}Avg.\\ Sentence\\ Length\end{tabular}} \\ \midrule
\multirow{3}{*}{\textbf{as}} & \textbf{Rand}  & \multirow{3}{*}{0.5}   & 0.154 & 7.224 & 0.964 & 0.939 & 0.981 & -     & 4.106 & 5.085 & 16.533 & \multirow{3}{*}{24.003} \\ \cmidrule(lr){2-2} \cmidrule(lr){4-12}
                             & \textbf{Phono} &                        & 0.188 & 6.696 & 0.971 & 0.944 & 0.983 & 0.881 & 3.196 & 5.046 & 14.038 &                         \\ \cmidrule(lr){2-2} \cmidrule(lr){4-12}
                             & \textbf{Ortho} &                        & 0.182 & 7.257 & 0.967 & 0.938 & 0.981 & -     & 2.308 & 5.046 & 13.474 &                         \\ \midrule
\multirow{3}{*}{\textbf{bd}} & \textbf{Rand}  & \multirow{3}{*}{0.45}  & 0.005 & 4.791 & 0.986 & 0.963 & 0.989 & -     & 5.582 & 6.039 & 14.425 & \multirow{3}{*}{22.205} \\ \cmidrule(lr){2-2} \cmidrule(lr){4-12}
                             & \textbf{Phono} &                        & 0.024 & 5.621 & 0.987 & 0.959 & 0.989 & -     & 3.883 & 5.98  & 13.936 &                         \\ \cmidrule(lr){2-2} \cmidrule(lr){4-12}
                             & \textbf{Ortho} &                        & 0.027 & 5.59  & 0.984 & 0.957 & 0.988 & -     & 2.375 & 5.954 & 12.883 &                         \\ \midrule
\multirow{3}{*}{\textbf{bn}} & \textbf{Rand}  & \multirow{3}{*}{0.639} & 0.08  & 8.509 & 0.955 & 0.929 & 0.972 & -     & 4.528 & 4.953 & 24.239 & \multirow{3}{*}{23.721} \\ \cmidrule(lr){2-2} \cmidrule(lr){4-12}
                             & \textbf{Phono} &                        & 0.128 & 8.35  & 0.96  & 0.932 & 0.975 & 0.959 & 3.293 & 4.988 & 22.376 &                         \\ \cmidrule(lr){2-2} \cmidrule(lr){4-12}
                             & \textbf{Ortho} &                        & 0.171 & 8.227 & 0.962 & 0.93  & 0.974 & -     & 2.292 & 4.974 & 19.556 &                         \\ \midrule
\multirow{3}{*}{\textbf{gu}} & \textbf{Rand}  & \multirow{3}{*}{0.653} & 0.063 & 7.517 & 0.958 & 0.932 & 0.974 & -     & 4.065 & 4.529 & 25.554 & \multirow{3}{*}{26.129} \\ \cmidrule(lr){2-2} \cmidrule(lr){4-12}
                             & \textbf{Phono} &                        & 0.125 & 7.659 & 0.963 & 0.93  & 0.976 & 0.972 & 2.916 & 4.522 & 23.573 &                         \\ \cmidrule(lr){2-2} \cmidrule(lr){4-12}
                             & \textbf{Ortho} &                        & 0.24  & 6.307 & 0.966 & 0.939 & 0.977 & -     & 1.59  & 4.58  & 19.484 &                         \\ \midrule
\multirow{3}{*}{\textbf{hi}} & \textbf{Rand}  & \multirow{3}{*}{0.715} & 0.042 & 8.084 & 0.953 & 0.918 & 0.97  & -     & 3.721 & 4.152 & 31.396 & \multirow{3}{*}{30.107} \\ \cmidrule(lr){2-2} \cmidrule(lr){4-12}
                             & \textbf{Phono} &                        & 0.129 & 8.723 & 0.961 & 0.916 & 0.973 & 0.973 & 2.647 & 4.048 & 30.221 &                         \\ \cmidrule(lr){2-2} \cmidrule(lr){4-12}
                             & \textbf{Ortho} &                        & 0.17  & 7.614 & 0.957 & 0.923 & 0.973 & -     & 1.793 & 4.036 & 27.012 &                         \\ \midrule
\multirow{3}{*}{\textbf{kn}} & \textbf{Rand}  & \multirow{3}{*}{0.702} & 0.123 & 8.501 & 0.966 & 0.947 & 0.974 & -     & 6.283 & 7.881 & 23.667 & \multirow{3}{*}{20.034} \\ \cmidrule(lr){2-2} \cmidrule(lr){4-12}
                             & \textbf{Phono} &                        & 0.228 & 8.258 & 0.97  & 0.949 & 0.977 & 0.98  & 4.197 & 7.682 & 20.554 &                         \\ \cmidrule(lr){2-2} \cmidrule(lr){4-12}
                             & \textbf{Ortho} &                        & 0.375 & 6.69  & 0.977 & 0.957 & 0.979 & -     & 2.053 & 7.576 & 15.024 &                         \\ \midrule
\multirow{3}{*}{\textbf{ml}} & \textbf{Rand}  & \multirow{3}{*}{0.653} & 0.142 & 8.965 & 0.963 & 0.948 & 0.974 & -     & 6.346 & 8.85  & 21.857 & \multirow{3}{*}{19.254} \\ \cmidrule(lr){2-2} \cmidrule(lr){4-12}
                             & \textbf{Phono} &                        & 0.279 & 7.072 & 0.977 & 0.96  & 0.98  & 0.951 & 3.982 & 8.59  & 16.531 &                         \\ \cmidrule(lr){2-2} \cmidrule(lr){4-12}
                             & \textbf{Ortho} &                        & 0.278 & 8.348 & 0.972 & 0.952 & 0.976 & -     & 3.099 & 8.547 & 15.546 &                         \\ \midrule
\multirow{3}{*}{\textbf{mr}} & \textbf{Rand}  & \multirow{3}{*}{0.673} & 0.06  & 8.191 & 0.962 & 0.935 & 0.973 & -     & 4.823 & 5.37  & 23.394 & \multirow{3}{*}{23.172} \\ \cmidrule(lr){2-2} \cmidrule(lr){4-12}
                             & \textbf{Phono} &                        & 0.147 & 8.69  & 0.97  & 0.933 & 0.976 & 0.972 & 3.38  & 5.345 & 21.614 &                         \\ \cmidrule(lr){2-2} \cmidrule(lr){4-12}
                             & \textbf{Ortho} &                        & 0.15  & 8.534 & 0.961 & 0.931 & 0.974 & -     & 2.579 & 5.437 & 20.358 &                         \\ \midrule
\multirow{3}{*}{\textbf{pa}} & \textbf{Rand}  & \multirow{3}{*}{0.664} & 0.05  & 7.288 & 0.959 & 0.925 & 0.973 & -     & 3.568 & 4.061 & 28.775 & \multirow{3}{*}{30.062} \\ \cmidrule(lr){2-2} \cmidrule(lr){4-12}
                             & \textbf{Phono} &                        & 0.091 & 7.595 & 0.958 & 0.922 & 0.973 & 0.935 & 2.618 & 4.035 & 27.119 &                         \\ \cmidrule(lr){2-2} \cmidrule(lr){4-12}
                             & \textbf{Ortho} &                        & 0.177 & 7.244 & 0.958 & 0.926 & 0.972 & -     & 1.723 & 4.023 & 24.219 &                         \\ \midrule
\multirow{3}{*}{\textbf{ta}} & \textbf{Rand}  & \multirow{3}{*}{0.688} & 0.103 & 9.327 & 0.96  & 0.946 & 0.974 & -     & 5.774 & 7.858 & 24.669 & \multirow{3}{*}{20.874} \\ \cmidrule(lr){2-2} \cmidrule(lr){4-12}
                             & \textbf{Phono} &                        & 0.197 & 8.854 & 0.97  & 0.949 & 0.978 & 0.974 & 4.073 & 7.792 & 18.654 &                         \\ \cmidrule(lr){2-2} \cmidrule(lr){4-12}
                             & \textbf{Ortho} &                        & 0.311 & 6.524 & 0.98  & 0.961 & 0.98  & -     & 2.136 & 7.756 & 15.734 &                         \\ \midrule
\multirow{3}{*}{\textbf{te}} & \textbf{Rand}  & \multirow{3}{*}{0.662} & 0.079 & 7.881 & 0.966 & 0.947 & 0.976 & -     & 5.749 & 6.966 & 23.23  & \multirow{3}{*}{21.76}  \\ \cmidrule(lr){2-2} \cmidrule(lr){4-12}
                             & \textbf{Phono} &                        & 0.114 & 7.591 & 0.97  & 0.949 & 0.978 & 0.975 & 4.153 & 6.851 & 20.644 &                         \\ \cmidrule(lr){2-2} \cmidrule(lr){4-12}
                             & \textbf{Ortho} &                        & 0.205 & 6.872 & 0.97  & 0.952 & 0.978 & -     & 2.071 & 6.825 & 17.146 &                         \\ \bottomrule
\end{tabular}%
}
\caption{The table presents the impact of different character-substitution attack strategies: random (\textbf{Rand}), phonetic (\textbf{Phono}), and orthographic (\textbf{Ortho}) on the \textit{mBERT} language model for the \textit{IndicSentiment} dataset.}
\label{tab:senti_mbert}
\end{table*}

\begin{table*}[]
\centering
\resizebox{\textwidth}{!}{%
\begin{tabular}{@{}ccccccccccccc@{}}
\toprule
\textbf{Lang} &
  \textbf{\begin{tabular}[c]{@{}c@{}}Type\\ of\\ Perturbation\end{tabular}} &
  \textbf{\begin{tabular}[c]{@{}c@{}}Original\\ Accuracy\end{tabular}} &
  \textbf{\begin{tabular}[c]{@{}c@{}}After\\ Attack\\ Accuracy\end{tabular}} &
  \textbf{\begin{tabular}[c]{@{}c@{}}\%\\ Perturbed\\ Words\end{tabular}} &
  \textbf{\begin{tabular}[c]{@{}c@{}}Semantic\\ Similarity\end{tabular}} &
  \textbf{\begin{tabular}[c]{@{}c@{}}Overlap\\ Similarity\end{tabular}} &
  \textbf{\begin{tabular}[c]{@{}c@{}}BERTScore\\ based\\ Similarity\end{tabular}} &
  \textbf{\begin{tabular}[c]{@{}c@{}}Phonetic\\ Similarity\end{tabular}} &
  \textbf{\begin{tabular}[c]{@{}c@{}}Avg. No. of\\ Candidates\\ per word\end{tabular}} &
  \textbf{\begin{tabular}[c]{@{}c@{}}Avg.\\ Word\\ Length\end{tabular}} &
  \textbf{\begin{tabular}[c]{@{}c@{}}Query\\ Number\end{tabular}} &
  \textbf{\begin{tabular}[c]{@{}c@{}}Avg.\\ Sentence\\ Length\end{tabular}} \\ \midrule
\multirow{3}{*}{\textbf{as}} &
  \textbf{Rand} &
  \multirow{3}{*}{0.52} &
  0.083/0.089 &
  8.234/7.707 &
  0.97/0.97 &
  0.941/0.946 &
  0.979/0.98 &
  - &
  4.779/4.805 &
  6.016/6.009 &
  15.342/14.638 &
  \multirow{3}{*}{17.814/17.628} \\ \cmidrule(lr){2-2} \cmidrule(lr){4-12}
 &
  \textbf{Phono} &
   &
  0.179/0.163 &
  8.27/8.332 &
  0.972/0.971 &
  0.943/0.943 &
  0.982/0.981 &
  0.896/0.897 &
  3.674/3.687 &
  5.912/5.874 &
  13.481/13.565 &
   \\ \cmidrule(lr){2-2} \cmidrule(lr){4-12}
 &
  \textbf{Ortho} &
   &
  0.137/0.128 &
  8.129/8.256 &
  0.971/0.97 &
  0.94/0.94 &
  0.98/0.98 &
  - &
  2.539/2.508 &
  5.873/5.841 &
  12.328/12.266 &
   \\ \midrule
\multirow{3}{*}{\textbf{bn}} &
  \textbf{Rand} &
  \multirow{3}{*}{0.498} &
  0.069/0.052 &
  9.375/8.572 &
  0.961/0.962 &
  0.935/0.941 &
  0.974/0.975 &
  - &
  5.483/5.43 &
  6.22/6.153 &
  13.928/13.238 &
  \multirow{3}{*}{15.031/15.299} \\ \cmidrule(lr){2-2} \cmidrule(lr){4-12}
 &
  \textbf{Phono} &
   &
  0.156/0.135 &
  9.192/8.755 &
  0.966/0.963 &
  0.937/0.94 &
  0.977/0.976 &
  0.978/0.978 &
  3.86/3.846 &
  6.182/6.122 &
  11.515/11.454 &
   \\ \cmidrule(lr){2-2} \cmidrule(lr){4-12}
 &
  \textbf{Ortho} &
   &
  0.116/0.1 &
  9.176/8.584 &
  0.966/0.964 &
  0.933/0.938 &
  0.975/0.975 &
  - &
  2.471/2.463 &
  6.126/6.053 &
  10.266/10.134 &
   \\ \midrule
\multirow{3}{*}{\textbf{gu}} &
  \textbf{Rand} &
  \multirow{3}{*}{0.77/0.77} &
  0.222/0.228 &
  13.265/12.169 &
  0.943/0.946 &
  0.901/0.912 &
  0.965/0.967 &
  - &
  4.95/4.931 &
  5.542/5.54 &
  22.734/21.576 &
  \multirow{3}{*}{16.53/16.594} \\ \cmidrule(lr){2-2} \cmidrule(lr){4-12}
 &
  \textbf{Phono} &
   &
  0.36/0.357 &
  13.591/12.845 &
  0.955/0.955 &
  0.901/0.909 &
  0.969/0.97 &
  0.968/0.969 &
  3.382/3.373 &
  5.506/5.487 &
  19.516/18.959 &
   \\ \cmidrule(lr){2-2} \cmidrule(lr){4-12}
 &
  \textbf{Ortho} &
   &
  0.357/0.346 &
  12.562/12.06 &
  0.947/0.946 &
  0.903/0.908 &
  0.966/0.966 &
  - &
  1.869/1.872 &
  5.401/5.385 &
  16.426/16.049 &
   \\ \midrule
\multirow{3}{*}{\textbf{hi}} &
  \textbf{Rand} &
  \multirow{3}{*}{0.5} &
  0.031/0.021 &
  7.253/6.041 &
  0.966/0.968 &
  0.936/0.946 &
  0.975/0.977 &
  - &
  4.639/4.577 &
  5.238/5.186 &
  15.99/14.521 &
  \multirow{3}{*}{19.749/20.201} \\ \cmidrule(lr){2-2} \cmidrule(lr){4-12}
 &
  \textbf{Phono} &
   &
  0.073/0.061 &
  8.495/7.076 &
  0.964/0.966 &
  0.927/0.938 &
  0.975/0.977 &
  0.979/0.981 &
  3.133/3.09 &
  5.095/5.035 &
  14.956/13.771 &
   \\ \cmidrule(lr){2-2} \cmidrule(lr){4-12}
 &
  \textbf{Ortho} &
   &
  0.059/0.045 &
  7.458/6.556 &
  0.964/0.966 &
  0.931/0.94 &
  0.975/0.977 &
  - &
  2.346/2.303 &
  5.071/5.007 &
  13.252/12.717 &
   \\ \midrule
\multirow{3}{*}{\textbf{kn}} &
  \textbf{Rand} &
  \multirow{3}{*}{0.514} &
  0.126/0.117 &
  8.42/8.013 &
  0.974/0.974 &
  0.953/0.957 &
  0.979/0.98 &
  - &
  5.406/5.418 &
  6.8/6.794 &
  12.82/11.913 &
  \multirow{3}{*}{14.402/14.236} \\ \cmidrule(lr){2-2} \cmidrule(lr){4-12}
 &
  \textbf{Phono} &
   &
  0.204/0.2 &
  8.054/8.076 &
  0.98/0.979 &
  0.956/0.957 &
  0.982/0.983 &
  0.982/0.982 &
  3.701/3.667 &
  6.831/6.826 &
  10.287/9.888 &
   \\ \cmidrule(lr){2-2} \cmidrule(lr){4-12}
 &
  \textbf{Ortho} &
   &
  0.222/0.221 &
  8.019/7.512 &
  0.976/0.977 &
  0.952/0.956 &
  0.981/0.982 &
  - &
  2.095/2.056 &
  6.933/6.875 &
  8.598/8.046 &
   \\ \midrule
\multirow{3}{*}{\textbf{ml}} &
  \textbf{Rand} &
  \multirow{3}{*}{0.548} &
  0.159/0.147 &
  10.267/9.452 &
  0.969/0.969 &
  0.949/0.955 &
  0.976/0.977 &
  - &
  5.798/5.776 &
  7.999/7.971 &
  13.688/12.737 &
  \multirow{3}{*}{13.238/13.265} \\ \cmidrule(lr){2-2} \cmidrule(lr){4-12}
 &
  \textbf{Phono} &
   &
  0.276/0.243 &
  9.022/9.464 &
  0.98/0.977 &
  0.956/0.956 &
  0.981/0.98 &
  0.952/0.953 &
  3.96/3.965 &
  7.997/8.012 &
  10.12/10.582 &
   \\ \cmidrule(lr){2-2} \cmidrule(lr){4-12}
 &
  \textbf{Ortho} &
   &
  0.228/0.201 &
  8.719/8.906 &
  0.972/0.971 &
  0.954/0.955 &
  0.977/0.977 &
  - &
  2.981/3.009 &
  8.043/8.036 &
  9.225/9.335 &
   \\ \midrule
\multirow{3}{*}{\textbf{mr}} &
  \textbf{Rand} &
  \multirow{3}{*}{0.525} &
  0.093/0.071 &
  8.289/7.994 &
  0.97/0.97 &
  0.946/0.949 &
  0.977/0.977 &
  - &
  5.582/5.608 &
  6.33/6.367 &
  14.304/14.079 &
  \multirow{3}{*}{15.938/16.051} \\ \cmidrule(lr){2-2} \cmidrule(lr){4-12}
 &
  \textbf{Phono} &
   &
  0.153/0.127 &
  8.683/8.608 &
  0.974/0.973 &
  0.944/0.946 &
  0.979/0.979 &
  0.979/0.98 &
  3.861/3.864 &
  6.273/6.303 &
  12.483/12.705 &
   \\ \cmidrule(lr){2-2} \cmidrule(lr){4-12}
 &
  \textbf{Ortho} &
   &
  0.125/0.11 &
  8.603/8.287 &
  0.971/0.97 &
  0.942/0.945 &
  0.977/0.977 &
  - &
  2.894/2.911 &
  6.206/6.27 &
  11.737/11.613 &
   \\ \midrule
\multirow{3}{*}{\textbf{pa}} &
  \textbf{Rand} &
  \multirow{3}{*}{0.595} &
  0.108/0.072 &
  9.233/8.549 &
  0.948/0.951 &
  0.918/0.923 &
  0.971/0.972 &
  - &
  4.192/4.164 &
  4.67/4.651 &
  20.059/19.385 &
  \multirow{3}{*}{20.124/20.372} \\ \cmidrule(lr){2-2} \cmidrule(lr){4-12}
 &
  \textbf{Phono} &
   &
  0.177/0.117 &
  9.026/9.366 &
  0.955/0.952 &
  0.92/0.917 &
  0.974/0.972 &
  0.939/0.937 &
  2.977/2.981 &
  4.583/4.589 &
  17.358/18.156 &
   \\ \cmidrule(lr){2-2} \cmidrule(lr){4-12}
 &
  \textbf{Ortho} &
   &
  0.195/0.167 &
  9.536/9.31 &
  0.95/0.95 &
  0.917/0.918 &
  0.971/0.971 &
  - &
  2.019/1.991 &
  4.496/4.449 &
  16.075/16.097 &
   \\ \midrule
\multirow{3}{*}{\textbf{te}} &
  \textbf{Rand} &
  \multirow{3}{*}{0.539} &
  0.13/0.114 &
  9.138/8.739 &
  0.969/0.97 &
  0.947/0.95 &
  0.977/0.978 &
  - &
  5.608/5.553 &
  6.928/6.864 &
  12.865/12.436 &
  \multirow{3}{*}{13.726/13.731} \\ \cmidrule(lr){2-2} \cmidrule(lr){4-12}
 &
  \textbf{Phono} &
   &
  0.209/0.197 &
  8.915/8.924 &
  0.976/0.975 &
  0.949/0.95 &
  0.981/0.981 &
  0.983/0.982 &
  3.898/3.836 &
  6.931/6.855 &
  10.753/10.821 &
   \\ \cmidrule(lr){2-2} \cmidrule(lr){4-12}
 &
  \textbf{Ortho} &
   &
  0.223/0.204 &
  8.284/8.501 &
  0.974/0.973 &
  0.948/0.948 &
  0.979/0.979 &
  - &
  2.251/2.23 &
  6.943/6.871 &
  8.827/9.063 &
   \\ \bottomrule
\end{tabular}%
}
\caption{The table presents the impact of different character-substitution attack strategies: random (\textbf{Rand}), phonetic (\textbf{Phono}), and orthographic (\textbf{Ortho}) on the \textit{mBERT} language model for the \textit{IndicParaphrase} dataset. Results are shown for perturbing both \textit{sentence1} and \textit{sentence2} of the IndicParaphrase dataset, separated by a forward slash (/).}
\label{tab:xnli_para}
\end{table*}

\begin{table*}[]
\centering
\resizebox{\textwidth}{!}{%
\begin{tabular}{@{}ccccccccccccc@{}}
\toprule
\textbf{Lang} &
  \textbf{\begin{tabular}[c]{@{}c@{}}Type\\ of\\ Perturbation\end{tabular}} &
  \textbf{\begin{tabular}[c]{@{}c@{}}Original\\ Accuracy\end{tabular}} &
  \textbf{\begin{tabular}[c]{@{}c@{}}After\\ Attack\\ Accuracy\end{tabular}} &
  \textbf{\begin{tabular}[c]{@{}c@{}}\%\\ Perturbed\\ Words\end{tabular}} &
  \textbf{\begin{tabular}[c]{@{}c@{}}Semantic\\ Similarity\end{tabular}} &
  \textbf{\begin{tabular}[c]{@{}c@{}}Overlap\\ Similarity\end{tabular}} &
  \textbf{\begin{tabular}[c]{@{}c@{}}BERTScore\\ based\\ Similarity\end{tabular}} &
  \textbf{\begin{tabular}[c]{@{}c@{}}Phonetic\\ Similarity\end{tabular}} &
  \textbf{\begin{tabular}[c]{@{}c@{}}Avg. No. of\\ Candidates\\ per word\end{tabular}} &
  \textbf{\begin{tabular}[c]{@{}c@{}}Avg.\\ Word\\ Length\end{tabular}} &
  \textbf{\begin{tabular}[c]{@{}c@{}}Query\\ Number\end{tabular}} &
  \textbf{\begin{tabular}[c]{@{}c@{}}Avg.\\ Sentence\\ Length\end{tabular}} \\ \midrule
\multirow{3}{*}{\textbf{as}} &
  \textbf{Rand} &
  \multirow{3}{*}{0.463} &
  0.092/0.055 &
  5.732/7.584 &
  0.971/0.96 &
  0.951/0.937 &
  0.983/0.98 &
  - &
  4.152/4.188 &
  5.153/5.112 &
  11.427/6.345 &
  \multirow{3}{*}{17.091/9.459} \\ \cmidrule(lr){2-2} \cmidrule(lr){4-12}
 &
  \textbf{Phono} &
   &
  0.131/0.079 &
  6.326/8.284 &
  0.973/0.964 &
  0.948/0.933 &
  0.984/0.981 &
  0.891/0.894 &
  3.265/3.249 &
  5.092/5.098 &
  11.012/6.38 &
   \\ \cmidrule(lr){2-2} \cmidrule(lr){4-12}
 &
  \textbf{Ortho} &
   &
  0.135/0.105 &
  5.832/7.721 &
  0.973/0.962 &
  0.951/0.935 &
  0.983/0.98 &
  - &
  2.144/2.065 &
  5.063/5.078 &
  9.801/5.478 &
   \\ \midrule
\multirow{3}{*}{\textbf{bn}} &
  \textbf{Rand} &
  \multirow{3}{*}{0.597} &
  0.121/0.063 &
  7.911/10.349 &
  0.952/0.934 &
  0.93/0.913 &
  0.973/0.969 &
  - &
  4.558/4.508 &
  5.027/4.958 &
  15.788/8.421 &
  \multirow{3}{*}{16.971/9.221} \\ \cmidrule(lr){2-2} \cmidrule(lr){4-12}
 &
  \textbf{Phono} &
   &
  0.168/0.104 &
  8.311/10.786 &
  0.956/0.941 &
  0.93/0.912 &
  0.974/0.97 &
  0.976/0.98 &
  3.308/3.23 &
  4.998/4.935 &
  14.68/7.866 &
   \\ \cmidrule(lr){2-2} \cmidrule(lr){4-12}
 &
  \textbf{Ortho} &
   &
  0.204/0.134 &
  7.376/10.694 &
  0.958/0.941 &
  0.934/0.911 &
  0.975/0.97 &
  - &
  2.061/1.969 &
  4.943/4.871 &
  12.481/7.155 &
   \\ \midrule
\multirow{3}{*}{\textbf{gu}} &
  \textbf{Rand} &
  \multirow{3}{*}{0.556} &
  0.115/0.06 &
  6.924/9.301 &
  0.955/0.942 &
  0.932/0.917 &
  0.975/0.973 &
  - &
  4.066/4.032 &
  4.548/4.528 &
  14.963/8.305 &
  \multirow{3}{*}{18.389/9.956} \\ \cmidrule(lr){2-2} \cmidrule(lr){4-12}
 &
  \textbf{Phono} &
   &
  0.171/0.096 &
  7.376/9.89 &
  0.962/0.95 &
  0.931/0.916 &
  0.978/0.975 &
  0.977/0.983 &
  2.836/2.795 &
  4.482/4.466 &
  13.843/7.765 &
   \\ \cmidrule(lr){2-2} \cmidrule(lr){4-12}
 &
  \textbf{Ortho} &
   &
  0.235/0.197 &
  6.348/8.263 &
  0.962/0.952 &
  0.939/0.924 &
  0.978/0.975 &
  - &
  1.415/1.376 &
  4.403/4.387 &
  11.631/6.31 &
   \\ \midrule
\multirow{3}{*}{\textbf{hi}} &
  \textbf{Rand} &
  \multirow{3}{*}{0.629} &
  0.113/0.049 &
  7.21/9.06 &
  0.948/0.934 &
  0.919/0.903 &
  0.971/0.968 &
  - &
  3.522/3.561 &
  3.928/4.037 &
  19.763/9.955 &
  \multirow{3}{*}{21.825/11.492} \\ \cmidrule(lr){2-2} \cmidrule(lr){4-12}
 &
  \textbf{Phono} &
   &
  0.182/0.085 &
  7.441/9.725 &
  0.957/0.942 &
  0.919/0.898 &
  0.974/0.97 &
  0.976/0.983 &
  2.446/2.482 &
  3.829/3.926 &
  18.193/9.56 &
   \\ \cmidrule(lr){2-2} \cmidrule(lr){4-12}
 &
  \textbf{Ortho} &
   &
  0.187/0.088 &
  6.632/9.363 &
  0.954/0.933 &
  0.924/0.9 &
  0.974/0.968 &
  - &
  1.683/1.696 &
  3.826/3.906 &
  16.209/9.025 &
   \\ \midrule
\multirow{3}{*}{\textbf{kn}} &
  \textbf{Rand} &
  \multirow{3}{*}{0.585} &
  0.165/0.088 &
  8.626/11.991 &
  0.962/0.947 &
  0.945/0.93 &
  0.976/0.973 &
  - &
  5.844/6.274 &
  7.378/7.897 &
  14.338/8.194 &
  \multirow{3}{*}{14.072/7.533} \\ \cmidrule(lr){2-2} \cmidrule(lr){4-12}
 &
  \textbf{Phono} &
   &
  0.238/0.177 &
  8.535/11.907 &
  0.969/0.958 &
  0.946/0.93 &
  0.979/0.977 &
  0.985/0.987 &
  3.953/4.129 &
  7.306/7.731 &
  12.128/6.951 &
   \\ \cmidrule(lr){2-2} \cmidrule(lr){4-12}
 &
  \textbf{Ortho} &
   &
  0.336/0.272 &
  6.567/10.238 &
  0.977/0.963 &
  0.958/0.939 &
  0.982/0.978 &
  - &
  2.024/2.114 &
  7.235/7.557 &
  8.139/5.053 &
   \\ \midrule
\multirow{3}{*}{\textbf{ml}} &
  \textbf{Rand} &
  \multirow{3}{*}{0.562} &
  0.172/0.093 &
  8.052/12.045 &
  0.962/0.95 &
  0.95/0.938 &
  0.976/0.972 &
  - &
  5.992/6.633 &
  8.428/9.213 &
  12.08/7.532 &
  \multirow{3}{*}{12.98/6.837} \\ \cmidrule(lr){2-2} \cmidrule(lr){4-12}
 &
  \textbf{Phono} &
   &
  0.262/0.213 &
  7.92/11.812 &
  0.974/0.964 &
  0.954/0.941 &
  0.98/0.977 &
  0.954/0.956 &
  3.945/4.136 &
  8.361/8.924 &
  10.142/6.223 &
   \\ \cmidrule(lr){2-2} \cmidrule(lr){4-12}
 &
  \textbf{Ortho} &
   &
  0.251/0.212 &
  6.676/10.309 &
  0.968/0.955 &
  0.958/0.946 &
  0.978/0.974 &
  - &
  3.032/3.152 &
  8.371/8.905 &
  8.215/4.872 &
   \\ \midrule
\multirow{3}{*}{\textbf{mr}} &
  \textbf{Rand} &
  \multirow{3}{*}{0.544} &
  0.115/0.058 &
  7.399/9.787 &
  0.96/0.944 &
  0.936/0.922 &
  0.976/0.973 &
  - &
  4.734/4.831 &
  5.335/5.464 &
  13.446/7.468 &
  \multirow{3}{*}{16.154/8.773} \\ \cmidrule(lr){2-2} \cmidrule(lr){4-12}
 &
  \textbf{Phono} &
   &
  0.171/0.103 &
  7.364/10.338 &
  0.967/0.955 &
  0.938/0.922 &
  0.979/0.976 &
  0.979/0.983 &
  3.191/3.271 &
  5.229/5.356 &
  11.592/6.87 &
   \\ \cmidrule(lr){2-2} \cmidrule(lr){4-12}
 &
  \textbf{Ortho} &
   &
  0.166/0.105 &
  7.13/9.866 &
  0.961/0.945 &
  0.938/0.921 &
  0.977/0.974 &
  - &
  2.405/2.494 &
  5.204/5.369 &
  10.979/6.279 &
   \\ \midrule
\multirow{3}{*}{\textbf{pa}} &
  \textbf{Rand} &
  \multirow{3}{*}{0.594} &
  0.102/0.049 &
  6.378/8.381 &
  0.957/0.94 &
  0.93/0.91 &
  0.974/0.972 &
  - &
  3.526/3.478 &
  3.972/3.972 &
  18.09/9.443 &
  \multirow{3}{*}{21.909/11.725} \\ \cmidrule(lr){2-2} \cmidrule(lr){4-12}
 &
  \textbf{Phono} &
   &
  0.153/0.064 &
  6.677/8.916 &
  0.963/0.948 &
  0.929/0.905 &
  0.976/0.973 &
  0.943/0.944 &
  2.568/2.469 &
  3.918/3.923 &
  17.149/9.106 &
   \\ \cmidrule(lr){2-2} \cmidrule(lr){4-12}
 &
  \textbf{Ortho} &
   &
  0.187/0.105 &
  6.571/8.679 &
  0.96/0.943 &
  0.93/0.908 &
  0.975/0.97 &
  - &
  1.645/1.606 &
  3.871/3.856 &
  15.551/8.402 &
   \\ \midrule
\multirow{3}{*}{\textbf{ta}} &
  \textbf{Rand} &
  \multirow{3}{*}{0.557} &
  0.148/0.077 &
  7.58/11.328 &
  0.96/0.946 &
  0.953/0.936 &
  0.978/0.975 &
  - &
  5.337/5.708 &
  7.757/8.1 &
  12.845/7.779 &
  \multirow{3}{*}{14.721/7.826} \\ \cmidrule(lr){2-2} \cmidrule(lr){4-12}
 &
  \textbf{Phono} &
   &
  0.253/0.176 &
  6.979/10.412 &
  0.972/0.962 &
  0.958/0.94 &
  0.982/0.978 &
  0.978/0.984 &
  3.765/3.931 &
  7.695/8.014 &
  9.161/5.487 &
   \\ \cmidrule(lr){2-2} \cmidrule(lr){4-12}
 &
  \textbf{Ortho} &
   &
  0.245/0.206 &
  6.214/9.933 &
  0.976/0.964 &
  0.96/0.944 &
  0.98/0.976 &
  - &
  2.104/2.189 &
  7.65/7.953 &
  8.676/5.175 &
   \\ \midrule
\multirow{3}{*}{\textbf{te}} &
  \textbf{Rand} &
  \multirow{3}{*}{0.545} &
  0.132/0.085 &
  7.612/10.955 &
  0.962/0.949 &
  0.944/0.929 &
  0.978/0.975 &
  - &
  5.253/5.511 &
  6.368/6.678 &
  13.042/7.327 &
  \multirow{3}{*}{14.655/7.706} \\ \cmidrule(lr){2-2} \cmidrule(lr){4-12}
 &
  \textbf{Phono} &
   &
  0.188/0.121 &
  8.012/11.275 &
  0.967/0.955 &
  0.944/0.928 &
  0.98/0.977 &
  0.976/0.985 &
  3.618/3.913 &
  6.3/6.638 &
  11.709/6.663 &
   \\ \cmidrule(lr){2-2} \cmidrule(lr){4-12}
 &
  \textbf{Ortho} &
   &
  0.253/0.197 &
  6.455/10.233 &
  0.969/0.954 &
  0.952/0.933 &
  0.981/0.977 &
  - &
  1.902/2.025 &
  6.281/6.592 &
  8.56/5.082 &
   \\ \bottomrule
\end{tabular}%
}
\caption{The table presents the impact of different character-substitution attack strategies: random (\textbf{Rand}), phonetic (\textbf{Phono}), and orthographic (\textbf{Ortho}) on the \textit{mBERT} language model for the \textit{IndicXNLI} dataset. Results are shown for perturbing both \textit{premise} and \textit{hypothesis} of the IndicXNLI dataset, separated by a forward slash (/).}
\label{tab:xnli_mbert}
\end{table*}

\section{Synonym-based Word Substitution}\label{sec:synonym}
We have performed word-level substitutions by replacing a word with its synonyms from IndoWordNet \cite{bhattacharyya-2010-indowordnet} for IndicBERTv2 language model. The results on IndicSentiment, IndicParaphrase and IndicXNLI datasets are presented in Tables \ref{tab:semantics_senti_indicbert} - \ref{tab:semantics_xnli_indicbert}.

The results reveal that word-level synonym substitution is quite effective in hampering the performance of IndicBERTv2. We observe that Hindi(hi) seems to be the most vulnerable among different languages. Similar to the observations in character-level attack, Dravidian languages are most robust to even synonym-baed substitution at the word level. Some languages such as Bengali, Marathi languages like Bengali and Marathi show greater robustness to word-level perturbations compared to character-level ones.

\begin{table*}[]
\centering
\resizebox{\textwidth}{!}{%
\begin{tabular}{@{}cccccccccc@{}}
\toprule
\textbf{Language} &
  \textbf{Original Accuracy} &
  \textbf{After-Attack Accuracy} &
  \textbf{\% Perturbed Words} &
  \textbf{Semantic Similarity} &
  \textbf{Overlap Similarity} &
  \textbf{BERTScore based Similarity} &
  \textbf{Avg. No. of Candidates per word} &
  \textbf{Query Number} &
  \textbf{Avg. Sentence Length} \\ \midrule
\textbf{as} & 0.929 & 0.415 & 12.461 & 0.923 & 0.851 & 0.962 & 3.376  & 50.776 & 24.003 \\ \midrule
\textbf{bd} & 0.856 & 0.262 & 12     & 0.947 & 0.874 & 0.968 & 3.039  & 38.801 & 22.205 \\ \midrule
\textbf{bn} & 0.954 & 0.513 & 12.718 & 0.942 & 0.842 & 0.96  & 2.108  & 40.79  & 23.721 \\ \midrule
\textbf{gu} & 0.941 & 0.269 & 12.251 & 0.938 & 0.827 & 0.962 & 6.038  & 70.848 & 26.129 \\ \midrule
\textbf{hi} & 0.955 & 0.16  & 11.167 & 0.931 & 0.833 & 0.961 & 10.806 & 98.525 & 30.107 \\ \midrule
\textbf{kn} & 0.937 & 0.624 & 12.555 & 0.942 & 0.871 & 0.968 & 3.184  & 44.858 & 20.034 \\ \midrule
\textbf{ml} & 0.939 & 0.771 & 7.917  & 0.961 & 0.916 & 0.977 & 1.868  & 29.659 & 19.254 \\ \midrule
\textbf{mr} & 0.946 & 0.569 & 12.436 & 0.956 & 0.854 & 0.969 & 2.46   & 40.175 & 23.172 \\ \midrule
\textbf{or} & 0.93  & 0.265 & 13.357 & 0.943 & 0.84  & 0.968 & 3.933  & 49.529 & 23.491 \\ \midrule
\textbf{pa} & 0.949 & 0.318 & 9.767  & 0.938 & 0.847 & 0.959 & 3.864  & 61.163 & 30.062 \\ \midrule
\textbf{ta} & 0.948 & 0.791 & 9.174  & 0.959 & 0.901 & 0.978 & 1.12   & 27.507 & 20.874 \\ \midrule
\textbf{te} & 0.947 & 0.673 & 9.372  & 0.95  & 0.889 & 0.969 & 1.626  & 32.92  & 21.76  \\ \bottomrule
\end{tabular}%
}
\caption{Impact of synonym-based word substitution on IndicSentiment dataset for IndicBERTv2 model}
\label{tab:semantics_senti_indicbert}
\end{table*}

\begin{table*}[]
\centering
\resizebox{\textwidth}{!}{%
\begin{tabular}{@{}cccccccccc@{}}
\toprule
\textbf{Language} &
  \textbf{Original Accuracy} &
  \textbf{After-Attack Accuracy} &
  \textbf{\% Perturbed Words} &
  \textbf{Semantic Similarity} &
  \textbf{Overlap Similarity} &
  \textbf{BERTScore based Similarity} &
  \textbf{Avg. No. of Candidates per word} &
  \textbf{Query Number} &
  \textbf{Avg. Sentence Length} \\ \midrule
\textbf{as} & 0.572 & 0.35/0.34   & 5.41/5.555  & 0.967/0.966 & 0.939/0.936 & 0.983/0.983 & 2.085/2.166  & 12.549/13.215 & 17.814/17.628 \\ \midrule
\textbf{bn} & 0.498 & 0.226/0.229 & 5.492/5.546 & 0.972/0.97  & 0.935/0.934 & 0.982/0.982 & 1.68/1.708   & 8.528/8.796   & 15.031/15.299 \\ \midrule
\textbf{gu} & 0.723 & 0.287/0.233 & 8.802/9.164 & 0.957/0.952 & 0.891/0.885 & 0.975/0.973 & 5.288/5.306  & 25.107/25.897 & 16.53/16.594  \\ \midrule
\textbf{hi} & 0.498 & 0.027/0.025 & 5.97/6.086  & 0.962/0.961 & 0.913/0.91  & 0.979/0.979 & 12.731/13.38 & 26.471/28.227 & 19.749/20.201 \\ \midrule
\textbf{kn} & 0.583 & 0.34/0.351  & 5.382/5.845 & 0.977/0.975 & 0.943/0.941 & 0.984/0.984 & 2.644/2.529  & 11.631/11.818 & 14.402/14.236 \\ \midrule
\textbf{ml} & 0.566 & 0.485/0.496 & 2.042/1.768 & 0.988/0.99  & 0.979/0.982 & 0.994/0.995 & 1.134/1.173  & 4.069/3.928   & 13.238/13.265 \\ \midrule
\textbf{mr} & 0.544 & 0.375/0.362 & 4.33/4.823  & 0.984/0.981 & 0.952/0.949 & 0.989/0.988 & 1.524/1.552  & 7.626/8.484   & 15.938/16.051 \\ \midrule
\textbf{or} & 0.576 & 0.254/0.231 & 6.724/7.481 & 0.97/0.965  & 0.922/0.916 & 0.981/0.979 & 2.355/2.43   & 12.555/13.692 & 15.884/16.035 \\ \midrule
\textbf{pa} & 0.543 & 0.198/0.17  & 6.11/6.795  & 0.96/0.957  & 0.915/0.906 & 0.978/0.976 & 2.644/2.683  & 16.466/17.944 & 20.124/20.372 \\ \midrule
\textbf{te} & 0.545 & 0.397/0.412 & 3.284/3.052 & 0.978/0.98  & 0.961/0.963 & 0.989/0.991 & 1.139/1.092  & 5.527/5.022   & 13.726/13.731 \\ \bottomrule
\end{tabular}%
}
\caption{Impact of synonym-based word substitution on IndicParaphrase dataset for IndicBERTv2 model}
\label{tab:semantics_para_indicbert}
\end{table*}

\begin{table*}[]
\centering
\resizebox{\textwidth}{!}{%
\begin{tabular}{cccccccccc}
\hline
\textbf{Language} &
  \textbf{Original Accuracy} &
  \textbf{After-Attack Accuracy} &
  \textbf{\% Perturbed Words} &
  \textbf{LaBSE Similarity} &
  \textbf{chrF Similarity} &
  \textbf{BERTScore} &
  \textbf{Avg. No. of Candidates (avg)} &
  \textbf{Query Number} &
  \textbf{Sentence Length (avg)} \\ \hline
\textbf{as} & 0.676 & 0.431/0.322 & 5.026/8.953 & 0.967/0.971 & 0.932/0.939 & 0.982/0.98  & 2.273/3.606   & 14.313/14.663 & 17.091/17.814 \\ \hline
\textbf{bn} & 0.711 & 0.437/0.351 & 5.303/8.083 & 0.964/0.949 & 0.922/0.894 & 0.977/0.972 & 1.781/2.058   & 12.739/8.452  & 16.971/9.221  \\ \hline
\textbf{gu} & 0.735 & 0.334/0.251 & 6.54/8.9    & 0.959/0.947 & 0.898/0.866 & 0.976/0.974 & 4.532/4.873   & 23.901/12.978 & 18.389/9.956  \\ \hline
\textbf{hi} & 0.718 & 0.2/0.075   & 6.06/8.016  & 0.953/0.937 & 0.893/0.863 & 0.974/0.971 & 10.641/12.697 & 35.366/18.206 & 21.825/11.492 \\ \hline
\textbf{kn} & 0.717 & 0.534/0.494 & 4.567/7.249 & 0.975/0.962 & 0.948/0.92  & 0.986/0.983 & 2.418/2.313   & 11.165/6.959  & 14.072/7.533  \\ \hline
\textbf{ml} & 0.727 & 0.601/0.584 & 3.351/5.934 & 0.978/0.967 & 0.957/0.936 & 0.987/0.987 & 1.348/1.376   & 7.984/4.667   & 12.98/6.837   \\ \hline
\textbf{mr} & 0.696 & 0.476/0.39  & 4.813/8.185 & 0.976/0.962 & 0.935/0.896 & 0.984/0.98  & 1.675/2.106   & 11.612/8.145  & 16.154/8.773  \\ \hline
\textbf{or} & 0.703 & 0.401/0.307 & 5.798/8.783 & 0.968/0.952 & 0.918/0.887 & 0.982/0.977 & 2.443/2.864   & 14.659/9.183  & 16.173/8.847  \\ \hline
\textbf{pa} & 0.738 & 0.353/0.258 & 5.848/8.014 & 0.955/0.941 & 0.9/0.871   & 0.974/0.971 & 3.042/3.52    & 23.775/13.031 & 21.909/11.725 \\ \hline
\textbf{ta} & 0.717 & 0.592/0.575 & 3.271/5.273 & 0.984/0.978 & 0.961/0.942 & 0.99/0.988  & 0.853/0.761   & 7.433/4.078   & 14.721/7.826  \\ \hline
\textbf{te} & 0.713 & 0.468/0.432 & 4.925/8.208 & 0.967/0.95  & 0.929/0.901 & 0.982/0.979 & 1.916/2.028   & 11.461/7.212  & 14.655/7.706  \\ \hline
\end{tabular}%
}
\caption{Impact of synonym-based word substitution on IndicXNLI dataset for IndicBERTv2 model}
\label{tab:semantics_xnli_indicbert}
\end{table*}


\begin{thebibliography}{44}
\expandafter\ifx\csname natexlab\endcsname\relax\def\natexlab#1{#1}\fi

\bibitem[{Alzantot et~al.(2018)Alzantot, Sharma, Elgohary, Ho, Srivastava, and Chang}]{alzantot2018generating}
Moustafa Alzantot, Yash Sharma, Ahmed Elgohary, Bo-Jhang Ho, Mani Srivastava, and Kai-Wei Chang. 2018.
\newblock Generating natural language adversarial examples.
\newblock \emph{arXiv preprint arXiv:1804.07998}.

\bibitem[{Bhattacharyya(2010)}]{bhattacharyya-2010-indowordnet}
Pushpak Bhattacharyya. 2010.
\newblock \href {http://www.lrec-conf.org/proceedings/lrec2010/pdf/939_Paper.pdf} {{I}ndo{W}ord{N}et}.
\newblock In \emph{Proceedings of the Seventh International Conference on Language Resources and Evaluation ({LREC}'10)}, Valletta, Malta. European Language Resources Association (ELRA).

\bibitem[{Bommasani et~al.(2021)Bommasani, Hudson, Adeli, Altman, Arora, von Arx, Bernstein, Bohg, Bosselut, Brunskill et~al.}]{bommasani2021opportunities}
Rishi Bommasani, Drew~A Hudson, Ehsan Adeli, Russ Altman, Simran Arora, Sydney von Arx, Michael~S Bernstein, Jeannette Bohg, Antoine Bosselut, Emma Brunskill, et~al. 2021.
\newblock On the opportunities and risks of foundation models.
\newblock \emph{arXiv preprint arXiv:2108.07258}.

\bibitem[{Brown et~al.(2020)Brown, Mann, Ryder, Subbiah, Kaplan, Dhariwal, Neelakantan, Shyam, Sastry, Askell et~al.}]{brown2020language}
Tom Brown, Benjamin Mann, Nick Ryder, Melanie Subbiah, Jared~D Kaplan, Prafulla Dhariwal, Arvind Neelakantan, Pranav Shyam, Girish Sastry, Amanda Askell, et~al. 2020.
\newblock Language models are few-shot learners.
\newblock \emph{Advances in neural information processing systems}, 33:1877--1901.

\bibitem[{Choksi and Thakkar(2012)}]{choksi2012recognition}
Amit~H Choksi and Shital~P Thakkar. 2012.
\newblock Recognition of similar appearing gujarati characters using fuzzy-knn algorithm.
\newblock \emph{International Journal of Computer Applications}, 55(6).

\bibitem[{Conneau et~al.(2020)Conneau, Khandelwal, Goyal, Chaudhary, Wenzek, Guzm{\'a}n, Grave, Ott, Zettlemoyer, and Stoyanov}]{conneau-etal-2020-unsupervised}
Alexis Conneau, Kartikay Khandelwal, Naman Goyal, Vishrav Chaudhary, Guillaume Wenzek, Francisco Guzm{\'a}n, Edouard Grave, Myle Ott, Luke Zettlemoyer, and Veselin Stoyanov. 2020.
\newblock \href {https://doi.org/10.18653/v1/2020.acl-main.747} {Unsupervised cross-lingual representation learning at scale}.
\newblock In \emph{Proceedings of the 58th Annual Meeting of the Association for Computational Linguistics}, pages 8440--8451, Online. Association for Computational Linguistics.

\bibitem[{Devlin et~al.(2019)Devlin, Chang, Lee, and Toutanova}]{devlin-etal-2019-bert}
Jacob Devlin, Ming-Wei Chang, Kenton Lee, and Kristina Toutanova. 2019.
\newblock \href {https://doi.org/10.18653/v1/N19-1423} {{BERT}: Pre-training of deep bidirectional transformers for language understanding}.
\newblock In \emph{Proceedings of the 2019 Conference of the North {A}merican Chapter of the Association for Computational Linguistics: Human Language Technologies, Volume 1 (Long and Short Papers)}, pages 4171--4186, Minneapolis, Minnesota. Association for Computational Linguistics.

\bibitem[{Dholakia et~al.(2007)Dholakia, Yajnik, and Negi}]{dholakia2007wavelet}
Jignesh Dholakia, Archit Yajnik, and Atul Negi. 2007.
\newblock Wavelet feature based confusion character sets for gujarati script.
\newblock In \emph{International Conference on Computational Intelligence and Multimedia Applications (ICCIMA 2007)}, volume~2, pages 366--370. IEEE.

\bibitem[{Doddapaneni et~al.(2023{\natexlab{a}})Doddapaneni, Aralikatte, Ramesh, Goyal, Khapra, Kunchukuttan, and Kumar}]{doddapaneni2023towards}
Sumanth Doddapaneni, Rahul Aralikatte, Gowtham Ramesh, Shreya Goyal, Mitesh~M Khapra, Anoop Kunchukuttan, and Pratyush Kumar. 2023{\natexlab{a}}.
\newblock Towards leaving no indic language behind: Building monolingual corpora, benchmark and models for indic languages.
\newblock In \emph{Proceedings of the 61st Annual Meeting of the Association for Computational Linguistics (Volume 1: Long Papers)}, pages 12402--12426.

\bibitem[{Doddapaneni et~al.(2023{\natexlab{b}})Doddapaneni, Aralikatte, Ramesh, Goyal, Khapra, Kunchukuttan, and Kumar}]{doddapaneni-etal-2023-towards}
Sumanth Doddapaneni, Rahul Aralikatte, Gowtham Ramesh, Shreya Goyal, Mitesh~M. Khapra, Anoop Kunchukuttan, and Pratyush Kumar. 2023{\natexlab{b}}.
\newblock \href {https://doi.org/10.18653/v1/2023.acl-long.693} {Towards leaving no {I}ndic language behind: Building monolingual corpora, benchmark and models for {I}ndic languages}.
\newblock In \emph{Proceedings of the 61st Annual Meeting of the Association for Computational Linguistics (Volume 1: Long Papers)}, pages 12402--12426, Toronto, Canada. Association for Computational Linguistics.

\bibitem[{Ebrahimi et~al.(2017)Ebrahimi, Rao, Lowd, and Dou}]{ebrahimi2017hotflip}
Javid Ebrahimi, Anyi Rao, Daniel Lowd, and Dejing Dou. 2017.
\newblock Hotflip: White-box adversarial examples for text classification.
\newblock \emph{arXiv preprint arXiv:1712.06751}.

\bibitem[{Feng et~al.(2020)Feng, Yang, Cer, Arivazhagan, and Wang}]{feng2020language}
Fangxiaoyu Feng, Yinfei Yang, Daniel Cer, Naveen Arivazhagan, and Wei Wang. 2020.
\newblock Language-agnostic bert sentence embedding.
\newblock \emph{arXiv preprint arXiv:2007.01852}.

\bibitem[{Gao et~al.(2018)Gao, Lanchantin, Soffa, and Qi}]{gao2018black}
Ji~Gao, Jack Lanchantin, Mary~Lou Soffa, and Yanjun Qi. 2018.
\newblock Black-box generation of adversarial text sequences to evade deep learning classifiers.
\newblock In \emph{2018 IEEE Security and Privacy Workshops (SPW)}, pages 50--56. IEEE.

\bibitem[{Garg and Ramakrishnan(2020)}]{garg2020bae}
Siddhant Garg and Goutham Ramakrishnan. 2020.
\newblock Bae: Bert-based adversarial examples for text classification.
\newblock \emph{arXiv preprint arXiv:2004.01970}.

\bibitem[{Guo et~al.(2021)Guo, Sablayrolles, J{\'e}gou, and Kiela}]{guo2021gradient}
Chuan Guo, Alexandre Sablayrolles, Herv{\'e} J{\'e}gou, and Douwe Kiela. 2021.
\newblock Gradient-based adversarial attacks against text transformers.
\newblock \emph{arXiv preprint arXiv:2104.13733}.

\bibitem[{Jangid and Srivastava(2016)}]{jangid2016similar}
Mahesh Jangid and Sumit Srivastava. 2016.
\newblock Similar handwritten devanagari character recognition by critical region estimation.
\newblock In \emph{2016 international conference on advances in computing, communications and informatics (ICACCI)}, pages 1936--1939. IEEE.

\bibitem[{Jin et~al.(2020{\natexlab{a}})Jin, Jin, Zhou, and Szolovits}]{jin2020bert}
Di~Jin, Zhijing Jin, Joey~Tianyi Zhou, and Peter Szolovits. 2020{\natexlab{a}}.
\newblock Is bert really robust? a strong baseline for natural language attack on text classification and entailment.
\newblock In \emph{Proceedings of the AAAI conference on artificial intelligence}, volume~34, pages 8018--8025.

\bibitem[{Jin et~al.(2020{\natexlab{b}})Jin, Jin, Zhou, and Szolovits}]{Jin_Jin_Zhou_Szolovits_2020}
Di~Jin, Zhijing Jin, Joey~Tianyi Zhou, and Peter Szolovits. 2020{\natexlab{b}}.
\newblock \href {https://doi.org/10.1609/aaai.v34i05.6311} {Is bert really robust? a strong baseline for natural language attack on text classification and entailment}.
\newblock \emph{Proceedings of the AAAI Conference on Artificial Intelligence}, 34(05):8018--8025.

\bibitem[{Khanuja et~al.(2021)Khanuja, Bansal, Mehtani, Khosla, Dey, Gopalan, Margam, Aggarwal, Nagipogu, Dave et~al.}]{khanuja2021muril}
Simran Khanuja, Diksha Bansal, Sarvesh Mehtani, Savya Khosla, Atreyee Dey, Balaji Gopalan, Dilip~Kumar Margam, Pooja Aggarwal, Rajiv~Teja Nagipogu, Shachi Dave, et~al. 2021.
\newblock Muril: Multilingual representations for indian languages.
\newblock \emph{arXiv preprint arXiv:2103.10730}.

\bibitem[{Kunchukuttan(2020)}]{kunchukuttan2020indicnlp}
Anoop Kunchukuttan. 2020.
\newblock {The IndicNLP Library}.
\newblock \url{https://github.com/anoopkunchukuttan/indic_nlp_library/blob/master/docs/indicnlp.pdf}.

\bibitem[{Kurakin et~al.(2016)Kurakin, Goodfellow, and Bengio}]{kurakin2016adversarial}
Alexey Kurakin, Ian Goodfellow, and Samy Bengio. 2016.
\newblock Adversarial machine learning at scale.
\newblock \emph{arXiv preprint arXiv:1611.01236}.

\bibitem[{Kurakin et~al.(2018)Kurakin, Goodfellow, and Bengio}]{kurakin2018adversarial}
Alexey Kurakin, Ian~J Goodfellow, and Samy Bengio. 2018.
\newblock Adversarial examples in the physical world.
\newblock In \emph{Artificial intelligence safety and security}, pages 99--112. Chapman and Hall/CRC.

\bibitem[{Li et~al.(2018)Li, Ji, Du, Li, and Wang}]{li2018textbugger}
Jinfeng Li, Shouling Ji, Tianyu Du, Bo~Li, and Ting Wang. 2018.
\newblock Textbugger: Generating adversarial text against real-world applications.
\newblock \emph{arXiv preprint arXiv:1812.05271}.

\bibitem[{Li et~al.(2020{\natexlab{a}})Li, Ma, Guo, Xue, and Qiu}]{li2020bert}
Linyang Li, Ruotian Ma, Qipeng Guo, Xiangyang Xue, and Xipeng Qiu. 2020{\natexlab{a}}.
\newblock Bert-attack: Adversarial attack against bert using bert.
\newblock \emph{arXiv preprint arXiv:2004.09984}.

\bibitem[{Li et~al.(2020{\natexlab{b}})Li, Ma, Guo, Xue, and Qiu}]{li-etal-2020-bert-attack}
Linyang Li, Ruotian Ma, Qipeng Guo, Xiangyang Xue, and Xipeng Qiu. 2020{\natexlab{b}}.
\newblock \href {https://doi.org/10.18653/v1/2020.emnlp-main.500} {{BERT}-{ATTACK}: Adversarial attack against {BERT} using {BERT}}.
\newblock In \emph{Proceedings of the 2020 Conference on Empirical Methods in Natural Language Processing (EMNLP)}, pages 6193--6202, Online. Association for Computational Linguistics.

\bibitem[{Liu et~al.(2019)Liu, Ott, Goyal, Du, Joshi, Chen, Levy, Lewis, Zettlemoyer, and Stoyanov}]{liu2019roberta}
Yinhan Liu, Myle Ott, Naman Goyal, Jingfei Du, Mandar Joshi, Danqi Chen, Omer Levy, Mike Lewis, Luke Zettlemoyer, and Veselin Stoyanov. 2019.
\newblock Roberta: A robustly optimized bert pretraining approach.
\newblock \emph{arXiv preprint arXiv:1907.11692}.

\bibitem[{Mehrabi et~al.(2022)Mehrabi, Beirami, Morstatter, and Galstyan}]{mehrabi2022robust}
Ninareh Mehrabi, Ahmad Beirami, Fred Morstatter, and Aram Galstyan. 2022.
\newblock Robust conversational agents against imperceptible toxicity triggers.
\newblock \emph{arXiv preprint arXiv:2205.02392}.

\bibitem[{Morris et~al.(2020)Morris, Lifland, Yoo, Grigsby, Jin, and Qi}]{morris2020textattack}
John~X Morris, Eli Lifland, Jin~Yong Yoo, Jake Grigsby, Di~Jin, and Yanjun Qi. 2020.
\newblock Textattack: A framework for adversarial attacks, data augmentation, and adversarial training in nlp.
\newblock \emph{arXiv preprint arXiv:2005.05909}.

\bibitem[{Naik and Desai(2017)}]{naik2017online}
Vishal~A Naik and Apurva~A Desai. 2017.
\newblock Online handwritten gujarati character recognition using svm, mlp, and k-nn.
\newblock In \emph{2017 8th international conference on computing, communication and networking technologies (icccnt)}, pages 1--6. IEEE.

\bibitem[{Popovi{\'c}(2015)}]{popovic-2015-chrf}
Maja Popovi{\'c}. 2015.
\newblock \href {https://doi.org/10.18653/v1/W15-3049} {chr{F}: character n-gram {F}-score for automatic {MT} evaluation}.
\newblock In \emph{Proceedings of the Tenth Workshop on Statistical Machine Translation}, pages 392--395, Lisbon, Portugal. Association for Computational Linguistics.

\bibitem[{Pruthi et~al.(2019{\natexlab{a}})Pruthi, Dhingra, and Lipton}]{pruthi2019combating}
Danish Pruthi, Bhuwan Dhingra, and Zachary~C Lipton. 2019{\natexlab{a}}.
\newblock Combating adversarial misspellings with robust word recognition.
\newblock \emph{arXiv preprint arXiv:1905.11268}.

\bibitem[{Pruthi et~al.(2019{\natexlab{b}})Pruthi, Dhingra, and Lipton}]{pruthi-etal-2019-combating}
Danish Pruthi, Bhuwan Dhingra, and Zachary~C. Lipton. 2019{\natexlab{b}}.
\newblock \href {https://doi.org/10.18653/v1/P19-1561} {Combating adversarial misspellings with robust word recognition}.
\newblock In \emph{Proceedings of the 57th Annual Meeting of the Association for Computational Linguistics}, pages 5582--5591, Florence, Italy. Association for Computational Linguistics.

\bibitem[{Purkaystha et~al.(2017)Purkaystha, Datta, and Islam}]{purkaystha2017bengali}
Bishwajit Purkaystha, Tapos Datta, and Md~Saiful Islam. 2017.
\newblock Bengali handwritten character recognition using deep convolutional neural network.
\newblock In \emph{2017 20th International conference of computer and information technology (ICCIT)}, pages 1--5. IEEE.

\bibitem[{Radford et~al.(2019)Radford, Wu, Child, Luan, Amodei, Sutskever et~al.}]{radford2019language}
Alec Radford, Jeffrey Wu, Rewon Child, David Luan, Dario Amodei, Ilya Sutskever, et~al. 2019.
\newblock Language models are unsupervised multitask learners.
\newblock \emph{OpenAI blog}, 1(8):9.

\bibitem[{Raffel et~al.(2020)Raffel, Shazeer, Roberts, Lee, Narang, Matena, Zhou, Li, and Liu}]{raffel2020exploring}
Colin Raffel, Noam Shazeer, Adam Roberts, Katherine Lee, Sharan Narang, Michael Matena, Yanqi Zhou, Wei Li, and Peter~J Liu. 2020.
\newblock Exploring the limits of transfer learning with a unified text-to-text transformer.
\newblock \emph{The Journal of Machine Learning Research}, 21(1):5485--5551.

\bibitem[{Ren et~al.(2019)Ren, Deng, He, and Che}]{ren2019generating}
Shuhuai Ren, Yihe Deng, Kun He, and Wanxiang Che. 2019.
\newblock Generating natural language adversarial examples through probability weighted word saliency.
\newblock In \emph{Proceedings of the 57th annual meeting of the association for computational linguistics}, pages 1085--1097.

\bibitem[{Ribeiro et~al.(2018)Ribeiro, Singh, and Guestrin}]{ribeiro2018semantically}
Marco~Tulio Ribeiro, Sameer Singh, and Carlos Guestrin. 2018.
\newblock Semantically equivalent adversarial rules for debugging nlp models.
\newblock In \emph{Proceedings of the 56th Annual Meeting of the Association for Computational Linguistics (volume 1: long papers)}, pages 856--865.

\bibitem[{Surinta et~al.(2015)Surinta, Karaaba, Schomaker, and Wiering}]{surinta2015recognition}
Olarik Surinta, Mahir~F Karaaba, Lambert~RB Schomaker, and Marco~A Wiering. 2015.
\newblock Recognition of handwritten characters using local gradient feature descriptors.
\newblock \emph{Engineering Applications of Artificial Intelligence}, 45:405--414.

\bibitem[{Wakabayashi et~al.(2009)Wakabayashi, Pal, Kimura, and Miyake}]{wakabayashi2009f}
Tetsushi Wakabayashi, Umapada Pal, Fumitaka Kimura, and Yasuji Miyake. 2009.
\newblock F-ratio based weighted feature extraction for similar shape character recognition.
\newblock In \emph{2009 10th International Conference on Document Analysis and Recognition}, pages 196--200. IEEE.

\bibitem[{Wallace et~al.(2019)Wallace, Feng, Kandpal, Gardner, and Singh}]{wallace2019universal}
Eric Wallace, Shi Feng, Nikhil Kandpal, Matt Gardner, and Sameer Singh. 2019.
\newblock Universal adversarial triggers for attacking and analyzing nlp.
\newblock \emph{arXiv preprint arXiv:1908.07125}.

\bibitem[{Wang et~al.(2019)Wang, Jin, and He}]{wang2019natural}
Xiaosen Wang, Hao Jin, and Kun He. 2019.
\newblock Natural language adversarial attack and defense in word level.

\bibitem[{Zang et~al.(2019)Zang, Qi, Yang, Liu, Zhang, Liu, and Sun}]{zang2019word}
Yuan Zang, Fanchao Qi, Chenghao Yang, Zhiyuan Liu, Meng Zhang, Qun Liu, and Maosong Sun. 2019.
\newblock Word-level textual adversarial attacking as combinatorial optimization.
\newblock \emph{arXiv preprint arXiv:1910.12196}.

\bibitem[{Zhang et~al.(2019)Zhang, Kishore, Wu, Weinberger, and Artzi}]{zhang2019bertscore}
Tianyi Zhang, Varsha Kishore, Felix Wu, Kilian~Q Weinberger, and Yoav Artzi. 2019.
\newblock Bertscore: Evaluating text generation with bert.
\newblock \emph{arXiv preprint arXiv:1904.09675}.

\bibitem[{Zou et~al.(2023)Zou, Wang, Kolter, and Fredrikson}]{zou2023universal}
Andy Zou, Zifan Wang, J~Zico Kolter, and Matt Fredrikson. 2023.
\newblock Universal and transferable adversarial attacks on aligned language models.
\newblock \emph{arXiv preprint arXiv:2307.15043}.

\end{thebibliography}
\end{document}